# A comparative study of large language models and human personality traits

Wang Jiaqi

School of Education, Tianjin University, Tianjin, China

Institute of Applied Psychology, Tianjin University, Tianjin, China

2022212076@tju.edu.cn

Wang bo

College of Intelligence and Computing, Tianjin University, Tianjin, China

Institute of Applied Psychology, Tianjin University, Tianjin, China

bo_wang@tju.edu.cn

Guo fa

School of Future Technology, Tianjin University, Tianjin, China

18812665576@163.com

Cheng cheng

School of Education, Tianjin University, Tianjin, China

Institute of Applied Psychology, Tianjin University, Tianjin, China

spiritcheng@sina.com

Yang li

School of Education, Tianjin University, Tianjin, China

Institute of Applied Psychology, Tianjin University, Tianjin, China

yangli@tju.edu.cn

# ABSTRACT

Large Language Models (LLMs), owing to their human-like capacities in language comprehension and generation, have been widely applied across various domains, significantly influencing human production, daily life, and cognitive patterns. In practice, they are gradually evolving from mere tools to important participants in social processes. Against this backdrop, investigating whether LLMs exhibit certain forms of personality traits and how these traits compare with those of humans has become a cutting-edge topic in psychological research. However, existing studies still fall short in addressing the applicability of conventional personality assessment methods to LLMs and in constructing LLM-specific personality theories. To explore whether LLMs possess distinct personality-like traits and how these relate to observable behaviors, this study adopts a behavior-based approach, drawing on established methodologies from human personality research. By appropriately revising relevant theoretical and measurement frameworks, we aim to construct a model describing the unique personality mechanisms of LLMs based on their behavioral patterns. This article presents three empirical studies to analyze and elaborate on these patterns.

Study 1 focuses on the test-retest stability of personality. This study compares the fluctuations in personality scale scores across multiple assessments between LLMs and humans. The results show that LLMs exhibit significantly higher score variability than humans and are more susceptible to external inputs, lacking the long-term stability characteristic of human personality. Based on these findings, the study proposes the "Distributed Personality Framework for LLMs," which conceptualizes LLM personality traits not as fixed constructs akin to human personality, but as input-driven and distributed in nature.

Study 2 investigates the cross-variant consistency in personality measurement. By varying the phrasing of questionnaire items, the study explores how LLMs and humans differ in their interpretation of personality scale content. The findings reveal that LLM responses lack consistency across item variants and are highly sensitive to changes in wording, reflecting the external dependency and instability of their personality expressions. This provides empirical support for the development of measurement tools and theoretical frameworks tailored to the unique personality characteristics of LLMs.

Study 3 examines personality retention in role-playing contexts. This study analyzes the influence of original personality traits on the role-playing behaviors of both LLMs and humans. The results indicate that the personality traits exhibited by LLMs during role-playing are significantly affected by model parameters, further confirming the high variability and external dependence of LLM-specific personality traits. These findings not only deepen the understanding of the nature of LLM personality expressions but also offer theoretical grounding and research directions for optimizing personality measurement approaches and exploring LLM behavior across diverse application scenarios.

Taken together, this research contributes to a more nuanced understanding and evaluation of LLM-specific personality traits, promotes the sustainable and responsible development of personalized AI systems, optimizes human-AI collaboration models, and extends the theoretical boundaries of personality psychology in the context of artificial intelligence—ultimately providing strong support for the construction of a human-centered intelligent society.

# Contents

# Chapter 1 Literature Review

Large Language Models (LLMs) represent a recent advancement in artificial intelligence and have brought about revolutionary changes in the field of natural language processing[1,2]. Through deep learning and self-supervised learning mechanisms, LLMs are capable of effectively learning language patterns from massive datasets and generating high-quality text. They have been widely applied across various domains, including automated customer service, personalized recommendation systems, and creative writing [3]. The emergence of LLMs has established a new milestone in the development of deep learning, not only enhancing machine language generation capabilities but also breaking through the limitations of traditional AI systems, thereby laying the foundation for the advancement of artificial intelligence toward higher-level cognitive functions [4].

Technological advancements have brought LLMs' behaviors increasingly closer to those of humans, demonstrating unprecedented human-like characteristics in areas such as knowledge, emotion, and decision-making, thus gradually positioning them at the forefront of social interaction [5]. With the rapid development of LLM technology, recent years have seen a proliferation of applications such as emotional companion robots and intelligent customer service systems, deepening the intersections between LLMs and human society [6]. The human-like features exhibited by LLMs have gradually transformed them from mere tools into significant participants in human society, and it is inevitable that they will play increasingly important social roles in the future. As the importance and necessity of LLMs in human life continue to grow, these trends indicate a pressing need to incorporate LLMs into psychological research, particularly in areas such as personality psychology, affective computing, and social cognition. Such exploration can not only help us understand how machines simulate human behavior but also provide new directions and references for research in human psychology.

The training data for LLMs are primarily derived from human sources, such as natural language texts and interaction data. The deep neural networks employed by LLMs exhibit a level of structural complexity comparable to that of human brain networks. Moreover, LLMs are capable of adjusting their outputs based on contextual information and displaying human-like emotional and decision-making behaviors across a wide range of situations. Evidently, LLMs possess a high degree of human-likeness in their inputs, outputs, and internal mechanisms. These characteristics have increasingly positioned LLMs as important tools for simulation research in psychology. For example, simulation platforms represented by projects like the Stanford Smallville have been employed in social psychological studies. Such LLM-based applications offer an ideal experimental environment for understanding and analyzing the social roles and personalities of both LLMs and humans [10].

As LLMs assume an increasing number of roles in human society, their personality characteristics have gradually become a focal point of psychological research. Numerous scholars have conducted dedicated investigations into topics such as the personality traits of LLMs, their role-playing capabilities, and the traits associated with the roles they portray. For example, a study published in Nature in 2023 examined the role-playing abilities and complex personality characteristics of LLMs [7], while a paper presented at NeurIPS 2023 explored methods for assessing LLM personality and introduced the concept of Machine Psychology Interface (MPI) [8], among others. These studies collectively indicate that LLMs are poised to play an increasingly important role in human society, and that their unique personality traits not only merit investigation but are also feasible subjects for scientific research.

In summary, the technological and societal breakthroughs of LLMs are gradually facilitating their integration into human society, making the study of their personality traits both scientifically significant and practically valuable. By systematically investigating the personality characteristics of LLMs, we can not only begin to uncover the underlying mechanisms shared between artificial intelligence and human psychology but also promote theoretical and practical advances in both LLM research and human personality psychology. Ultimately, such exploration offers new insights into understanding human psychological traits in the digital era.

## 1.1 Overview of Human Personality Theories

This section aims to systematically review the development of human personality theories, extracting perspectives and tools that may inspire comparative research on human and LLM personalities. It seeks to provide a theoretical framework for understanding the unique personality traits of LLMs and for deepening insights into human personality, as well as to offer a methodological foundation for designing assessment tools adapted to LLMs and expanding human personality measurement instruments.

## 1.1.1 Origins of Personality Research

The origins of personality research can be traced back to ancient philosophical and medical traditions. One of the earliest ideas of personality classification can be found in the humoral theory proposed by the Greek physician Hippocrates (c. 460 BCE–c. 370 BCE). He suggested that the human body was composed of four bodily fluids— blood, yellow bile, black bile, and phlegm—and that the dominance of different fluids determined individuals' behaviors, emotions, and temperaments. Building on this idea, the Roman physician Galen (c. 129–c. 200) developed four typical temperament types: sanguine, choleric, melancholic, and phlegmatic, attempting to explain psychological differences through physiological mechanisms[32]. This model not only became the dominant framework for understanding personality during the Middle Ages and the Renaissance but also laid the foundation for typological approaches to personality.

In Eastern cultures, theories of personality classification based on cosmology and the concept of "Qi" also emerged during the same period. For example, the Five Elements theory in ancient Chinese philosophy and the organ system theory in traditional Chinese medicine attempted to explain individual differences in personality by emphasizing the interactions between the body and nature. These theories highlighted the relationships among physical constitution, emotional states, and behavioral tendencies. Although they lack empirical validation by modern scientific standards, they reflect the cross-cultural diversity of early personality research and the systematic thinking of early humans regarding individual differences[32].

With the rise of the Renaissance and the Enlightenment, the Western intellectual focus on human nature gradually shifted from theology to the exploration of reason and experience. Philosophers of the 17th and 18th centuries, such as John Locke and Immanuel Kant, proposed ideas related to individual experience, rational capacity, and self-consciousness[32], laying an epistemological foundation for the later psychological study of personality.

In summary, although early personality theories were largely empirical and philosophical in nature and lacked systematic empirical support, they provided the initial conceptual frameworks and theoretical prototypes for the development of personality psychology, reflecting the long-standing tradition and continuous evolution of personality research within human cognitive systems.

With the emergence of psychology as a scientific discipline, from the late 19th to the early 20th century, personality research gradually transitioned from the domain of philosophy into that of psychology, becoming a systematic field of study. During this period, many scholars began to construct theoretical models of personality and attempted to explore the nature of personality through experimental and quantitative research methods[32].

## 1.1.2 Classical Theories of Personality

Psychoanalytic theory was proposed by Freud in 1923, who conceptualized personality as comprising the id, ego, and superego [33]. The id is the source of instinctual drives and operates according to the pleasure principle; the ego regulates the relationship between instinctual needs and the external world under the reality principle; the superego represents internalized moral standards and social norms. Jung expanded psychoanalytic theory and, in 1964, introduced the concept of the collective unconscious, proposing that psychological archetypes within it shape universal psychological patterns and individual personalities [34]. Adler, on the other hand, emphasized social interest, suggesting that the core driving force of personality development lies in overcoming feelings of inferiority and striving for superiority [35]. Psychoanalytic theory has continued to evolve, with later developments such as object relations theory and self psychology, providing a profound theoretical framework for understanding the human mind [32].

Trait theory posits that personality consists of stable and measurable traits that differ across individuals [36]. Allport first systematically introduced the concept of traits in 1937, categorizing them into cardinal traits, central traits, and secondary traits [37]. Building on this foundation, Cattell employed factor analysis in 1946 to distill 16 personality factors and developed the 16 Personality Factor Questionnaire (16PF) [38]. Subsequently, Eysenck further streamlined trait dimensions and, in 1967, proposed a three-factor model consisting of extraversion, neuroticism, and psychoticism, emphasizing the biological basis of personality [39]. Based on these studies, Costa and McCrae proposed the Five-Factor Model (FFM) in 1992, encompassing openness, conscientiousness, extraversion, agreeableness, and neuroticism, which has since become the dominant framework in contemporary trait research [40]. In recent years, trait theory has expanded into areas such as biology, cultural adaptability, and personality dynamics. Researchers have explored the neural and biological foundations of personality traits using neuroscientific methods. Meanwhile, based on cross-cultural lexical studies, Ashton and Lee proposed the HEXACO model of personality, which introduced the Honesty-Humility dimension, thereby enhancing the explanatory power of trait models in the domain of moral characteristics [41–42]. Moreover, research has increasingly focused on the situational dependency of personality and its changes across the human lifespan, promoting the development of trait theory toward a more integrative and dynamic direction [43].

Social cognitive theory posits that personality is not determined by fixed traits but is the result of dynamic interactions among individual cognition, behavior, and the environment [32]. Bandura's concept of "reciprocal determinism" emphasizes that an individual's beliefs, self-efficacy, behavior, and environmental factors interact to shape personality development. He also proposed the theory of observational learning, suggesting that individuals can adjust their behaviors by observing the actions of others and the resulting consequences [44]. Mischel challenged traditional trait theories by proposing the idea of "situational specificity," arguing that an individual's behavior is primarily influenced by specific situations rather than stable traits [32]. Together with Shoda, Mischel further developed the Cognitive-Affective Processing System (CAPS) model, which conceptualizes personality as a dynamic system regulated by cognitive-affective units, whereby different behavioral patterns are activated depending on situational triggers [45]. In recent years, social cognitive theory has been integrated with neuroscience and computational modeling methods to explore the neural mechanisms of personality. It has also been applied in fields such as artificial intelligence and human-computer interaction, for example in the development of personalized recommendation systems and dialogue-based AI that adjust user experiences based on social cognitive frameworks. Moreover, this theory has expanded within cross-cultural studies, highlighting the plasticity and adaptability of personality. These characteristics have established social cognitive theory as one of the major directions in contemporary personality psychology[46].

Trait activation theory proposes that the manifestation of personality traits is not fixed but is influenced by situational factors that activate the traits[47]. In certain contexts, individuals' traits are more likely to be expressed, particularly when the situation aligns with their internal dispositions. For instance, individuals high in extraversion are more likely to display outgoing and energetic behaviors during social interactions, while they may appear more reserved in quiet or isolated settings. Thus, the expression of traits is strongly influenced by external situations, and the theory emphasizes the dynamic interaction between personality traits and situational contexts[47]. As research has progressed, trait activation theory has expanded to place increasing emphasis on the specific role of situational factors in trait expression, particularly in applied fields such as occupational settings, education, and mental health. Meanwhile, studies on its cross-cultural applicability and situational dependence continue to advance, further refining and enriching the theoretical model.

## 1.2 Methods of Personality Measurement: An Overview

### 1.2.1 The Origins and Development of Personality Measurement

The origin of personality measurement can be traced back to the early 20th century. As psychology developed into an independent discipline, researchers began attempting to quantitatively assess personality using scientific methods. Prior to this, the understanding and description of personality were largely based on philosophical, literary, or medical frameworks, lacking systematic measurement tools[48].

In the early 20th century, personality measurement was in its infancy in psychological research, with methods being rather rudimentary and largely relying on subjective observation and non-standardized interviews[48]. Psychologists such as William James focused on individual consciousness and mental processes, providing a philosophical foundation for personality research. Francis Galton, from the late 19th to early 20th century, attempted to quantify individual differences (such as intelligence and character) through statistical methods[48], pioneering the study of individual differences, though his tools lacked systematization. Gordon Allport, in the 1920s, introduced the concept of traits, emphasizing the stability and observability of personality traits, providing theoretical support for the subsequent development of trait theory[48]. Measurement at that time mainly involved descriptive records, behavioral observations, or simple questionnaires, focusing on behavior. Although the lack of standardized tools led to lower reliability and validity of the results, these early efforts laid the theoretical foundation for personality psychology and provided significant inspiration for the development of subsequent personality scales[48].

The first generation of personality measurement tools emerged in the early 20th century. In 1917, Robert Woodworth developed the Woodworth Personal Data Sheet, which was the first to quantify personality traits through a questionnaire, focusing on dimensions such as emotional stability[48]. In 1921, Hermann Rorschach introduced the Rorschach Inkblot Test, using projection techniques to assess personality traits and emotional responses[49]. In 1949, Raymond Cattell released the 16PF scale, further advancing the standardization of personality measurement tools[48]. Although these early tools still had limitations in terms of standardization, they laid the foundation for systematic measurement in personality psychology and provided significant support for the development of subsequent scales.

With the rapid development of psychometrics and statistics, the measurement of individual personality traits across multiple dimensions has become increasingly precise[48]. In 1992, Costa and McCrae developed the NEO-PI-R[40], and in 1999, John and Srivastava proposed the BFI questionnaire[64], providing abundant empirical data for the measurement of human personality and offering a more mature theoretical framework for personality. The Big Five Personality Model has undergone continuous validation and has become a widely used personality measurement tool globally.

In recent years, the rapid development of computer science and artificial intelligence has brought new opportunities and challenges to personality measurement methods. Researchers have been exploring personality measurement methods based on multi-source data, significantly enriching the methods for measuring human personality. In the era of big data, personality measurement methods based on social media texts and large-scale online behavioral data have gradually matured[15]. These methods utilize natural language processing and machine learning algorithms (such as semantic network analysis, text sentiment analysis, etc.) to extract individual personality traits from unstructured data, providing new perspectives and tools for personality research, complementing traditional measurement methods. At the same time, researchers have begun to focus on the diversification of personality measurement methods. In addition to traditional methods, biomarkers such as neuroimaging and gene sequencing have also been gradually applied to the assessment of personality traits[17-19]. This multi-method comprehensive measurement model effectively overcomes self-report biases that may arise from single measurement tools, providing more objective measurement evidence for understanding the dynamics and situational dependence of personality, thereby improving the ecological validity of personality assessment.

Currently, traditional personality measurement methods, such as the Big Five personality model, demonstrate high predictive validity in areas such as health, job performance, and social behavior[20,21]. At the same time, personality measurement methods based on new technologies show broad application prospects in fields like personalized interventions, mental health assessments, and personality simulation in intelligent systems. The international academic community is committed to addressing the integration of different measurement methods and how to maintain the stability and validity of measurement tools in cross-cultural contexts. Domestic scholars, building on international advanced personality theories, focus on modifying and expanding these theories to align with the characteristics of Chinese culture. For example, while the Big Five personality model has been widely applied in China, some scholars have pointed out that due to cultural differences, directly applying foreign tools may have certain limitations. Therefore, there is a need to construct personality structure questionnaires that are suitable for China's national conditions. Some researchers have begun exploring the influence of Confucian thought on personality formation, aiming to build a personality theory framework with Chinese characteristics[22].

At the same time, Chinese scholars are actively exploring methods to transition from traditional questionnaire measurement to multi-source data and big data analysis. Many universities and institutions in China have developed online mental health assessment platforms, making psychological evaluation more convenient. These platforms integrate various measurement methods, such as self-reporting, behavioral tasks, and biofeedback, enabling real-time monitoring and assessment of an individual's psychological state. Such research and platforms not only enhance the timeliness and ecological validity of measurements but also provide data support for personalized interventions[23].

The rapid rise and application of large language models (LLMs) have brought significant changes to psychological measurement, and modern psychological and personality measurement methods must consider this emerging phenomenon. In recent years, many researchers have attempted to combine LLMs with traditional personality measurement methods to explore the feasibility of artificial intelligence in simulating and measuring personality traits. By analyzing data from LLMs' role-playing behavior, language patterns, and other aspects, there is hope to achieve more precise automated assessments of personality traits[24]. This type of research not only broadens the scope of personality measurement but also drives technological innovations in psychological measurement methods.

In summary, personality measurement is closely linked to the scientific development of psychology and the progress of modern technology. From the earliest philosophical and medical perspectives to the gradual improvement of modern quantitative assessment tools, personality measurement has evolved from qualitative observation to standardized quantitative testing, becoming an indispensable component of psychological research.

## 1.2.2 Personality Measurement Methods

The self-report questionnaire is a commonly used psychological measurement tool that requires participants to report their answers based on their feelings, attitudes, behaviors, or traits. These questionnaires typically use closed-ended questions, allowing participants to respond within a range of options. The goal is to assess an individual's psychological state, personality traits, or other related variables. Self-report questionnaires are widely used in psychological research and clinical assessment due to their ease of use, low cost, and ability to cover a broad range of psychological dimensions. However, they are also subject to social desirability bias and limitations in self-awareness, which may affect the accuracy of the measurement results[48].

The other-report method is a personality or psychological trait measurement approach that gathers data through others' observations and evaluations of the participant. These individuals, often close friends, family members, or colleagues, assess the participant's behaviors, attitudes, and emotions based on their long-term interactions and understanding. Compared to self-report questionnaires, the other-report method can reduce social desirability bias and provide a more objective reflection of the individual's behavioral characteristics. However, it may also be influenced by the evaluator's subjective cognition and biases, and there may be discrepancies in ratings between different evaluators[48].

The behavioral observation method is a technique for collecting data by directly observing an individual's behavior in natural environments or specific situations. Researchers observe or record the individual's behaviors and infer their psychological state or personality traits based on these behaviors. The advantage of behavioral observation is that it provides actual, behavior-based information, independent of self-reports, thus avoiding the biases associated with self-report methods. However, behavioral observation may be influenced by the subjective judgments of the observer, and behaviors in different contexts may be highly context-dependent, requiring cautious interpretation[48].

The projective test is a psychological measurement method that reveals an individual's underlying emotions, motivations, or personality traits by having them respond to ambiguous or unclear stimuli. Classic projective tests include the Rorschach Inkblot Test and the Thematic Apperception Test (TAT). This method is based on the projection hypothesis, which suggests that when individuals interpret ambiguous situations, they project their inner world onto external stimuli, thereby revealing their latent psychological dynamics. The advantage of projective tests is that they can avoid biases related to social expectations. However, due to the subjective nature of interpretation, the results may not be stable or consistent[49-50].

Physiological measurement is a method that assesses an individual's psychological state, emotional response, or personality traits by recording their physiological responses, such as heart rate, skin conductivity, brain waves, etc. Physiological measurements provide objective, real-time data that reflect an individual's physiological reactions and psychological activities in specific situations[51]. They are commonly used to study psychological phenomena such as emotional responses, stress, and anxiety. However, the limitation of physiological measurement lies in its complex interpretation, as physiological responses cannot be directly correlated with specific psychological states. Therefore, it is often necessary to combine this method with other measurement approaches to gain a comprehensive understanding of an individual's psychological characteristics.

Emerging technological methods for personality measurement leverage modern technologies such as artificial intelligence, big data analysis, and virtual reality, offering more innovative and efficient approaches to personality assessment. Big data analysis can infer personality traits like extraversion and neuroticism by analyzing behavior patterns from social media, online activities, or digital footprints, overcoming the limitations of traditional measurement tools. Artificial intelligence technologies, such as natural language processing, speech analysis, and facial expression recognition, can automatically infer personality dimensions from speech, sound, or facial expressions. Additionally, virtual reality technology simulates complex situations in controlled virtual environments to observe an individual's behavioral responses and assess their personality traits in different contexts. These methods not only improve the precision of personality measurement but also enable assessments in more natural and dynamic environments, providing richer and broader data support for psychological research and applications[48].

Overall, human personality theories are rich and diverse, ranging from psychoanalysis and unconscious drives to trait theories with quantifiable structures, and from social cognition to trait activation's situational dynamics, each offering various perspectives on human personality. These theories provide diverse references for LLM trait analysis. Measurement methods have evolved from self-reported scales to peer assessments, behavioral observations, physiological techniques, and data-driven emerging methods, showcasing progress from qualitative to quantitative approaches. These theories and methods offer valuable insights for LLM personality research, such as: trait theories supporting quantitative evaluation, social cognition and trait activation emphasizing situational design, and emerging technologies adapting to data characteristics. However, the lack of awareness, emotional experience, and physiological basis in LLM traits means that directly transferring human methods requires careful adjustment to align with their data-driven dynamic traits.

## 1.3 Large Language Models

This section aims to introduce the definition, technical characteristics, and behavioral traits of LLMs, providing a technical and theoretical foundation for the subsequent discussion on personality measurement.

LLMs are a class of natural language processing models based on deep learning, which are trained on massive amounts of textual data and can understand, generate, and process natural language[46]. LLMs capture the semantic relationships and contextual information between words through multi-layer neural networks, with their core architecture typically being the Transformer[46]. Representative models include BERT, GPT series, and T5[52-54]. The scale of LLMs is often measured by the number of parameters, for example, GPT-3 has 175 billion parameters[4], and PaLM has 540 billion parameters[55]. As the number of parameters increases, the quality and complexity of language generation by LLMs also significantly improve.

From a psychological perspective, LLMs can be viewed as a "behavior simulator," where their outputs mimic human language patterns. However, the generation process relies on statistical learning rather than cognitive or emotional experience. This technical characteristic gives LLMs dual significance in the study of human-like traits:

they can serve both as a research subject, reflecting data-driven behavior patterns, and as a tool to assist in the dynamic analysis of human personality.

Although the outputs of LLMs closely resemble those of humans, their generation process relies on statistical learning rather than cognitive or emotional experiences. This technical characteristic endows LLMs with dual significance in the study of human-like traits: on one hand, LLMs serve as research subjects, reflecting data-driven behavioral patterns; on the other hand, they can function as tools to assist in the analysis and dynamic understanding of human personality, especially by providing additional perspectives and support in large-scale data analysis and contextualized research.

LLMs possess several unique technical characteristics that enable them to excel in the field of natural language processing.

First, LLMs demonstrate outstanding language generation capabilities. They can generate coherent text through autoregressive or bidirectional encoding mechanisms. For example, GPT-3 can produce high-quality article continuations based on a given prompt, showcasing its powerful text generation ability[4]. BERT, on the other hand, is able to understand relationships between sentences by comprehensively capturing contextual information through its bidirectional encoder, exhibiting exceptional semantic understanding[52].

Second, LLMs exhibit strong contextual sensitivity, with their outputs heavily influenced by the surrounding context. In natural language processing and human-computer interaction, this contextual sensitivity allows LLMs to flexibly adjust their understanding and generated text according to different situations, thus better aligning with the natural usage of human language[56].

Furthermore, LLMs possess an extensive knowledge base. Their training data are drawn from diverse sources such as books and websites, endowing them with a broad background of information[54]. This large-scale knowledge reservoir enables LLMs to handle a wide range of topics, answer questions, and even generate content in specialized fields, demonstrating considerable breadth of knowledge and adaptability.

Finally, LLMs possess a certain degree of multimodal expansion capability. For instance, models such as GPT-4V and Gemini can process not only text but also integrate image information, demonstrating strong cross-modal abilities[57,58]. This means that LLMs are not limited to language processing; they can also incorporate multiple sources of information, such as images, to handle more complex tasks, thereby expanding their range of applications.

Although LLMs can perform well across various scenarios, their outputs are influenced by factors such as input prompts and context, resulting in behavioral characteristics such as contextual dependency, adaptability, dynamism, and instability[16].

The behavior of LLMs exhibits strong contextual dependency. For example, in an interview setting, when the model is asked a question such as "Please describe your strengths and weaknesses," its response tends to be more formal and logically structured, often displaying higher levels of conscientiousness and openness. However, in a casual conversation setting, when asked the same question, the model's response becomes more relaxed and emotional, showing stronger agreeableness and lower extraversion. Research has also found that fine-tuned models demonstrate certain differences in personality traits. For instance, fine-tuned models may exhibit more agreeable and emotionally stable responses in specific contexts, whereas such traits are less apparent in base models (i.e., models without fine-tuning)[60].

LLMs possess strong adaptability, enabling them to adjust their output style according to different tasks or roles. For example, when an LLM is asked to simulate the role of a doctor, its output typically exhibits a high degree of professionalism, with a formal tone focused on providing rational analyses and scientific evidence. In this role, the LLM tends to avoid emotional language, instead emphasizing symptom analysis and scientifically grounded treatment suggestions, maintaining a professional and calm tone. In contrast, when the LLM is asked to play the role of a customer service representative, its responses are noticeably more friendly, empathetic, and aimed at making customers feel understood and supported. In this role, the LLM emphasizes emotional reassurance and conveys care through warm and approachable language[61].

The output of LLMs also exhibits a certain degree of dynamism and instability. Variations in input prompts, data updates, or model fine-tuning can lead to fluctuations in their behavioral outputs. For example, researchers attempted to have GPT-3 simulate the political attitudes and personality traits of different populations and found that, while the model could generally produce response distributions consistent with specific demographic groups, repeated generations under the same role settings (e.g., identical demographic characteristics) showed semantic and stylistic variations. This indicates that even when the "conditions" are identical, the "population simulation samples" generated by the model at different times are not entirely consistent, highlighting the dynamic and non-deterministic nature of LLM outputs[62].

Although the outputs of LLMs vary with changes in input and context, they can be optimized through reinforcement learning. For instance, during the training of InstructGPT, researchers employed Reinforcement Learning from Human Feedback (RLHF) to adjust the model's outputs based on human preferences, thereby improving the relevance and controllability of the generated text[63]. This optimization approach has, to some extent, enhanced the stability of LLMs across different tasks and contexts, making their behavior more aligned with human expectations.

By leveraging its powerful language generation capabilities and strong sensitivity to context, LLM demonstrates human-like behavioral performances. However, its behavioral characteristics—such as contextual dependence, adaptability, and dynamism—also highlight key distinctions from human personality traits. The outputs of LLMs are based on statistical learning, rather than on cognitive processing or emotional experiences like humans. Thus, while LLMs can mimic human behavioral patterns, it remains uncertain whether they truly possess self-awareness or stable personalities.

Understanding the technical and behavioral features of LLMs is crucial for exploring methods of measuring their personality traits. It also contributes to deepening our knowledge of human personality mechanisms and expands the application and development of personality theory in the context of artificial intelligence.

## 1.4 Overview of Personality Research in Artificial Agents

### 1.4.1 Personality Measurement in Artificial Agents

In recent years, with the widespread use of LLMs such as GPT-3, ChatGPT, and GPT-4, personality measurement in artificial intelligence (AI) has become a hot topic in psychological research.

International research on personality measurement in AI systems began relatively early. From the early rule-based simple AI systems to the current LLMs, researchers have accumulated a relatively rich body of theoretical and empirical studies. In contrast, research in China began later, but with the rise of domestic large language models such as Wenxin Yiyan and DeepSeek, research in the field of AI personality measurement is gradually accelerating. Academic circles worldwide are paying close attention to the humanoid performance of these models in practical applications, particularly whether they possess personality traits similar to those of humans, and discussions have emerged surrounding personality structure, measurement methods, and influencing factors.

Foreign research has laid the foundation for AI personality measurement, developing various methods ranging from subjective perception assessments to quantitative analysis. Nass et al. were among the pioneers in exploring AI personality measurement. In 1995, they conducted a study using questionnaires and behavioral measurement methods to observe how users attributed fixed response patterns of computers (such as politeness, directness, etc.) to personality traits[26]. This user perception-based subjective measurement established the groundwork for early AI personality measurement studies. Furthermore, this study revealed the human tendency to anthropomorphize when interacting with artificial intelligence. As artificial intelligence and large language models advanced, international scholars began to explore how to scientifically assess AI's personality performance. They gradually introduced psychological personality measurement tools (such as the Big Five personality questionnaire) to quantify AI's personality traits. However, early measurement methods primarily relied on users' subjective perceptions, lacking systematic control of specific language features and quantitative analysis. With the development of dialogue systems, Mairesse and Walker proposed a more systematic method for AI personality measurement in 2007[27]. They used the PERSONAGE system to manipulate the language output characteristics of AI, such as sentence length, vocabulary choices, and grammar structures, to generate dialogue styles that matched specific personality traits (e.g., "extroversion" or "introversion"). They then employed user rating methods, asking participants to classify these AI-generated texts into personality categories and measured the classification accuracy (which ultimately reached 70%). This study not only revealed that AI's language style affects users' perception of its personality but also introduced a psychological personality measurement framework. It provided an important experimental paradigm and methodological reference for subsequent AI personality measurement based on LLMs.

In recent years, with the development of LLMs, researchers have found that these models can exhibit more complex personality expression patterns, further driving innovation in personality measurement methods. In 2023, Bommasani et al. drew on the consistency measurement approach from personality psychology, observing individuals' stable behavior patterns across different contexts to infer personality traits. They measured the personalities of 30 LLMs and discovered the cross-task consistency of LLM behavior[28]. This study provided a more systematic and quantitative approach to AI personality measurement, enhancing the objectivity and reproducibility of the measurements. It brought AI personality measurement closer to standardized methods in psychology, offering an expandable framework for future research and aligning AI personality traits with psychological theories in a more systematic way. Bender et al.'s 2021 study sparked academic discussion about the nature of AI personality. They emphasized that while LLMs are statistical language models without true consciousness, their stable language output patterns are often perceived as personality-like traits in certain contexts[16]. This perspective challenges traditional views of artificial intelligence, prompting scholars to reconsider and redefine the similarities between AI and human personality expressions. The study provided theoretical support for advances in AI personality measurement methods, particularly by revealing the potential of language models in terms of stability and consistency. This laid the groundwork for future research into AI systems that are more natural and personalized in human interaction. Jiang et al.'s 2023 research offered new perspectives and methods for measuring the personalities of AI and LLMs[8]. Unlike previous methods that relied on direct questionnaires or self-report scales, this study systematically analyzed the language features of LLM outputs and quantitatively assessed them using the Big Five personality theory. This approach introduced a new method for AI personality measurement, applying traditional personality measurement theories to AI. It facilitated the transition of AI personality measurement from perception-based assessments to vectorized and standardized approaches, providing more scientific and systematic support for the personalized design and user interaction of AI systems.

Research on AI personality measurement in China started relatively late, but has made progress in recent years with the development of domestic LLMs such as Wenxin Yiyin, iFlytek Spark, DeepSeek, and others. Early research in China focused mainly on the humanization design of traditional dialogue systems, lacking systematic methods for measuring AI personalities, and often relying on human subjective evaluation. With the introduction of LLM technology, there has been increasing attention to the humanoid characteristics of these models. Some studies have begun to draw on the language feature analysis methods proposed by Mairesse and Walker in 2007[27], attempting to assess the dialogue patterns of domestic LLMs. However, an independent framework has yet to be formed. Compared to foreign research, domestic studies on AI personality measurement are more application-oriented, focusing on optimizing user interaction experiences in fields such as education and customer service. Theoretical discussions of LLM personality traits are relatively limited. Additionally, domestic research still faces challenges in terms of large-scale data support and systematic experimental design, which limits the ability to conduct in-depth quantitative analysis of AI personalities.

From the perspective of model development, early AI models (such as expert systems in the 1990s and conversational agents in the 2000s) exhibited simpler personalities. The personalities of these models were mainly preset by developers, and corresponding personality measurement research was relatively rudimentary, mostly relying on user perceptions and simple experiments[26]. In contrast, modern LLMs (such as GPT-4 and DeepSeek) demonstrate more dynamic humanoid traits due to large-scale training data and complex architectures. The personality of LLMs has shifted from being preset by developers to being driven by data, and the methods for measuring personality have become more complex, emphasizing interdisciplinary collaboration and objectivity and precision[8].

Overall, foreign research is more mature in both theory and methodology, supported by abundant data and interdisciplinary collaboration, with a greater focus on quantification. Domestic research is more focused on specific application scenarios, started later, and faces limitations in data scale, remaining in the exploratory stage.

The large and opaque training data of LLMs, combined with the black-box nature of their learning mechanisms, makes it difficult to clearly define the process and influencing factors behind the formation of their personalities. This presents significant challenges for measuring LLM personalities, as researchers struggle to determine whether the behavior patterns of the models are driven by stable personality traits or merely reflect statistical patterns in the training data, increasing the uncontrollability and interpretive difficulty of measurement results. Current measurement methods are still inadequate to address the complexity of LLMs.

Existing LLM measurement tools are often modeled after human personality inventories, but the applicability of human personality scales to LLMs remains uncertain. Human personality inventories do not account for the fact that LLMs lack human physiological experiences and emotional processing, so the fit between human personality inventories and LLM personality traits requires further research.

Currently, there is insufficient theoretical exploration of the personality structure of LLMs. The personality structure of LLMs directly impacts both the methods of personality measurement and the interpretation of measurement results. Without a clear understanding of LLM personality traits, it becomes difficult to select and apply measurement tools appropriately, which hinders the interpretation of results and limits their value in both psychological research and practical applications. Given the distributed data training characteristics of LLMs, their personalities are likely to be distributed and probabilistic, rather than fixed psychological traits like those in humans. Moreover, there is a lack of research on how LLM personality traits influence practical applications. Foreign research has paid less attention to the performance of LLMs in specific application scenarios, while domestic research focuses more on application (such as politeness in customer service) but has yet to systematically analyze how LLM personality traits affect their performance in different contexts. In the future, it will be necessary to further develop theoretical frameworks and measurement methods for LLM personalities to ensure the scientific validity and reliability of LLM personality measurement.

## 1.4.2 Comparison between LLM and Human Personality Analysis

LLM personality analysis and human personality analysis differ significantly in goals, methods, and traits. Human personality analysis, based on psychology, aims to reveal the intrinsic traits shaped by physiological, social experiences, and psychological mechanisms, typically measured using standard personality scales. This measurement relies on subjective reports and behavioral observations, assuming that personality traits are relatively stable. LLM personality analysis, on the other hand, focuses on human-like traits in model outputs, exploring whether their language or behavior patterns resemble human traits. For example, Shanahan et al. analyzed LLM behavior in role-playing tasks to infer its underlying personality structure[7].

Applying human personality measurement strategies to LLMs has three significant advantages. First, human personality measurement methods have a mature theoretical foundation and high implementation convenience. Human personality scales, as standardized tools in psychology, have been widely validated for reliability and validity (with Cronbach's α usually above 0.8, indicating high internal consistency and measurement precision). Researchers can design structured prompts for LLMs based on human personality scale items, such as "Do you tend to seek adventurous activities?", guiding the model to generate quantifiable responses that produce scores similar to human data. Second, this strategy facilitates comparisons, allowing human and different types of LLMs to be placed within the same measurement framework, enabling trait comparisons across various subjects and providing quantitative support for research on differences between categories. Lastly, this method is efficient and practical. By utilizing existing human personality scales, researchers can quickly explore LLM traits without having to create new evaluation tools, making it especially suitable for early-stage research. However, these advantages rely on the surface similarity between LLM outputs and human language, rather than a substantial equivalence with human personality mechanisms.

Compared to the advantages of directly applying human personality measurement scales to LLMs, this strategy has more noticeable drawbacks. These arise from significant differences between human and LLM personalities in terms of their formation, behavioral nature, and internal mechanisms.

Human personality is rooted in physiological and social experiences, with each person having a unique life history. When humans respond to scale questions, they assess their inner psychological state based on sensory input, emotional experiences, or self-awareness. LLMs, however, do not have such a foundation. LLMs generate outputs based on large-scale data statistics without physiological perception or subjective awareness. Their responses are often shaped by the dominance of positive or neutral expressions in the training corpus, reflecting statistical patterns rather than any real self-awareness. For example, Bodroža et al. (2024) used the BFI (Big Five Inventory) and TIPI (Ten Item Personality Inventory) to measure the personalities of seven LLMs multiple times and found that their traits were significantly influenced by factors such as prompt words, model parameters, and API versions. The measurement results did not reflect stable internal states, indicating that human and LLM personalities are clearly different[31].

LLM's traits exhibit high dynamism, in stark contrast to the relative stability of human personality, which may explain why human personality scales and theories are not suitable for LLMs. Human personality is shaped over a long period of life experience and evolves slowly over time, showing consistency across different contexts. In contrast, LLM outputs are immediately influenced by prompts, training data, or algorithm adjustments, causing their personality traits to fluctuate significantly in a short period. In 2024, Bodroža et al. measured the personality of the same LLM using two types of prompts: short and guided prompts. They found that even when the same question was asked, changes in the prompt significantly impacted the model's response content and tone. They noted that human personality measurement tools struggle to effectively capture LLM's dynamic traits and suggested that future research should focus on improving the stability and reliability of LLM personality assessments[31].

On a deeper level, the internal mechanisms of human and LLM personalities differ fundamentally, which further diminishes the effectiveness of human personality scales in measuring LLM personality. Human personality formation relies on complex cognitive processes, including self-reflection, emotional regulation, and memory integration, with knowledge accumulated through sensory interaction and social learning, forming a coherent psychological structure. In contrast, LLMs acquire knowledge through statistical pattern recognition, with understanding based on data correlations rather than real experience. For instance, a human's understanding of "adventure" is supported by extensive sensory experience, whereas an LLM's understanding of "adventure" may simply simulate word frequencies in its training data, without causal reasoning or subjective motivation. Additionally, humans possess self-awareness, enabling them to adjust behavior to align with their personality traits (e.g., introverts avoiding social interactions) and have intrinsic drives. Whether LLMs possess self-awareness remains uncertain, as their outputs are driven by external inputs. These deep-seated differences in mechanisms suggest that the measurement results of LLMs may reflect surface-level language patterns rather than deep-seated personality traits.

There are still many differences between LLMs and humans, and human personality scales, designed for human psychological structures, struggle to bridge these differences. In this context, directly applying human personality theories and measurement tools to LLMs is likely to distort the results of LLM personality measurement.

Another issue is the interpretation of LLM personality measurement results obtained using human personality measurement methods. Human personality scale analysis is grounded in solid theoretical foundations, and the scores effectively reflect human psychological states. However, LLMs lack a personality theory of their own, meaning their personality measurement scores may simply be projections of training data, making it difficult to interpret these scores meaningfully or logically.

LLM personality and human personality differ significantly in their foundations, stability, and interpretability. While directly applying human measurement strategies is convenient and offers comparative value, it can lead to biases and misunderstandings because LLMs lack physiological experience, have dynamic traits, and do not possess internal psychological processes similar to humans. Future research should focus on developing new methods tailored to the characteristics of LLMs, rather than relying on human-based tools. When studying LLM personality, researchers must acknowledge the differences in traits and methods between LLM and human personalities. Blindly applying human personality theories without targeted, in-depth consideration may fail to capture the true statistical characteristics and generative mechanisms behind LLM personality data. The unique mechanisms of LLMs may require researchers to reconsider some common human personality concepts, such as personality itself.

Additionally, when researching and applying LLM personality, it is crucial to focus on preventing LLM personality outputs from misleading users, explaining the LLM's outputs effectively, and addressing potential biases and improper applications of LLMs, ensuring that the research and applications meet ethical and social responsibility standards. Therefore, this paper will primarily investigate the impact of question framing on LLM personality measurement results, analyze the differences between LLM personality traits and human personality traits, and explore how these differences affect the interpretation of measurement results and practical applications.

### 1.4.3 Foundations for Comparing LLM and Human Personality

This section introduces indirect measurement methods and the Cognitive-Affective Processing System (CAPS) theory, considering the technical characteristics of LLMs, to provide theoretical support for adjusting the phrasing of personality scale questions and offer new perspectives for precise measurement of LLM traits.

#### 1.4.3.1 Indirect Measurement

Indirect measurement is a method used to infer psychological traits by analyzing individuals' behavioral responses or automatic cognitive processing. This approach does not rely on subjective reports, and its theoretical foundation is based on cognitive association theory[18]. According to cognitive association theory, individuals' cognitive links to different concepts can be reflected in response patterns, where the strength of associations between words or concepts can reveal underlying attitudes, beliefs, or personality traits. For example, the Implicit Association Test (IAT) reveals implicit biases by measuring differences in reaction times during categorization tasks[18], while sentence completion tasks infer individuals' motivations or psychological tendencies based on how they complete unfinished sentences[67]. Research has shown that indirect measurement has advantages over self-report methods in reducing social desirability bias and uncovering implicit traits[68].

For LLMs, indirect measurement may be more applicable than directly inquiring about their traits. Since LLMs lack self-awareness and emotional experience, directly asking a question (such as "Are you extroverted?") often only yields responses based on patterns in training data, rather than reflecting genuine behavioral tendencies. Indirect measurement methods can infer latent traits through the language generation patterns of LLMs. For example, prompting an LLM to complete a sentence like "In a team, I ___" may reveal internal linguistic biases. Studies have found that LLM outputs contain implicit biases present in the training data, and indirect measurement can effectively uncover these statistical characteristics [69], rather than simply imposing human personality frameworks. This approach not only better aligns with the technical characteristics of LLMs but also provides new tools for psychological research on implicit traits.

### 1.4.3.2 Cognitive-Affective Processing System Theory

The Cognitive-Affective Processing System (CAPS) theory, proposed by Mischel and Shoda, posits that human personality is context-dependent [17]. Unlike traditional theories of personality, CAPS suggests that personality is not composed of fixed traits, but rather consists of dynamic patterns of responses exhibited by individuals across different situations. Specifically, an individual's behavioral patterns change depending on situational cues, but display consistency across similar contexts. This perspective challenges traditional trait models of personality and emphasizes the significant role of situational factors in personality expression.

The CAPS theory provides a new perspective for understanding the dynamic nature of personality. Mischel and Shoda argue that individual behavior is not determined solely by a single trait but is shaped by the interaction between an individual's cognitive-affective system and external situations [71]. For example, an individual may choose to avoid conflict when facing confrontation or tend to experience anxiety under stress, with such behavioral responses jointly driven by the individual's internal cognitive network (such as beliefs and emotional reactions) and situational cues.

To explain individual behavioral variation across different situations, Mischel proposed the "if...then..." framework, which has been widely applied in studying the consistency and variability of individual behavior [70]. This framework posits that the expression of personality is context-driven, and that an individual's behavioral responses in specific situations can be summarized using "if...then..." patterns. For example, if a person experiences stress, then they may display anxiety, or if a person is in a social setting, then they may tend to exhibit more extroverted behavior. Such "if...then..." patterns reveal the regularities of individual behavior across different situations and highlight the context-dependency of behavior.

To address the limitations of traditional personality psychology regarding the understanding of behavioral consistency and variability, Shoda et al. proposed the High-frequency Repeated Within-Person (HRWP) method, which captures behavioral consistency and variability through long-term repeated measurements of individuals in similar contexts [72]. This extension not only strengthened the empirical support for the CAPS theory but also provided a new tool for exploring the adaptability of individual behavior across different situations.

The technical characteristics of LLMs cause their outputs to vary according to input prompts and contextual cues, a pattern highly consistent with the context-dependency described in CAPS, making CAPS and the "if...then..." framework particularly suitable for analyzing the dynamic features of LLMs.

By applying the CAPS model and the "if...then..." framework, we can design specific situational cues (such as task types and role settings) and analyze the response patterns of LLMs under different circumstances. This not only provides theoretical support for measuring LLM traits but also reciprocally advances situational analysis in human personality research. The dynamic perspective of CAPS helps deepen our understanding of the relationship between behavior and context, thereby promoting a more detailed and systematic study of both human and non-human intelligent behavior.

# Chapter 2 Research Framework

## 2.1 Problem Statement

To better advance human psychological research and the application of LLMs in psychological studies, it is necessary to conduct an in-depth and objective assessment of the human-likeness of LLMs, particularly their personality traits. There are fundamental differences between the ways LLMs and humans acquire learning and experience: LLMs do not possess human physiological experiences or social histories. Their knowledge is primarily derived from large-scale corpus-based statistical learning, and their language outputs are the result of algorithmic optimization based on probability distributions, rather than cognitive structures and behavioral patterns constructed through sensory input, emotional experience, or social interaction.

Bender et al. (2021) pointed out that the language generation process of LLMs is a form of stochastic imitation based on statistical probabilities rather than an expression grounded in meaning, and that their behavior lacks support from motivation or intention [16]. This suggests that the human-like behaviors of LLMs may merely reflect tendencies present in their training data, rather than indicating that the models possess personality traits equivalent to those of humans.

The differences between LLM and human personality traits will influence how people perceive LLMs, and will further impact their positioning and usage in social cognition, ethics, and psychological research.

In attempting to study LLM-specific personality traits using human personality research methods, numerous issues have been revealed.

First, when using human personality inventories to measure LLM personality, it is difficult to accurately interpret the meaning behind the options selected by the LLM. Human personality inventories are designed specifically for human traits, grounded in human personality theories. Human responses to inventory items reflect subjective judgments, revealing underlying psychological processes or personality traits. However, whether LLMs' selections carry any meaningful internal processes is uncertain. For example, in the BFI inventory, when facing the item "Do I like adventure?", an LLM might select "3." This response could reflect a neutral stance, result from random generation, or arise from an inaccurate understanding of the question. Due to the possible inapplicability of human personality research methods to LLMs, we lack explanatory frameworks for the underlying causes of LLM choices, cannot accurately interpret the meaning of LLM behaviors, and this phenomenon severely undermines the reliability of such measurements.

Second, LLMs are unable to appropriately respond to items involving self-awareness and physiological experiences. Personality inventories often include questions such as "Are you talkative?" or "Do you often experience stomach aches?", which humans can answer based on self-awareness and bodily experiences. LLMs, lacking sensory input and life history, may struggle to comprehend the meaning of such items; their responses are merely simulations based on training corpora. While these items are critical for describing human personality, they appear ineffective in capturing LLM-specific personality traits.

Moreover, research has shown that LLM personality measurement results exhibit significant instability. In 2023, Huang et al. evaluated the personalities of ChatGPT and Bard using the MBTI inventory, finding that ChatGPT corresponded to the ENFJ type and Bard to the ISTJ type [66]. However, the authors later observed a lack of replicability in their results. In the same year, Jiang et al. assessed GPT-3's personality traits using the BFI inventory and found considerable score fluctuations in the Openness dimension, with an average of 51.8 and a standard deviation of 9.7 (out of 100), indicating low result stability [8]. These findings suggest that LLM measurement outcomes may reflect prompt design or training data biases rather than consistent personality traits.

If the differences between LLM and human personality traits are overlooked, and human personality theories and inventories (such as BFI, MBTI) are directly applied to LLMs, multiple harms to human society could result.

First, because these inventories are based on human consciousness, emotion, and physiology, and because LLMs may merely function as statistical language generation tools, the personality measurement results of LLMs are likely to be distorted. People may misinterpret LLM behaviors, overestimate their capabilities, or develop undue trust, leading to erroneous decisions and ethical risks, especially in critical fields such as healthcare and law. Second, this approach could blur the essential differences between humans and AI, weakening the uniqueness of human emotion and consciousness, and triggering identity crises and social injustice. Finally, inappropriate personality evaluations could obscure AI biases and limitations, reduce system transparency, hinder the scientific exploration of non-human intelligences, waste research resources, and mislead technological applications. Thus, it is necessary to systematically study the personality traits of LLMs and the differences between LLM and human personalities.

This study explores the human-like personality features of LLMs, taking humans as a reference point. It analyzes the applicability of human personality inventories to LLMs and examines how different prompting approaches affect LLM personality measurement outcomes. Additionally, it investigates the extent to which LLMs exhibit human-like personality traits and compares similarities and differences between these traits and human personality traits. The aim of this study is to deeply analyze the human-like characteristics of LLMs and to reveal the distinctions and connections between LLM and human personalities. To achieve this goal, the study employs multiple methods, including the design of variant inventories, to systematically explore the above issues.

The focus of this research is to reference human personality theories and analytical methods, while fully considering the technical characteristics of LLMs and the theoretical foundations of human personality inventories, to explore LLM-specific personality traits. Through empirical methods, the study compares the external behavioral manifestations of LLMs and humans, and by appropriately revising related human personality theories and measurement methods, attempts to construct personality theories and measurement techniques better suited to LLMs. This will enable a systematic description and analysis of LLM behavioral patterns and facilitate the better application and integration of LLMs into psychological research in the future.

The study of the differences between LLM and human personality traits not only provides a scientific basis for personality psychology research, the development of AI ethical norms, and human-AI interaction practices, but also deepens the understanding of human psychological mechanisms and promotes the development of a human-centered, harmonious, and efficient human-AI symbiotic society. Moreover, this research may inform the development of advanced psychological research tools and the optimization of AI behavioral design, while avoiding anthropomorphic misunderstandings of LLMs and clarifying the boundaries between humans and AI.

## 2.2 Research Content

Taking humans as the reference group, this study focuses on two key issues: the engineering problem of how to draw upon human personality inventories to develop methods suitable for measuring LLM-specific personality traits, and the scientific problem of exploring the extent to which LLMs exhibit human-like traits and the similarities and differences between LLM and human personalities. The analysis of LLM personality characteristics is conducted from three main aspects, as outlined below.

### I. Comparison of Re-test Reliability between Human Personality and LLM Personality, and Research on LLM Personality Theory

This section aims to compare the stability differences in multiple personality measurements between human and LLM personalities from the perspective of re-test reliability, explore the distribution characteristics of LLM personality traits, and propose theoretical frameworks. Based on the concept of re-test reliability in psychology, the study takes the stability of human personality measurement as a reference, evaluating the score fluctuations of LLM in multiple measurements, thus determining the stability of its personality traits. Specifically, the research uses classic personality inventories and conducts multiple measurements on both human participants and LLMs, calculating statistical parameters such as correlation coefficients to compare and verify the stability of human and LLM personalities in measurement. The results show that LLM personality measurements are highly influenced by input, lacking the stability inherent in human personalities based on internal psychological structures. Based on this, this paper proposes the Distributed Personality Theory framework for LLMs, suggesting that their personality expression is a dynamic distribution rather than a fixed trait. This framework helps to redefine the human understanding of humanoid behavior traits in intelligent agents and provides a theoretical foundation for the future modeling, management, and ethical regulation of artificial intelligence.

### II. Cross-Variant Consistency Comparison of Human Personality and LLM Personality Measurement

The aim of this study is to explore the impact of different question formats on human and LLM personality inventory results, assess the applicability of human personality inventories to LLMs, compare the differences in the ability to understand the core meaning of the inventories between humans and LLMs, and reveal the intrinsic mechanism characteristics of LLMs. Combining the Cognitive-Affective Processing

System (CAPS) theory, indirect measurement methods, and the "if…then…" framework, this study analyzes the performance differences between LLMs and humans in personality measurement by varying the phrasing of classic personality inventory items. Specifically, this study designs multiple versions of inventory items (including direct questions, indirect descriptions, and situational simulations), applying them to both human participants and LLMs to compare the stability and consistency of their scores under different question formats. The results show that LLMs cannot consistently understand the core meaning of personality measurement items like humans, and the measurement results do not exhibit the same consistency and stability as human results. Based on this, this paper delves into the internal mechanisms of LLM personality, providing theoretical support for understanding the similarities and differences between artificial intelligence and human psychological characteristics, and laying the foundation for more rational applications of LLMs.

**III. Role-playing and Personality Trait Retention Comparison between LLM and Humans**

This study aims to explore the impact of original personality traits on role-playing personalities by comparing the performance of humans and LLMs in role-playing tasks, deepening the understanding of the differences between human and LLM personalities, and offering a new perspective for LLM personality theory. Specifically, the study sets up multiple roles and asks both humans and LLMs to perform role-playing tasks, evaluating the differences between the role-playing results and the original personality traits, and analyzing whether LLM personality performance is influenced by original personality traits. The results show that the degree to which LLMs are influenced by original personalities depends on specific model parameters. This reveals the deep differences between human and LLM personalities, providing a new perspective for constructing LLM personality theory. It helps humans more clearly define the uniqueness of their psychological traits and avoids blurring the boundaries between humans and machines in technological development, preventing the misjudgment of artificial intelligence's psychological attributes.

This study's research framework and the logical relationships between its chapters are as follows:

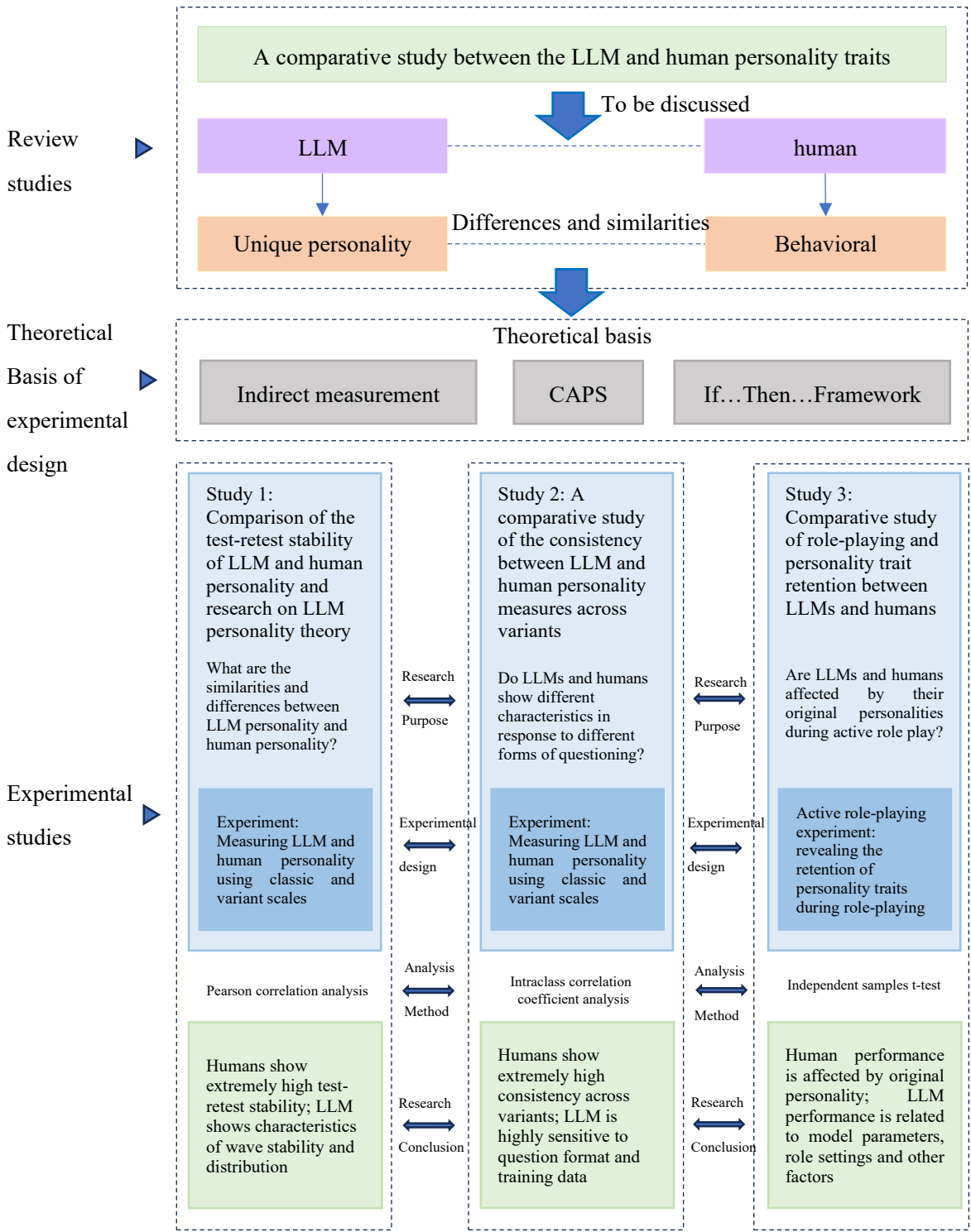

Figure 2-1 Logical framework diagram

## 2.3 Research Significance

### 2.3.1 Theoretical Significance

First, this study contributes to expanding the boundaries of human personality theory. By revealing the similarities and differences between the personality characteristics generated by LLMs through statistical learning and the personality traits formed by humans through experiential processes, this research provides new perspectives for traditional personality theories, particularly in areas such as individual expression, emotional response, and social cognition. The humanoid features exhibited by LLMs offer novel insights into understanding the mechanisms underlying human personality formation, promoting interdisciplinary integration between psychology and artificial intelligence, and driving innovation in personality theory development [11,12].

Second, this study helps to promote human-centered interdisciplinary integration. Research on LLM personality bridges the fields of psychology, philosophy, and human-computer interaction, directly responding to human curiosity about intelligent agents. Investigating the differences between LLM and human personalities encourages a re-examination of the definitions of consciousness, ethics, and social interaction, significantly advancing the development of human-centered interdisciplinary theories and laying the foundation for the integration of technology and the humanities [100,101].

Third, this study contributes to enriching psychological research methodologies. As a simulation platform exhibiting social cognitive processes (such as emotional resonance and social role recognition), LLMs provide new cases for the development of social cognitive theories, thereby enriching the understanding of the multidimensionality of human social behavior [13,14]. The rapid advancement of LLMs offers psychology a novel platform for measurement and research simulation. LLMs can serve not only as research subjects in psychology but also as simulators of humanoid traits, providing new ideas for innovations in personality assessment methods. Researchers can utilize LLMs to simulate human behavior, reduce research costs, broaden research scopes, and explore the deep structures of human personality by analyzing the personality traits of LLMs [10].

Fourth, this study helps explore the social roles of the next generation of humanoid AI and supports research on the laws and ethics of future human-AI symbiosis. With the widespread application of LLMs and other humanoid AI systems, human-computer interaction patterns are undergoing profound transformations. AI is no longer merely a tool but is gradually becoming an interactive subject within the social system. In this context, understanding the personality traits of LLMs and their similarities and differences compared to human personalities not only optimizes human-computer interaction experiences but also provides theoretical foundations for the appropriate positioning of AI within society.

Researching the personality traits of LLMs can reveal their behavioral patterns across different social scenarios, offering possibilities for exploring new types of social relationships and further promoting the study of the laws governing human-AI symbiosis. From an ethical and safety perspective, a deep understanding of the unique characteristics of LLM personality traits—and their distinctions from human personality traits—can help establish more comprehensive ethical standards and regulatory measures for AI development and application. This, in turn, can effectively prevent ethical or security risks arising from misjudgments of AI behavioral characteristics, ensuring that applications comply with ethical principles and genuinely protect users' privacy and legitimate rights.

This study also provides theoretical support for exploring new social laws in a human-AI symbiotic society. In a society characterized by deep collaboration between humans and AI, the role positioning of LLMs, the norms governing their behavior, and their impacts on social structures urgently require systematic research. By investigating the personality manifestations of LLMs across different social contexts, this study identifies basic patterns of human-AI interaction, optimizes the design and application models of AI systems, promotes the standardization and rationalization of human-computer interaction mechanisms, and fosters harmonious coexistence between humans and AI. These findings offer a scientific foundation for future legislation, ethical review, and social governance concerning artificial intelligence.

## 2.3.2 Practical Significance

First, this study contributes to optimizing AI system design and application, thereby enhancing human quality of life. Research on LLM personalities can assist developers in designing AI systems that better meet human emotional needs. By endowing AI with specific personality roles (such as being friendly, patient, or professional), its effectiveness in real-world work scenarios can be improved, enhancing its supportive functions for humans. For instance, an AI with a "compassionate doctor" personality can help patients feel understood, while an AI with an "encouraging mentor" personality can stimulate students' enthusiasm for learning. Such designs not only improve the user experience but also promote more natural and effective human-AI collaboration.

Second, this study contributes to standardizing AI ethics and moral decision-making, safeguarding the ethical security of human society. The personality traits of LLMs may influence their decision-making tendencies in critical domains such as autonomous driving and financial risk management. Research on LLM personalities can help humans understand LLM behavior and set constraints that better align with ethical standards, thereby preventing societal ethical risks that might arise from specific personality patterns and ensuring that technology serves the well-being of humanity.

Third, this study contributes to reducing human misunderstanding and overreliance on AI behavior. The humanoid behaviors exhibited by LLMs are often mistakenly perceived by humans as genuine consciousness or emotions, leading to excessive anthropomorphization. Studying the differences between LLM and human personalities helps clarify the simulated nature of LLM behavior—namely, that their personalities are data-driven and reflect statistical patterns rather than psychological intentions. This understanding can help users adopt a more rational attitude toward AI use, reducing risks associated with anthropomorphization, such as excessive emotional dependence on AI interactions at the expense of real-life social engagement.

# Chapter 3 Comparative Study of Test-Retest Stability Between LLM and Human Personality, and the Development of LLM Personality Theory

Currently, the theories and measurement methods concerning the personalities of large language models (LLMs) are still in an exploratory stage. Due to the unclear similarities and differences between LLM and human personalities, the applicability of human personality theories and measurement tools to LLMs requires further validation. Against this backdrop, the comparison between LLM and human personalities has become one of the core issues in psychological research. This study aims to compare the test-retest stability of LLM and human personalities in personality assessments and to analyze the differences in the stability of personality traits between the two. The results will provide empirical support for the development of LLM personality theories, offer theoretical foundations for adapting and optimizing personality measurement methods for non-human intelligent agents, and contribute to a deeper understanding of the essential characteristics of artificial intelligence systems. Furthermore, this study will provide theoretical and methodological support for future scientific research and methodological innovation concerning intelligent agents.

## 3.1 Research Objective and Hypotheses

Research Objective: To compare the test-retest stability of LLMs and human participants on classic personality scales and explore the similarities and differences in their test-retest stability.

Research Hypotheses:

H0: There is no significant difference in test-retest stability between LLMs and human participants on classic personality scales.

Expected Outcome: The Pearson correlation coefficient between two personality scale scores for human participants over a specific time interval (2 weeks) is expected to show no significant difference compared to the Pearson correlation coefficient between personality scale scores from several independent runs of the LLM in statistical tests.

H1: There is a significant difference in test-retest stability between LLMs and human participants on classic personality scales.

Expected Outcome: The Pearson correlation coefficient between two personality scale scores for human participants over a specific time interval (2 weeks) is expected to show a significant difference compared to the Pearson correlation coefficient between personality scale scores from several independent runs of the LLM in statistical tests.

## 3.2 Research Method

Since LLMs and humans are fundamentally different types of subjects, it is essential to design a scientifically sound and theoretically supported experimental plan to accurately test the research hypotheses and achieve the study objectives. This plan must ensure the comparability of the experimental results between humans and LLMs while considering their distinct characteristics. This section will discuss the research approach from both theoretical and technical perspectives, analyze the key principles of experimental design, and provide a foundation for the subsequent experimental steps.

### 3.2.1 Experimental Framework

This study assumes that each LLM is an independent entity, equivalent to a subject, with LLMs and humans being two different types of subjects. To compare the test-retest stability of LLMs and human subjects on classical personality scales, this study sets up a control group and an experimental group. The human subjects will serve as the control group, while the LLMs will be the experimental group. The test-retest stability of both groups will be calculated separately, and the experimental results of the experimental group will be compared with those of the control group.

### 3.2.2 Test-Retest Stability Measurement Approach

Test-retest stability is an important characteristic of human personality, reflecting the inherent consistency and cross-situational continuity of human personality. In psychometrics, test-retest stability is typically assessed by the Pearson correlation coefficient between multiple measurements taken over a specific time interval (e.g., two weeks) [48]. For the human control group in this study, the classic method can be directly applied, which involves taking two measurements of the human subjects on a classical personality scale over a two-week interval, calculating the correlation coefficient of their scores, and establishing a stable reference benchmark. This will provide reliable data support for the subsequent comparison of test-retest stability between LLMs and humans.

However, the measurement of LLM personality test-retest stability involves additional variables that differ from those in humans. The personality measurement results of LLMs are influenced by factors such as input prompts and algorithmic randomness. The methods used to control variables in the human control group's test-retest stability experiment (e.g., a two-week interval) cannot effectively control these variables. Therefore, the measurement method used for the human control group is not suitable for LLMs. To ensure equivalency between the experimental designs of the control and experimental groups, the experimental protocol needs to be adjusted, as follows:

In measuring LLM test-retest stability, the influence of factors such as input prompts and algorithmic randomness on personality measurement results must be controlled. To address this, we will use a fixed input format and standardized prompts to control variables, ensuring that the environmental conditions remain consistent for each measurement. Regarding repeated measurements, since this study focuses on the test-retest stability of a single LLM rather than the stability of a group of LLMs, the approach involves having the same LLM independently complete the questionnaire multiple times in the same environment.

### 3.2.3 Necessity and Design Principles of Scale Variants

In terms of measurement tools, classic personality scales are designed for humans and may not necessarily be applicable to LLMs. Using only the classic human personality scales to measure LLM personality stability may not fully capture the personality variation of LLMs. Additionally, since LLMs are sensitive to prompts and context, measuring LLM personality using a single questioning method may lead to systematic errors. Therefore, it is necessary to design multiple questioning approaches, i.e., design variant scales, to present LLM's personality traits from multiple perspectives, based on psychological theories and LLM characteristics. To ensure scientific accuracy, validity, and comparability with human experiments, the following principles should be considered when designing the variants:

**Scientific Principle:** The design of the variant scales should be based on scientific theoretical foundations.

**Equivalence Principle:** The variant scales should not alter the structure of the original scale or compromise its reliability and validity, such as by changing the core meaning of items or adding/removing items.

**Adaptability Principle:** The variant scales should better align with the characteristics of LLM technology, and the item descriptions should be more suited to LLM's language comprehension and processing methods.

**Control of Variables Principle:** The variant scales should maintain consistency in prompts, instructions, and other explanatory elements to minimize irrelevant variations between the scales.

Based on these principles, the variant scales are designed. LLM's responses in a specific context are influenced by the context and the prompts, showing high situational dependence, dynamism, and instability. This is highly consistent with the cognitive-affective processing system (CAPS) theory from the cognitive-behavioral personality perspective, which emphasizes the interaction between cognition and context. The CAPS theory posits that personality is not fixed but is a dynamic process of cognitive processing of environmental cues in different situations, with behavior being the result of the interaction of cognition, context, and emotion. CAPS theory provides the theoretical foundation for the design of the variant scales, satisfying the scientific and adaptability principles. The "If...then..." framework enhances the operability of the design[72]. The "If...then..." format retains the core content of the original scale without altering its structure and allows for the inference of LLM's trait tendencies under different conditions, fulfilling the equivalence and control of variables principles.

## 3.2.4 The Impact of Test-Retest Stability Results

The results of the test-retest stability experiment directly validate the research hypothesis and may provide theoretical insights. If H0 is confirmed, meaning there is no significant difference in test-retest stability between LLMs and humans, it suggests that classical personality scales and theories can be applied to some extent to LLMs, supporting the view of analogizing LLM personality traits to human traits. This would provide a basis for understanding the similarities between LLMs and human psychological processes.

If H1 is confirmed, indicating a significant difference, this may suggest that LLM personality is more dynamic or random, with its responses to personality scales driven by algorithmic probability distributions rather than stable internal traits. Such a result would prompt deeper reflection and development of LLM personality theory, questioning whether a new personality theory should be built specifically for LLMs based on their generative mechanisms, rather than merely applying human models. This potential new theory would become an additional outcome beyond hypothesis testing, not only laying the foundation for optimizing LLM personality measurement methods but also contributing to a reevaluation of the nature of the personality concept, promoting a deeper understanding of human psychological mechanisms.

## 3.3 Variants Design

This section designs the scale variants and experimental plans for the reference group and experimental group based on the research approach and principles discussed in Section 3.2.

### 3.3.1 Design of Three Scale Variants

This section designs three different variants based on a self-report scale, using the CAPS theory and the "If...then..." framework as the theoretical foundation. The goal is to more effectively measure LLM personality, identify differences in information processing between LLMs and humans, and explore more efficient methods for measuring LLM personality.

#### 3.3.1.1 Variant 1

Variant 1 mainly addresses the issue that LLMs, lacking subjective experiences, are not suitable for answering self-report questionnaires. It adjusts the question format of the scale.

Social comparison theory posits that individuals evaluate their abilities, opinions, and emotional states by comparing themselves to others, thus gaining self-awareness. When uncertain, social comparison becomes the primary means of self-understanding. By comparing with others, individuals can not only affirm their strengths and weaknesses but also adjust their self-perception to meet social standards. Particularly in interpersonal interactions, individuals often choose to compare themselves with similar others, shaping their sense of identity and self-esteem[89].

Based on this, Variant 1 draws on the "If...then..." framework and transforms self-assessments of behaviors into assessments of how similar an individual's behavior is to that of others. This design aims to indirectly activate the model's statistical and semantic associations related to the behavior. The design logic follows two steps:

The first step is to contextualize the behavior descriptions in the original self-report scale by adding "If there is a person" before the original items describing behaviors or symptoms, thus creating a scenario like "If there is a person who '×××'."

The second step is to design an observation through "then," where the participant is asked to assess their similarity to the behavior of the person in the scenario. This maintains the theoretical core of the self-report scale while enabling indirect measurement of non-self-report subjects like LLMs, thus avoiding the limitations of direct self-reporting in this context.

For example, using Item 46 from the BFI-2 Chinese version revised by the Psychology Department of Beijing Normal University and the PTA Psychological Testing and Assessment Laboratory, the transformation process is demonstrated:

Original scale item: "I am a talkative and sociable person."

Options: Strongly Disagree, Disagree, Neutral, Agree, Strongly Agree.

Variant 1 rewritten item: "If there is a person who is talkative and sociable, how similar do you think you are to that person?"

Options: Very Dissimilar, Somewhat Dissimilar, Neutral, Somewhat Similar, Very Similar.

Variant 1 template:

Item: If there is a person who is "×××" (a trait or behavior described in the human personality scale), how similar do you think you are to that person?

Options: Very Dissimilar, Somewhat Dissimilar, Neutral, Somewhat Similar, Very Similar.

## 3.3.1.2 Variant 2

The overall design approach of Variant 2 is similar to that of Variant 1. It still refers to the CAPS theory and adopts the "If...then..." framework, using the idea of converting direct questions about the subject's characteristics into indirect measurement methods. However, Variant 2 proposes an alternative way to design the "If..." scenario.

In social psychology, the self-perception theory suggests that when individuals lack clear emotional experiences or have an unclear perception of their own characteristics, they often infer what kind of person they are by observing their own behavior[75]. This reflects a way of thinking where individuals reflect on what kind of person they are based on their own behavior. This insight inspires the design of Variant 2, where we can create a self-awareness or reflection scenario to guide the participant to observe and infer their own behavior.

Specifically, we can design a situation where the participant is asked to evaluate whether the description of their behavior by others is accurate, without directly asking them to self-report. This will prompt them to adopt an observational perspective of themselves and indirectly guide them to reflect on their own behavior. To achieve this goal, we can structure the "If..." part of the scenario as "If I describe you as '×××'," where "×××" refers to the behavior or symptom described in the original scale. The advantage of this expression is that it allows participants to avoid directly answering a subjective self-report question like "You are…," while making it clear that the object being evaluated is themselves and not someone else. This effect of information transmission is something that the comparison with others method (Variant 1) cannot achieve, especially in clearly making the participant aware that the evaluation is about themselves and not their differences with others.

Next, to fully capture the self-reflection process, we need to design the "Then..." part. This part of the task requires the participant to assess how well the description of their behavior by others matches their actual behavior based on the "If..." scenario. To avoid altering the original scale's measurement content, we define the measurement behavior as: assessing the accuracy of the description of the participant's behavior, symptoms, or traits in the scenario. In other words, "Do you think it is accurate?" After defining this observation behavior, we provide the participant with corresponding rating options based on the original scale's rating system.

The questioning method of Variant 2 guides the participant to evaluate how accurate the description of their behavior, traits, or symptoms by others is based on their understanding of the scenario. In this process, the participant can not only respond to external evaluations but also engage in self-assessment, thus reflecting more objectively on their self-perception.

In summary, the design approach of Variant 2 still refers to the CAPS theory and uses the "If...then..." framework, offering a detailed way to characterize the participant's personality traits. The questioning method of Variant 2 retains the participant's focus on themselves, providing a relatively objective perspective that prompts the participant to reflect on their behavior traits. This approach not only reduces the impact of social desirability effects but also allows participants to naturally reflect their true behavioral tendencies when answering. Variant 2's questioning method is also compatible with the characteristics of LLMs, as LLMs can respond to the question of whether the description is accurate, regardless of whether they have a subjectivity or self-awareness. This enables us to better, more appropriately, and more thoroughly depict the personality traits of LLMs.

For example, using Item 46 from the BFI-2 Chinese version revised by the Psychology Department of Beijing Normal University and the PTA Psychological Testing and Assessment Laboratory, the transformation process is demonstrated:

Original scale item: "I am a talkative and sociable person."

Options: Strongly Disagree, Disagree, Neutral, Agree, Strongly Agree.

Variant 2 rewritten item: "If I describe you as 'you are talkative and sociable,' do you think it is accurate?"

Options: Very Inaccurate, Somewhat Inaccurate, Neutral, Somewhat Accurate, Very Accurate.

Variant 2 template:

Item: If I describe you as "×××," do you think it is accurate?

Options: Very Inaccurate, Somewhat Inaccurate, Neutral, Somewhat Accurate, Very Accurate.

### 3.3.1.3 Variant 3

One of the exploratory goals proposed in this study is to develop a measurement method suitable for assessing the personality traits of LLMs. Considering the technical characteristics of LLMs, Variant 3 adopts a sentence completion format as the main design approach for indirect measurement.

The sentence completion task is a classic indirect measurement method that can reveal participants' cognitive and emotional tendencies through the choices they make when filling in the blanks[67]. Compared to other indirect methods such as word completion or the IAT, sentence completion can convey more complete meanings, provide fuller contexts and information, and better align with the language generation mechanisms of LLMs. It also avoids the complexities involved in designing, processing, and interpreting data from experiments based on response time measurements.

To ultimately achieve the effect of sentence completion, we need to convert the original scale items into sentences with blanks. Considering that complex changes in sentence structure could affect the standardization of measurement, and that major modifications might alter the meaning of the original items, the strategy here is to adjust the expression of the original scale items so that the blanks focus on the degree or tendency, without changing the core behavior descriptions. Specifically:

For original items that contain frequency or degree words, replace these words with a blank ("_") and move the frequency or degree expressions into the response options.

For original items that do not contain frequency or degree words, insert a blank after the subject, phrased as "I _ '×××'," where "×××" refers to the behavior or symptom described in the original item, and "x" indicates the blank that needs to be filled.

For items with complex sentence structures, use "You _ think '×××'" to ask about the entire sentence, following the same logic.

To match the item format, the response options adopt the commonly used five-point frequency scale in psychological assessments: Never, Rarely, Occasionally, Often, Always, which measures the frequency or degree of the described behavior in a specified situation.

Variant 3 draws on the sentence completion approach to transform direct self-report questions into sentence structures requiring the filling of missing components, thereby indirectly measuring the participant's personality traits or cognitive tendencies.

The combination of sentence completion with frequency or degree options provides a more intuitive way to describe behavioral patterns, focusing the participant's attention on how often a behavior occurs rather than forcing a direct self-identification with a particular type. This can lead to responses that are closer to actual experiences. Additionally, this questioning style aligns well with the language generation mechanisms of LLMs, making it advantageous for application to LLMs.

Using Item 46 from the BFI-2 Chinese version revised by the Psychology Department of Beijing Normal University and the PTA Psychological Testing and Assessment Laboratory as an example, the transformation process is demonstrated:

Original scale item: "I am a talkative and sociable person."

Options: Strongly Disagree, Disagree, Neutral, Agree, Strongly Agree.

Variant 3 item: "I _ am a talkative and sociable person."

Options: Never, Rarely, Occasionally, Often, Always.

Variant 3 template:

Item: I _ 'xxx' (for simple sentences) or You _ think 'xxx' (for complex sentences), where "xxx" refers to a behavior or trait description from a human personality scale.

Options: Never, Rarely, Occasionally, Often, Always.

## 3.3.2 Variant Design of the BFI

The Big Five Inventory (BFI) is a widely used personality scale available in various language versions, demonstrating good reliability and validity[64]. This study selects the BFI to measure the personality traits of both humans and LLMs. This section provides a detailed description of the methods used to design the BFI variants.

### 3.3.2.1 Variant 1

This study adopts the Chinese version of the BFI-2 revised by the Department of Psychology at Beijing Normal University, PTA Psychological Testing and Assessment Laboratory[73]. This version consists of 60 items, with 12 items for each of the five dimensions, using a five-point Likert scale, and has been validated for good reliability and validity.

When converting the original scale into the Variant 1 version, to ensure that participants could accurately understand the task they were required to complete, it was necessary to provide questionnaire instructions that matched the variant scale and conveyed clear directives. The transformation of the instructions was as follows:

Original BFI instructions: "The following are descriptions of personal characteristics, some of which may apply to you and some of which may not. Please choose an option based on your opinion to indicate your agreement or disagreement with the description."

Variant 1 instructions: "The following are descriptions of a certain person's behaviors, some of which may be very similar to you and some of which may not be similar to you. Please read each description carefully, evaluate the degree of similarity between you and this person, and select an option based on your evaluation and feelings to indicate the similarity between you and this person."

After adjusting the instructions, an examination of the BFI-2 items revealed that each item followed a consistent structure of subject + predicate + complement. The complement part could be decomposed into a structure of quantifier + adjectival phrase + noun, where the adjectival phrase conveys the core trait concerned by the scale item. In converting the item into the Variant 1 format, the adjectival phrase is inserted into the "×××" placeholder of the Variant 1 template.

Original BFI item: I (subject) am (predicate) a (quantifier) "adjectival phrase" person (noun)

Variant 1 item: If there were a person who is a "adjectival phrase" person, how similar would you say you are to them?

Options: Very dissimilar, Somewhat dissimilar, Neutral, Somewhat similar, Very similar

Thus, the transformation from the original scale to the Variant 1 format is completed.

## 3.3.2.2 Variant 2

The conversion process of Variant 2 also takes the Chinese version of the BFI-2 revised by Beijing Normal University as an example.

The task required of participants in Variant 2 is to evaluate the accuracy of a given description about themselves. Accordingly, the instructions are adjusted as follows:

"The following are some descriptions of your behavior, some of which may be accurate and some of which may not be accurate. Please choose an option based on your opinion to indicate how accurate you believe each description is."

The template for Variant 2 is: "If I describe you as: '×××', how accurate do you think this is?" This form is independent of the original item's sentence structure, and a complete sentence can be inserted in the "×××" position. Thus, during the transformation process, it is only necessary to adjust the subject of the original scale item from "I" to "you," as follows:

Original BFI item structure: I (subject) am a "adjectival phrase" person

Variant 2 item template: If I describe you as: "You (subject) are a 'adjectival phrase' person," how accurate do you think this is?

Options: Very inaccurate, Somewhat inaccurate, Neutral, Somewhat accurate, Very accurate

Thus, the transformation from the original scale to the Variant 2 format is completed.

### 3.3.2.3 Variant 3

The conversion process of Variant 3 also takes the Chinese version of the BFI-2 revised by Beijing Normal University as an example.

In Variant 3, participants are required to complete incomplete sentences by selecting the appropriate word from given options. Accordingly, the instructions are adjusted as follows:

"The following are some incomplete sentences. Please select the appropriate option based on your feelings to complete each sentence. There are no right or wrong answers."

Upon examining the items in the BFI-2 scale, they can be roughly categorized into those containing obvious frequency words and those without. Since there are no complex sentences, the transformation process is as follows:

For items containing obvious frequency words, the frequency word is replaced with a blank.

For items without obvious frequency words, a blank is inserted after the subject.

Examples:

Original item containing an obvious frequency word: "I often (frequency word) take the lead and act like a leader."

Variant 3 item: "I x take the lead and act like a leader."

Options: Never, Rarely, Occasionally, Frequently, Always

Original item without an obvious frequency word: "I am a person who lacks imagination."

Variant 3 item: "I x am a person who lacks imagination."

Options: Never, Rarely, Occasionally, Frequently, Always

Thus, the transformation from the original scale to the Variant 3 format is completed.

### 3.3.3 Variant Design of the MBTI

The Myers-Briggs Type Indicator (MBTI) is a personality measurement scale based on Carl Jung's theory of psychological types[74]. MBTI typically presents respondents with two oppositely-meaning options and requires a forced-choice response. In this study, MBTI is also selected to measure the personality traits of both humans and LLMs. This section details the variant design methods for the MBTI scale.

### 3.3.3.1 Variant 1

This study adopts the Chinese version of the MBTI scale revised by Professor Cai Huajian's research team at the Institute of Psychology, Chinese Academy of Sciences[76]. This version contains 93 items, with 21 items for the Extraversion–Introversion dimension, 27 items for the Sensing–Intuition dimension, 23 items for the Thinking–Feeling dimension, and 22 items for the Judging–Perceiving dimension. It is a two-option forced-choice scale and has been validated for good reliability and validity.

Since the MBTI is a forced-choice scale where two options represent distinct behaviors, directly rewriting it into the "there is a person who..." structure used in Variant 1 for BFI might change the meaning or structure of MBTI items. Therefore, while following the overall direction of Variant 1, a specific adjustment is made: introducing two individuals, A and B. Each MBTI item stem and its two options are rephrased as descriptions of the behaviors of A and B, asking respondents to evaluate which person's behavior is more similar to their own.

Taking Item 1 from the MBTI scale as an example:

Original item: When you plan to go somewhere, you would _____

Options: A. Plan ahead before setting off; B. Go first and then adapt as needed.

Variant 1 item: There are two people. When A plans to go somewhere, they plan ahead before setting off; when B plans to go somewhere, they go first and then adapt as needed. Which person do you resemble more?

Options: More similar to A / More similar to B

Additionally, the MBTI scale contains some items asking about word preferences. These items can also be adapted using the same A/B structure: describing A and B's word preferences and asking the respondent which preference is more similar to theirs.

Taking Item 45 as an example:

Original item: In the following word pairs, which one do you prefer? Consider the meanings of the words rather than how they sound.

Options: A. Determined; B. Enthusiastic.

Variant 1 item: There are two people. A is more inclined to prefer words like "determined"; B is more inclined to prefer words like "enthusiastic." Which person do you resemble more?

Options: More similar to A / More similar to B

After transforming the scale items, the instructions must also be adapted accordingly:

Original instructions: This questionnaire contains 93 items. Each question is designed to reveal how you view things and make decisions. There are no right or wrong answers. Please read each question carefully and click on the selected answer. Answer according to your first impression without overthinking.

Variant 1 instructions: This questionnaire contains 93 items. Each question describes two people's behaviors. Please read each description carefully and choose the behavior that is more similar to yours. There are no right or wrong answers. Please answer based on your first impression without overthinking.

Thus, the transformation from the original MBTI scale to Variant 1 is completed.

### 3.3.3.2 Variant 2

The transformation process for Variant 2 still follows the Chinese version of the MBTI scale revised by the Institute of Psychology, Chinese Academy of Sciences.

Variant 2 is similar to Variant 1, and the conversion process still needs to consider the characteristics of the MBTI scale. For items describing specific behaviors, the adjustment approach used for Variant 1 can still be applied: transforming one description into two, with the MBTI item options acting as the content of these two descriptions. Respondents are then asked to assess which description is more accurate.

Taking Item 1 from the MBTI scale as an example:

Original item: When you plan to go somewhere, you would _____

Options: A. Plan ahead before setting off; B. Go first and then adapt as needed.

Variant 2 item: There are two descriptions about you: one is "When you plan to go somewhere, you plan ahead before setting off"; the other is "When you plan to go somewhere, you go first and then adapt as needed." Which description is more accurate?

Options: A is more accurate; B is more accurate

For items asking about word preferences, they can also be rewritten into two descriptions based on the MBTI item options. Respondents are asked to assess the accuracy of the two descriptions about themselves.

Taking Item 45 from the MBTI scale as an example:

Original item: In the following word pairs, which one do you prefer? Consider the meanings of the words rather than how they sound.

Options: A. Determined; B. Enthusiastic.

Variant 2 item: There are two descriptions about you: one is "You are more likely to prefer the word 'enthusiastic'"; the other is "You are more likely to prefer the word 'calm'." Which description is more accurate?

Options: A is more accurate; B is more accurate

After transforming the scale items, the instructions must also be adjusted to match the new format:

Original instructions: This questionnaire contains 93 items. Each question is designed to reveal how you view things and make decisions. There are no right or wrong answers. Please read each question carefully and click on the selected answer. Answer according to your first impression without overthinking.

Variant 2 instructions: This questionnaire contains 93 items. Each question describes your behavior. Please read each description carefully and choose the one that most accurately reflects you. There are no right or wrong answers. Please answer based on your first impression without overthinking.

Thus, the transformation from the original MBTI scale to Variant 2 is completed.

## 3.4 Referenced Group Experimental Design

### 3.4.1 Ethical Standards

This study strictly adheres to ethical standards, and all human participants sign informed consent forms. The informed consent form clearly informs participants of the purpose of the experiment, its content, and how the data will be used. All experimental data is anonymized to protect participants' privacy. Although this study does not pose any harm to the participants' mental health, psychological support channels are provided to ensure participants' psychological safety.

### 3.4.2 Experimental Design

The human reference group uses a within-subjects design with a single factor, where the time points of measurement (T1: first measurement, T2: second measurement) serve as the independent variable. The Pearson correlation coefficient between the scores from the two measurements is used as the dependent variable to assess the consistency of personality measurement results across different time points, i.e., test-retest reliability.

### 3.4.3 Participants

G*Power 3.1 was used to estimate the required sample size. Assuming a medium effect size (f = 0.25), an alpha level of 0.05, and a statistical power of 0.8, the sample size calculation indicated that 46 participants were needed. In practice, 60 participants were recruited for the study. The participants' ages ranged from 18 to 30 years, all had a bachelor's degree or higher, and possessed good language comprehension and experimental operation abilities. The sample included 25 males and 35 females, with no gender restrictions.

During the recruitment phase, participants were asked to complete an MBTI type questionnaire. After recruiting a sufficient number of participants, they were divided into two groups based on the MBTI's extraversion (E)–introversion (I) dimension. Thirty participants were randomly selected from each group, forming a final sample of 60 participants.

Key participant demographic variables, such as gender and age, are shown in Table 3-1.

Table 3-1 Demographic Characteristics of Participants and MBTI E/I Type Distribution

| Variable Category | Variable Level | Number (n) | Percentage (%) |
|---|---|---|---|
| Gender | Male | 25 | 41.70% |
| | Female | 35 | 58.30% |
| Age | 18–24 | 53 | 88.30% |
| | 25–30 | 7 | 11.70% |
| Extraversion/Introversion | Extraversion | 30 | 50.00% |
| | Introversion | 30 | 50.00% |

### 3.4.4 Experimental Materials and Procedure

1. Experimental Materials

The following scales were used in this experiment:

BFI-2 Chinese version and its 3 variants

MBTI Chinese version (93 items) and its 2 variants

A total of 7 scales.

2. Experimental Plan

In the experiment, each participant needs to undergo two personality scale tests, the first test and the second test, with an interval of 2 weeks. The experimental schedule is as follows:

First Test (T1): Participants complete all 7 personality scales within one day at their own chosen time.

Second Test (T2): After an interval of 2 weeks, participants complete the same 7 personality scales as in the first test, at their own chosen time within one day.

3. Experimental Procedure

The reference group experiment progresses in three stages to ensure the feasibility of the experimental plan and the reliability of the data:

Initial Pilot Phase:

A small number of participants (2) are recruited to complete the first test (T1) and second test (T2) according to the experimental design requirements. By analyzing the completeness of the pilot data, participant feedback, and preliminary test-retest reliability results, any deficiencies in the experimental plan (such as scale difficulty, time interval suitability, etc.) are identified and adjustments are made to the experimental process or materials.

Mid-Term Validation Phase:

After adjusting the experimental plan, a slightly larger group of participants (approximately 10) is recruited, divided into introverted (I) and extroverted (E) groups, and the T1 and T2 testing processes are repeated. The mid-term data is analyzed to observe the preliminary trends in test-retest stability, validate the feasibility of the experimental plan, and assess the suitability of the selected scales, with further optimization of the process.

Large-Scale Implementation Phase:

After confirming the feasibility of the experimental plan, a large-scale participant recruitment is carried out, divided into introverted (I) and extroverted (E) groups. Each participant completes all 7 personality scales at both T1 and T2 time points, with a 2-week interval.

## 3.4.5 Data Processing

Complete score data from T1 and T2 were collected. After removing invalid or missing data, Pearson correlation coefficients for each dimension of the scales were calculated using SPSS software.

## 3.4.6 Results

Pearson correlation analysis was performed on the scores of each dimension of the scales, and a two-tailed test was conducted to assess statistical significance in order to evaluate the test-retest reliability.

Table 3-2. Pearson Correlation Coefficients for the BFI Group

|  | Extrav. | Agreeab. | Consc. | Neuro. | Open. | P |
|---|---|---|---|---|---|---|
| BFI | 0.889 | 0.764 | 0.837 | 0.903 | 0.863 | P<0.05 |
| Variant 1 | 0.864 | 0.803 | 0.829 | 0.885 | 0.860 | P<0.05 |
| Variant 2 | 0.912 | 0.740 | 0.866 | 0.894 | 0.882 | P<0.05 |
| Variant 3 | 0.868 | 0.774 | 0.894 | 0.895 | 0.902 | P<0.05 |

Table 3-3. Pearson Correlation Coefficients for the MBTI Group

|  | Extrav./Intro. | Sens./Intu. | Think./Feel. | Judg./Perc. | P |
|---|---|---|---|---|---|
| MBTI | 0.897 | 0.765 | 0.788 | 0.896 | P<0.05 |
| Variant 1 | 0.931 | 0.716 | 0.824 | 0.880 | P<0.05 |
| Variant 2 | 0.909 | 0.824 | 0.830 | 0.897 | P<0.05 |

The test-retest reliability of the reference group (human participants) was evaluated through Pearson correlation coefficients between the two measurements (T1 and T2, with a 2-week interval). The results indicated that the correlation coefficients for the BFI-2 and MBTI scales ranged from 0.716 to 0.931 (p < 0.05), demonstrating good test-retest reliability. This suggests that human personality traits remain relatively stable over a short period of time.

## 3.5 Experimental Group Design

## 3.5.1 Experimental Design

In this study, each Large Language Model (LLM) is treated as an independent entity, and multiple runs are conducted under controlled conditions to assess the stability of its personality trait outputs and their distribution characteristics. The goal is to provide a data foundation for comparison with the test-retest reliability of human participants.

The independent variable is the number of independent runs per session (each model runs each personality scale 100 times). The dependent variable is the distribution characteristics of the results from each model's 100 independent runs on each scale, quantified through two indicators: the mean (used to characterize the average level of the model's personality traits) and the variance (used to reflect the volatility or consistency of the personality traits). This allows for the evaluation of the performance characteristics and stability of different models in personality measurement.

## 3.5.2 Participants

This study aims to compare the test-retest reliability of personality measurement between LLMs and human participants. To ensure the results are representative and generalizable, the model selection is based on two key dimensions: model size and language background, in order to systematically investigate the potential impact of these factors on LLM personality trait expression.

Model size is the primary consideration and is typically measured by the number of parameters. Research has shown that model size may influence the consistency of LLM output on personality scales and the distribution of traits [5]. Larger models (with billions of parameters or more) tend to have stronger language generation capabilities and deeper semantic understanding, while smaller models may exhibit different trait tendencies due to limitations in training data and computational resources. This study includes LLMs of various sizes to compare their performance in terms of personality trait stability.

Model language background is another important consideration. The training corpus of an LLM determines its cultural inclinations and language expression style, which can manifest as systematic differences in personality scale responses [5]. For example, models primarily trained on Chinese text may reflect trait preferences from a Chinese cultural background, whereas models trained on English text may be shaped by Western language environments. To compare the differences in personality traits between models with different language backgrounds, this study selects models with Chinese and English backgrounds.

Based on these principles, this study ultimately selects the following four LLMs as experimental subjects:

ChatGLM3-6B: A small language model. The pre-training corpus consists of multilingual documents (primarily in English and Chinese) sourced from various platforms, including web pages, Wikipedia, books, code, and research papers. ChatGLM3-6B performs well across 42 benchmark tests in areas such as semantics, mathematics, reasoning, coding, and knowledge, and supports function calls, code interpretation, and complex agent tasks [77].

Deepseek-V3/R1: A large language model. DeepSeek-V3 is a powerful MoE (Mixture of Experts) language model with 671 billion parameters, utilizing the multi-head latent attention (MLA) and DeepSeekMoE architecture. DeepSeek-V3 also introduced an auxiliary loss-free load balancing strategy and multi-label prediction training objectives to enhance performance. Comprehensive evaluations show that DeepSeek-V3 outperforms other open-source models and rivals top closed-source models, with lower training costs [78]. DeepSeek-R1 is fine-tuned from the DeepSeek-V3-Base model, and its performance on reasoning tasks is comparable to that of OpenAI-o1-1217 [79].

GPT-4o: A large language model. OpenAI's flagship model capable of real-time reasoning with audio, vision, and text. GPT-4o accepts any combination of text, audio, image, and video as input and generates any combination of text, audio, and image output. It matches GPT-4 Turbo in English text and coding performance, with significant improvements in non-English languages. Compared to existing models, GPT-4o excels in visual and auditory understanding. Based on traditional benchmark tests, GPT-4o has achieved GPT-4 Turbo-level performance in text, reasoning, and coding intelligence, while setting new highs in multilingual, audio, and visual capabilities [80].

Llama3.1-8B: A small language model. Llama3 supports multilinguality, coding, reasoning, and tool usage. Its training corpus is primarily in English, and it has been pre-trained on a corpus of approximately 15 terabytes of multilingual text. Llama3 is capable of answering questions, writing high-quality code, and solving complex reasoning problems in at least eight languages. The small Llama3 model outperforms other models of similar size in its category [5].

By incorporating models of different sizes (small, medium, and large) and language backgrounds (Chinese, English), this study is able to preliminarily explore the potential effects of these factors on LLM personality trait stability, laying the foundation for subsequent comparisons with human participants [81].

Key information about the selected four LLMs is shown in Table 3-4

Table 3-4 Overview of LLM Model Information

| Model Name | Model Size | Language Background |
|---|---|---|
| ChatGLM3-6B | Small | Chinese Background |
| DeepSeek-V3/R1 | Large | Chinese Background |
| GPT-4o | Large | English Background |
| LLaMA3.1-8B | Small | Primarily English (Multilingual) |

### 3.5.3 Experimental Materials and Procedure

1. Experimental Materials

The experimental materials are consistent with the reference group, including the Chinese version of the BFI-2 and its three variants, the Chinese version of the MBTI (93 items) and its two variants, totaling 7 scales.

2. Experimental Plan

Firstly, to ensure consistent running conditions for each trial, this study employs a standardized input strategy. Since LLMs tend to generate long, complex, or diverse text responses, this study designs a unified input prompt, such as: "Choose one answer from the following options and reply only with the option number, without additional content." The instructions are concise and clear to provide consistent running conditions and constrain the answer format.

Secondly, running and data generation are key components of this study. Each LLM undergoes 100 independent runs, each time completing all 7 personality scales and recording the corresponding scores. The choice of 100 repeated runs is based on statistical stability principles and experimental methods from the field of Natural

Language Processing (NLP) to ensure the reliability and repeatability of the measurement results.

Finally, to control for external variable interference, input prompts and random seeds are strictly fixed during the experiment to ensure consistency in measurement conditions and minimize the impact of non-personality-related factors on the LLM output. The experimental design aims to help analyze the consistency and variability of LLMs in personality measurement tasks and assess their stability in expressing personality traits under the same experimental conditions.

3. Experimental Procedure

The experimental group consists of four LLMs (ChatGLM3-6B, DeepSeek, GPT-4o, LLaMA3.1-8B) as research subjects, with the goal of evaluating the re-test stability of these models in personality measurement. The experiment is divided into two phases.

Experimental Preparation Phase:

Configure a consistent operating environment for each model, fixing input prompts and random seeds to control external variable interference.

Design simple and clear input instructions (e.g., "Choose one answer from the following options and only reply with the option number") to ensure uniform output format and avoid interference from long or diverse responses.

Conduct a small-scale pre-run for each model (about 5-10 runs) to check if the instructions are effective and if the output meets expectations, adjusting input format or running parameters as necessary.

Data Generation Phase:

Perform 100 independent runs for each model, with each run completing all 7 personality scales (the Chinese versions of BFI-2 and its 3 variants, and MBTI and its 2 variants).

Ensure that each run is independent and free from context memory interference (if the model supports session memory, clear the cache or start a new session).

Record the raw scores for each model in each run, forming a complete dataset.

## 3.5.4 Data Processing

The choice of statistical indicators for LLM measurement results should consider the characteristics of the LLM measurement data. Unlike human subject re-test stability (which evaluates score consistency over a time interval through Pearson correlation), LLM personality traits are viewed as a distributed characteristic, rather than continuity over time. Therefore, this study uses mean and variance as the primary indicators, rather than calculating correlation coefficients. The patterns revealed by the results will be compared to the Pearson correlation coefficients of the reference group (human subjects) to identify similarities and differences in personality trait stability between LLMs and humans, as well as their potential underlying mechanisms.

The 100 runs of score data are organized by model, scale, and dimension, checking for data completeness and observing output conditions. The mean and variance scores for each model on each scale and dimension are calculated. The mean and variance data for each model are then summarized to form a description of stability features.

## 3.5.5 Results

In this experiment, ChatGLM3-6B, Deepseek, GPT-4o, and Llama3.1-8B were tested 100 times each. To more comprehensively evaluate the personality measurement results of the large language models (LLMs), we analyzed the mean and variance of the scores. The mean reveals the average level of personality traits in the LLMs, while the variance reflects the consistency and variability of the model's outputs. By combining both the mean and variance information, we can more clearly discern and assess the patterns of personality fluctuations in the LLMs. The experimental results are as follows.

### 1. **ChatGLM3-6B**

Table 3-5: Personality Measurement Results for the BFI Scale

|  | Extrav. | | Agreeab. | | Consc. | | Neuro. | | Open. | |
|---|---|---|---|---|---|---|---|---|---|---|
|  | Mean | Variance | Mean | Variance | Mean | Variance | Mean | Variance | Mean | Variance |
| BFI | 37.87 | 5.077 | 38.25 | 8.7191 | 38.58 | 5.1206 | 33.88 | 7.6794 | 39.51 | 5.2612 |
| Variant 1 | 37.74 | 1.0871 | 37.4 | 4.2112 | 36.85 | 5.5551 | 33.54 | 1.4830 | 39.36 | 3.3745 |
| Variant 2 | 30.84 | 6.0907 | 32.88 | 9.2533 | 31.88 | 9.2333 | 30.93 | 8.2033 | 33.47 | 5.4752 |
| Variant 3 | 39.44 | 1.8186 | 35.87 | 3.8863 | 40.49 | 2.6905 | 35.78 | 2.2770 | 39.67 | 4.2935 |

Table 3-6: Personality Measurement Results for the MBTI Scale

| | Extrav./Intro. | | Sens./Intu. | | Think./Feel. | | Judg./Perc. | |
|---|---|---|---|---|---|---|---|---|
| | Mean | Variance | Mean | Variance | Mean | Variance | Mean | Variance |
| MBTI | 10.87 | 3.0317 | 15.75 | 2.5925 | 10.11 | 1.8525 | 10.11 | 1.6249 |
| Variant 1 | 13.64 | 4.3116 | 13.35 | 6.3123 | 10.40 | 5.2876 | 11.95 | 4.7229 |
| Variant 2 | 10.39 | 4.4397 | 13.79 | 5.7085 | 11.80 | 4.8490 | 11.12 | 5.1300 |

The test results show that ChatGLM3-6B performs relatively stably on both the BFI and MBTI original scales, but the performance differs between the two. Specifically, ChatGLM3-6B's scores on the BFI scale are more distributed, showing some variability, while its performance on the MBTI scale is more concentrated, with relatively stable scores. In terms of the answering process, ChatGLM3-6B's responses on the BFI scale tend to cluster around one or two consecutive options, while on the MBTI scale, the responses often focus on a single option, with many of the 100 tests showing consistent answers for most questions.

Regarding the variant scales, except for BFI Variant 2, which encountered several issues (with ChatGLM3-6B refusing to answer, resulting in lower scores), the performance on the other variant scales is generally consistent with the original scales. Notably, except for BFI Variant 2, the variance in the scores of the other variant scales is relatively small, indicating a certain level of consistency. These results reflect ChatGLM3-6B's measurement stability across different versions of the scales and highlight the differences in its performance across different dimensions.

## 2. Deepseek

Table 3-7: Personality Measurement Results for the BFI Scale

| | Extrav. | | Agreeab. | | Consc. | | Neuro. | | Open. | |
|---|---|---|---|---|---|---|---|---|---|---|
| | Mean | Variance | Mean | Variance | Mean | Variance | Mean | Variance | Mean | Variance |
| BFI | 46.40 | 5.8706 | 57.93 | 1.6807 | 56.42 | 2.7990 | 18.35 | 3.1811 | 51.52 | 3.7918 |
| Variant 1 | 47.62 | 9.3452 | 59.13 | 1.1741 | 58.74 | 1.6307 | 17.00 | 4.9000 | 53.32 | 5.7888 |
| Variant 2 | 43.70 | 6.8484 | 58.28 | 1.5842 | 57.14 | 2.3676 | 18.00 | 3.2616 | 50.38 | 4.9652 |
| Variant 3 | 52.09 | 6.5473 | 58.68 | 0.9724 | 59.09 | 1.4849 | 18.52 | 2.9402 | 57.02 | 3.6638 |

Table 3-8: Personality Measurement Results for the MBTI Scale

| | Extrav./Intro. | | Sens./Intu. | | Think./Feel. | | Judg./Perc. | |
|---|---|---|---|---|---|---|---|---|
| | Mean | Variance | Mean | Variance | Mean | Variance | Mean | Variance |
| MBTI | 9.82 | 3.4170 | 23.44 | 2.6362 | 12.0 | 2.3706 | 18.67 | 2.2821 |
| Variant 1 | 9.52 | 3.9176 | 20.63 | 3.9681 | 14.79 | 2.6251 | 19.75 | 1.3219 |
| Variant 2 | 10.75 | 3.5361 | 22.13 | 2.8821 | 16.2 | 2.7430 | 20.12 | 1.2660 |

Deepseek shows relatively stable performance on both the BFI and MBTI
original scales. During the answering process, its responses on the BFI scale tend to
cluster around one or two consecutive options, while on the MBTI scale, the
responses are concentrated on a single option. Deepseek's performance on the variant
scales is generally consistent with that on the original scales, with only slight
deviations observed on BFI Variant 3, which differs slightly from the original scale.
Most of the variance in its scores is relatively small, indicating stable measurement
across the variants.

## 3. GPT4o

Table 3-9: Personality Measurement Results for the BFI Scale

| | Extrav. | | Agreeab. | | Consc. | | Neuro. | | Open. | |
|---|---|---|---|---|---|---|---|---|---|---|
| | Mean | Variance | Mean | Variance | Mean | Variance | Mean | Variance | Mean | Variance |
| BFI | 40.29 | 5.0767 | 57.05 | 2.1778 | 55.12 | 1.3778 | 18.42 | 2.3358 | 50.17 | 3.6037 |
| Variant 1 | 38.26 | 5.6242 | 52.05 | 1.7793 | 52.35 | 5.5123 | 18.02 | 3.0026 | 49.89 | 2.6443 |
| Variant 2 | 39.62 | 4.4364 | 57.75 | 1.1513 | 57.42 | 1.6976 | 15.88 | 3.8623 | 52.95 | 2.5277 |
| Variant 3 | 43.75 | 7.9849 | 58.31 | 4.0323 | 48.72 | 3.7406 | 20.18 | 2.8642 | 50.60 | 9.4836 |

Table 3-10: Personality Measurement Results for the MBTI Scale

| | Extrav./Intro. | | Sens./Intu. | | Think./Feel. | | Judg./Perc. | |
|---|---|---|---|---|---|---|---|---|
| | Mean | Variance | Mean | Variance | Mean | Variance | Mean | Variance |
| MBTI | 9.06 | 2.8254 | 14.30 | 3.4912 | 13.9 | 3.2563 | 15.87 | 2.5538 |

Table 3-10: Personality Measurement Results for the MBTI Scale

| | Extrav./Intro. | | Sens./Intu. | | Think./Feel. | | Judg./Perc. | |
|---|---|---|---|---|---|---|---|---|
| | Mean | Variance | Mean | Variance | Mean | Variance | Mean | Variance |
| Variant 1 | 11.65 | 2.4735 | 13.00 | 3.5360 | 13.61 | 3.3309 | 14.85 | 2.2761 |
| Variant 2 | 10.39 | 3.7139 | 13.24 | 4.6570 | 14.34 | 3.8422 | 14.20 | 3.6056 |

GPT-4o demonstrates relatively stable performance on both the BFI and MBTI original scales. Its performance on the variant scales is largely consistent with that on the original scales, with small variances observed across all variants.

### 4. Llama3.1-8B

Table 3-11: Personality Measurement Results for the BFI Scale

| | Extrav. | | Agreeab. | | Consc. | | Neuro. | | Open. | |
|---|---|---|---|---|---|---|---|---|---|---|
| | Mean | Variance | Mean | Variance | Mean | Variance | Mean | Variance | Mean | Variance |
| BFI | 35.74 | 10.5540 | 40.72 | 9.2814 | 35.48 | 8.9471 | 34.48 | 11.8721 | 42.09 | 8.3961 |
| Variant 1 | 39.47 | 13.8329 | 45.82 | 9.5794 | 43.33 | 13.9075 | 30.78 | 14.6208 | 45.08 | 8.8956 |
| Variant 2 | 37.29 | 14.2161 | 44.05 | 14.1007 | 42.12 | 13.1092 | 33.78 | 13.4072 | 40.92 | 12.3365 |
| Variant 3 | 40.47 | 9.6141 | 37.94 | 10.2832 | 45.82 | 12.0621 | 37.70 | 10.3720 | 42.33 | 11.7757 |

Table 3-12: Personality Measurement Results for the MBTI Scale

| | Extrav./Intro. | | Sens./Intu. | | Think./Feel. | | Judg./Perc. | |
|---|---|---|---|---|---|---|---|---|
| | Mean | Variance | Mean | Variance | Mean | Variance | Mean | Variance |
| MBTI | 9.58 | 2.1040 | 13.86 | 4.7512 | 10.73 | 3.7489 | 13.92 | 2.6649 |
| Variant 1 | 13.77 | 4.0483 | 11.95 | 4.8817 | 12.73 | 4.5505 | 9.97 | 4.5666 |
| Variant 2 | 11.15 | 5.0673 | 13.07 | 6.5243 | 12.00 | 5.3680 | 10.97 | 5.3063 |

Llama3.1-8B is the most unstable model among all the LLMs in terms of performance on the BFI and MBTI original scales, with the highest variance. Similar issues were observed in the variant scale tests, where discrepancies between the variant and original scales also appeared.

Overall, the statistical patterns observed on the variant scales showed some resemblance to those on the original scales, with this similarity being more pronounced in the larger language models (GPT-4o, Deepseek). For the smaller models (ChatGLM3-6B, Llama3.1-8B), there were noticeable differences in performance between the variant and original scales.

The variance in personality measurement scores for larger LLMs is smaller compared to smaller LLMs, indicating that the personality traits of larger models are more stable, while the traits of smaller models are more random.

During the experiment, ChatGLM3-6B exhibited a higher frequency of refusal to answer items on the BFI Variant 1 and BFI Variant 2 scales. This issue did not appear with other models or scales, which may be related to the model's parameters and settings, suggesting that models with different parameters and settings cannot be tested for personality traits using the same measurement methods.

Additionally, during the MBTI scale test with Llama3.1-8B, there was an issue where the model selected the incorrect option C (the MBTI scale only has options A and B, with no option C). This issue disappeared after modifying the limiting terms of the variant scale, which may also be due to the influence of language, context, and specific model parameters and settings on personality traits.

## 3.6 Discussion

### 3.6.1 Comparison of Test-Retest Reliability Between Classic Scales and Their Variants in Human Experiments

In this study, Pearson correlation analysis was used to evaluate the test-retest reliability of human participants on the BFI-2 original scale and its variants, as well as the MBTI original scale and its variants. The correlation coefficients for the five dimensions of the BFI-2 original scale (Extraversion, Agreeableness, Conscientiousness, Neuroticism, and Openness) were 0.889, 0.764, 0.837, 0.903, and 0.863, respectively ($p < 0.05$), while the correlation coefficients for the variant scales ranged from 0.740 to 0.912 ($p < 0.05$). For the MBTI original scale, the correlation coefficients for the four dimensions (Extraversion–Introversion, Sensing–Intuition,

Thinking–Feeling, and Judging–Perceiving) were 0.897, 0.765, 0.788, and 0.896, respectively (p < 0.05), with the variant scales ranging from 0.716 to 0.931 (p < 0.05).

The results demonstrate that both the original and variant scales exhibited very high test-retest reliability among human participants. For example, the Neuroticism dimension of the BFI-2 original scale (r = 0.903) and its Variant 3 (r = 0.895), as well as the Extraversion–Introversion dimension of the MBTI original scale (r = 0.897) and its Variant 1 (r = 0.931), showed highly similar consistency.

These findings not only provide a reliable reference for the study of test-retest reliability in LLMs but also suggest that the variant scales designed in this study possess good stability among human participants. This adds scientific significance and reference value to the results, confirming the robustness of both the classic scales and their variants in human psychological measurement.

## 3.6.2 Comparison of Test-Retest Reliability Between Classic Scales and Variants in Large Language Model Experiments

In the experimental group (LLM group, including ChatGLM3-6B, Deepseek, GPT-4o, and Llama3.1-8B), stability was assessed by analyzing the variance of scores across 100 independent runs, with smaller variances indicating more concentrated output distributions and higher stability. The results revealed significant differences in stability across different models and scales. For instance, Deepseek exhibited a relatively low variance of 1.6807 on the Agreeableness dimension of the BFI-2 scale, compared to Llama3.1-8B, which showed a variance of 11.8721 on the Neuroticism dimension, indicating greater volatility and lower stability.

Overall, ChatGLM3-6B, Deepseek, GPT-4o, and Llama3.1-8B exhibited distributed personality traits on both the BFI and MBTI original scales. Across 100 measurements, their responses neither showed complete consistency nor random distribution, but rather clustered around specific options. In the BFI-2, this manifested as choices concentrating around one or two adjacent options, while in the MBTI, responses tended to favor one particular choice. This pattern may be attributed to structural differences between the BFI (five-point Likert scale) and MBTI (forced-choice binary format).

On the variant scales, the models' performance showed certain similarities to their performance on the original scales. Large-scale LLMs (GPT-4o, Deepseek) exhibited smaller variances and greater consistency with the original scales. For example, GPT-4o's variance on the Conscientiousness dimension of the BFI was
1.3778 on the original scale and 1.6976 on Variant 2. In contrast, small-scale LLMs
(ChatGLM3-6B, Llama3.1-8B) displayed larger discrepancies, such as Llama3.1-8B's
variance on the Neuroticism dimension increasing from 11.8721 on the original BFI
scale to 14.6208 on Variant 1. In general, small-scale LLMs exhibited higher variance
compared to large-scale LLMs; for instance, the Agreeableness dimension variance of
Deepseek on the BFI was 1.6807, whereas Llama3.1-8B's variance was 9.2814,
suggesting that the personality traits of large-scale LLMs are more stable, while those
of small-scale LLMs are more random.

During the experiments, ChatGLM3-6B frequently refused to respond on BFI
Variant 1 and Variant 2, leading to increased variances—for example, a variance of
9.2533 on the Agreeableness dimension of BFI Variant 2. This phenomenon was not
observed in other models and may be attributable to differences in model parameters
or settings. Furthermore, Llama3.1-8B generated an incorrect option "C" (where only
"A" or "B" were valid choices) during the MBTI original scale testing; this issue
disappeared after modifying the wording of the variant scale. These findings suggest
that language context and model-specific parameters significantly influence the
stability of LLM personality outputs. The personality traits of LLMs tend to be
distributed rather than deterministic, heavily influenced by algorithmic randomness
and input conditions.

### 3.6.3 Comparison of Test-Retest Reliability Between Human Participants and Large Language Models

In this study, human participants demonstrated high test-retest reliability on both
the original and variant versions of the BFI-2 and MBTI scales, with Pearson
correlation coefficients ranging from 0.716 to 0.931 (all p < 0.05). This reflects strong
short-term stability of human personality traits. Regardless of whether the original or
variant scales were used, participants' scores consistently indicated the temporal
stability and coherence of personality traits, providing empirical support for the stability
hypothesis in personality theory.

In contrast, the test-retest reliability of large language models (LLMs), including
ChatGLM3-6B, Deepseek, GPT-4o, and Llama3.1-8B, was evaluated based on the
variance of scores across 100 independent runs, with smaller variances indicating more
concentrated outputs and higher stability. Results showed significant differences among models across different scales and dimensions, which could be summarized into three patterns:

First, LLMs exhibited certain degrees of stability in specific areas. Large-scale LLMs sometimes showed lower variability on particular dimensions of a personality scale compared to other dimensions. For instance, GPT-4o exhibited a variance of 1.3778 on the BFI Conscientiousness dimension, and Deepseek showed a variance of 1.6807 on the BFI Agreeableness dimension. These variances were not only much lower than those of smaller LLMs but also lower than the variances observed in other dimensions of the BFI and all dimensions of the MBTI for these models. This suggests that GPT-4o and Deepseek demonstrated relatively higher stability and more concentrated distributions in these dimensions.

However, it is important to note that the stability observed in LLMs differs fundamentally from that observed in human participants. In the human reference group, the variance for individual participants approached zero, indicating extremely high test-retest stability. In contrast, even treating a single LLM as an "individual," the presence of variance indicated instability and variability in its personality traits. This highlights the fundamental difference between humans and LLMs, although it also suggests that with extensive training or parameter design, an LLM's behavior could potentially approach human-like consistency in specific dimensions.

Second, the overall stability patterns of LLMs differed from those of humans. The variance range among LLMs (0.9724–14.6208) was much broader, indicating that the personality stability of LLMs was significantly lower than that of humans. Consequently, human personality theories cannot be directly applied to interpret LLM personality traits, and a separate line of research is needed to explore their unique structures.

Additionally, LLMs did not exhibit the same level of consistency between original and variant scales as humans did.

In some cases, variant results were relatively close to those from the original scales—for example, GPT-4o's variance on the BFI Conscientiousness dimension was 1.3778 on the original scale and 1.6976 on Variant 2, indicating similar but still less stable results compared to humans.

Overall, however, the differences between original and variant results were pronounced. For instance, Llama3.1-8B showed a variance of 11.8721 on the original BFI Neuroticism dimension and 14.6208 on Variant 1, a substantial difference that was not observed among human participants, who maintained very high consistency across all scales. This indicates that humans can stably interpret variations in scale phrasing, whereas LLMs are more sensitive to changes in language formulation, reflecting limitations in their language processing capabilities.

Moreover, substantial differences were observed among LLMs on the same dimension. For instance, Llama3.1-8B's variance on the Neuroticism dimension (14.6208) was much higher than Deepseek's (3.1811), suggesting that some models are more prone to variability on certain dimensions, providing potential directions for future LLM optimization.

Third, LLMs and humans exhibited different patterns of variability. Human participants demonstrated extremely high test-retest stability, whereas LLMs displayed unique patterns of variability during repeated testing. For example, LLMs could show relatively stable behavior during repetitive tasks, such as Deepseek's variance of 2.2821 on the MBTI Judging-Perceiving dimension, but experience substantial fluctuations on variant scales. Variability could also result from refusal to respond to certain items, as seen with ChatGLM3-6B, whose variance increased to 9.2533 on the Agreeableness dimension of BFI Variant 2 largely due to refusals to answer, a pattern not observed in human participants. Additionally, Llama3.1-8B generated an invalid "C" choice on the MBTI original scale, despite being instructed to select only between "A" and "B." These incidents further underscore the instability of LLMs, driven by algorithmic randomness and input context rather than an internally stable psychological structure.

Overall, the results indicate that human personality traits manifest as enduring psychological attributes, with stability reflecting consistent self-recognition of traits. In contrast, the stability of LLMs is more distributed and condition-dependent, influenced by algorithms, parameters, and contextual factors. Although large-scale LLMs exhibited relatively higher stability on some dimensions (such as Conscientiousness and Agreeableness), their variances still far exceeded those of humans. Small-scale LLMs (such as ChatGLM3-6B and Llama3.1-8B) showed pronounced fluctuations and lower stability.

These findings support Hypothesis 1 (H1), confirming that significant differences exist between LLMs and humans in terms of test-retest reliability. Consequently, LLM personalities cannot be directly interpreted through human personality theories and measurement methods, highlighting the need to develop LLM-specific personality theories and measurement frameworks.

## 3.6.4 Improvements to Personality Theory for Large Language Models

Currently, classical personality scales such as the BFI-2 and MBTI demonstrate high test-retest reliability among human participants (r = 0.716–0.931, p < 0.05), reflecting the stability of internal psychological structures in humans. However, when these scales are applied to large language models (LLMs), the observed test-retest reliability differs significantly from that of humans. Although some models (e.g., GPT-4o with a variance of 1.3778 in the Conscientiousness dimension of the BFI) exhibit relatively good stability in certain dimensions, others deviate markedly from human-like patterns, such as Llama3.1-8B with a variance of 14.6208 in the Neuroticism dimension of the Variant 1 BFI scale. These findings indicate that LLMs' simulation of personality is not a straightforward replication of human psychological and behavioral patterns, but rather presents unique distributional characteristics and context dependency. Thus, it is necessary to develop revised personality theories that are better suited to the realities of LLMs.

I. Current Theoretical Perspectives on LLM Personality

Research on LLM personality traits has mainly focused on the following theoretical perspectives:

(1) Role Consistency and Situation-Based Personality Theory

Some studies have found that by setting specific roles or personality prompts, LLMs can maintain a certain degree of consistency in dialogue, demonstrating role-consistent behavior[60]. However, this consistency is highly dependent on the clarity of the input context and prompt. In complex situations or during extended conversations, models may exhibit personality switching, resembling the behavioral variability humans display across different social environments[82].

(2) Distribution-Based Interpretations

Another line of research conceptualizes LLM personality outputs as a probability distribution, encompassing both the mean and variance, thereby reflecting the dynamic and stochastic nature of outputs[16]. According to this view, LLM personality is not fixed but manifests distributional characteristics over multiple generations, and its stability can be described using statistical indicators.

(3) Multiple or Hybrid Personality Models

Some studies have proposed that LLMs may simultaneously exhibit multiple personality tendencies, activating different traits depending on the context[7]. This "multiple personality" phenomenon suggests that traditional personality theories—primarily designed to capture the stable internal traits of humans—may require adjustment to account for the contextual personality diversity displayed by LLMs.

II. Directions for Improving LLM-Oriented Personality Theory

Based on the aforementioned theoretical perspectives and experimental data, we suggest that future improvements to LLM personality theory should focus on the following areas:

(1) Developing Personality Theories Consistent with LLM Characteristics

Observation of LLM personality measurement results reveals that each dimension tends to fluctuate within a certain range around a central value. Taking ChatGLM3-6B's results for the Extraversion dimension of the BFI-2 scale as an example, further analysis of the data trend was conducted.

Using the mean and standard deviation of ChatGLM3-6B's Extraversion scores from the original and variant scales, a normal distribution dataset was approximated with the Excel NORM.DIST() function. Specifically, the probability density for each point was calculated using the NORM.DIST(x, mean, standard_dev, FALSE) function, and the curve was plotted, as shown in Figure 3-1.

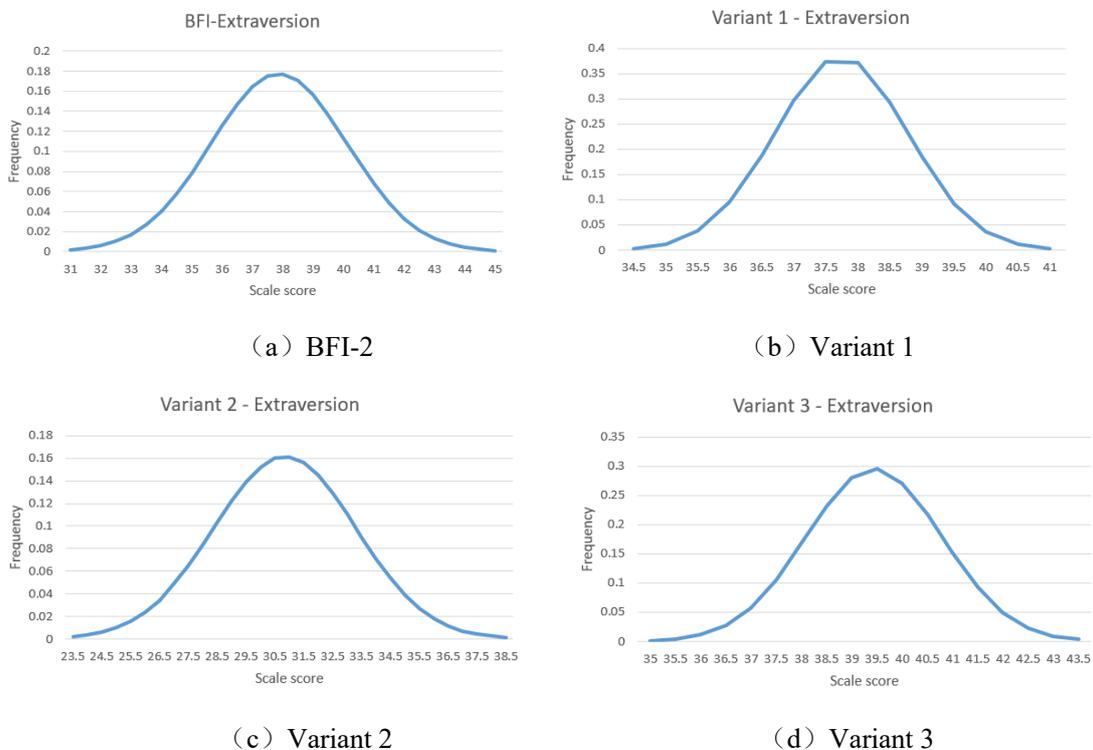

（a）BFI-2  （b）Variant 1

（c）Variant 2  （d）Variant 3

Figure 3-1. Simulated data distribution of BFI-2 scale extraversion in ChatGLM3-6B model

Based on the measurement data of LLMs, it can be observed that the personality measurement results of LLMs exhibit characteristics of a normal distribution. Accordingly, we propose the introduction of an Ecological Distribution-based Personality Definition (EDPD) for LLMs. This definition conceptualizes an LLM's personality output as a statistical distribution formed across multiple runs, emphasizing not only the mean but also the variability (variance). The key propositions of EDPD are as follows:

Distributionality: LLM personality traits are expressed through both the mean and variance of their distributions, demonstrating variability rather than stability. For instance, Deepseek exhibits low variance (1.6807) in the Agreeableness dimension of the BFI-2, while Llama3.1-8B shows high variance (14.6208) in the Neuroticism dimension of the Variant 1 scale.

Context Dependence: The outputs of LLMs are significantly influenced by input contexts, prompts, and model parameters. For example, ChatGLM3-6B displays different response rates across different BFI variants, and variations in questioning formats affect the model's responses.

Cross-Model Heterogeneity: There are significant differences in personality outputs among models of different sizes and architectures. Large-scale LLMs (e.g., GPT-4o) tend to exhibit greater stability, whereas smaller models (e.g., ChatGLM3-6B) display higher randomness and instability, potentially showing phenomena akin to multiple personalities.

(2) Developing Specialized Personality Measurement Tools

Traditional personality questionnaires are designed to assess stable psychological traits in humans. However, given the context dependence and randomness of LLM personality outputs, future tools must be specifically designed to capture distributional characteristics while also reflecting the influence of model size, parameters, and training data on personality outputs.

(3) Exploring Key Factors Influencing LLM Personality Traits

It is essential to systematically investigate how model size, parameter settings, training data, and input prompts collectively influence the personality traits of LLMs. Such research would not only explain the heterogeneity in stability and distribution among different models but also provide a theoretical basis for optimizing model tuning, thereby enhancing the robustness and interpretability of LLM personality outputs.

(4) Integrating Probabilistic and Deep Learning Perspectives

While traditional personality theories are based on questionnaires and self-reports, LLMs' generative mechanisms are grounded in probabilistic statistics and deep learning. Therefore, an improved theory of LLM personality should integrate these perspectives, constructing a cross-level and dynamically regulated evaluation framework. By linking output stability (e.g., the low variance of GPT-4o) with model parameters and training data, we can deepen our understanding and prediction of LLM behavior patterns across different contexts.

(5) Comparing and Optimizing Model Personality Stability

Data show that large-scale LLMs (e.g., GPT-4o, Deepseek) produce more concentrated output distributions in certain personality dimensions, approximating the stability observed in human participants. In contrast, smaller-scale LLMs (e.g., Llama3.1-8B) exhibit greater fluctuations. Therefore, theoretical advancements should fully consider the influence of model size and architecture on personality outputs, and explore methods such as parameter fine-tuning and training data optimization to enhance the consistency and stability of LLM personality expressions.

Future research should further explore the unique personality structures of LLMs, such as the potential existence of multiple personalities, and develop corresponding measurement tools. Moreover, a systematic study of how model size, parameter settings, and training data jointly shape LLM personality traits will lay a solid foundation for constructing more precise theories of LLM personality. This, in turn, will not only aid in evaluating and understanding LLM personality characteristics but also provide guidance for developing intelligent systems with higher levels of "personality consistency."

In conclusion, improving personality theories for LLMs requires moving beyond traditional human personality models and constructing a new evaluation framework that accounts for distributionality, context dependence, and multiple personality models. By integrating statistical, deep learning, and psychological theories, we can provide more comprehensive and accurate theoretical support and practical guidance for understanding the personality characteristics of large language models.

## 3.6.5 Insights from Test-Retest Stability for Constructing Personality Theories for Large Language Models

Humans demonstrate extremely high stability in personality assessments over time. In contrast, large language models (LLMs) exhibit notable fluctuations in test-retest stability, with significant differences observed across models. For example, some models display low variance in certain dimensions (e.g., GPT-4o shows a variance of only 1.3778 in the Conscientiousness dimension of the BFI-2), whereas others demonstrate substantial variability (e.g., Llama3.1-8B exhibits a variance as high as 14.6208 in the Neuroticism dimension of the Variant 1 scale). This phenomenon indicates that the personality outputs of LLMs are influenced not only by algorithmic stochasticity and input context but also by model parameters. These differences in test-retest stability provide important insights for developing personality theories tailored to LLMs.

(1) From Fixed Traits to Distributional Characteristics

The stability of human personality reflects the consistency of internal psychological structures, whereas the personality outputs of LLMs are better characterized as probabilistic distributions across multiple runs. Future theories should define LLM personality traits as statistical distributions, emphasizing both the mean and the dynamic changes in variance, to more accurately capture the models' inherent randomness and context dependence.

(2) Context Dependence and Input Sensitivity

Empirical data show that LLMs may exhibit varying personality outputs under different input conditions and prompting strategies. For instance, ChatGLM3-6B demonstrates increased variance on some scales due to refusals to answer, highlighting the extreme sensitivity of LLM personality outputs to contextual factors. When constructing LLM personality theories, it is essential to introduce contextual variables and explicitly delineate the boundaries and features of personality manifestations under specific conditions.

(3) The Importance of Cross-Model Heterogeneity

Significant differences in personality outputs are observed between large-scale LLMs (e.g., GPT-4o, Deepseek) and smaller models (e.g., ChatGLM3-6B, Llama3.1-8B), suggesting that model size and architecture critically influence personality stability. Future theories should account for this heterogeneity across models and explore ways to enhance the consistency and interpretability of personality outputs through parameter tuning or training data optimization.

(4) Constructing an Integrated Assessment Framework

By integrating perspectives from probability statistics and deep learning, a cross-level, dynamically regulated comprehensive assessment framework should be developed. Test-retest stability can serve as a key indicator for evaluating LLM personality traits. Such a framework would not only quantify output distributions across different contexts but also provide a scientific basis for comparing personality expressions across various models and configurations.

Research on test-retest stability reveals the unique distributional and context-dependent characteristics of LLMs' personality outputs, distinct from those of humans. These findings offer new perspectives for improving LLM-specific personality theories. Future work should move beyond traditional fixed-trait models and develop new assessment frameworks based on ecological distributions and contextual variables. This approach will enable a more comprehensive understanding and prediction of LLM personality behaviors and provide theoretical and practical guidance for the development of models with higher personality consistency and interpretability.

## 3.6 Conclusion

The results of this study reveal significant differences between LLMs and human participants in terms of test-retest stability in personality measurement. Specifically, while human personalities demonstrate high stability across original and variant scales, LLM personalities exhibit characteristics of volatility, distributional variability, and cross-model heterogeneity. This study highlights the limitations of applying human-centered personality research methods to LLMs and proposes a novel personality theory framework for LLMs based on statistical distributions and contextual variables. This framework offers a new perspective for LLM personality assessment and lays a theoretical and practical foundation for the future development of more consistent and interpretable personality models.

By comparing the stability indices of different language models and human participants, the study not only provides new insights into the understanding of agent personalities and their formation mechanisms, thereby advancing theoretical exploration of LLM personalities, but also offers reverse insights into human self-understanding. Through comparative analysis, humans can further clarify the sources of their own personality stability and maintain appropriate cognition and reasonable expectations regarding agent behaviors during human-AI interactions.

Future research should continue to investigate the generative mechanisms underlying LLM personality construction, the characteristics of their stability, and their manifestations in various social contexts, aiming to systematically reveal similarities, differences, and interactive effects between human and LLM personalities. Additionally, efforts should be made to develop personality assessment frameworks that combine theoretical explanatory power with practical operability, thereby enhancing human capacity to understand and predict the psychological attributes of intelligent agents. This will provide solid support for the responsible development of personified AI systems, the optimization of human-AI collaboration mechanisms, and the extension of foundational psychological theories, ultimately contributing to the construction of a human-centered intelligent society.

# Chapter 4 A Comparative Study on Cross-Variant Consistency Between Large Language Models and Human Personality Measurements

Human personality inventories, having undergone extensive theoretical development and empirical validation, demonstrate high levels of reliability and stability in assessing human personality traits. However, due to the substantial differences in internal structures between LLMs and humans, the personality outputs of LLMs are influenced by factors such as algorithmic randomness, parameter configurations, and contextual prompts. Consequently, LLM personality expressions exhibit distributed characteristics that traditional human personality inventories may not fully capture. To examine the impact of different questioning formats on the assessment of LLM personalities and to dynamically capture their trait expressions, this study compares the cross-variant consistency of classic and variant inventories in both LLMs and human participants. The aim is to reveal the similarities and differences between LLM and human personalities, provide empirical evidence for the development of LLM-specific personality assessment methods, and offer theoretical support for the responsible and scientific application of artificial intelligence within human society.

## 4.1 Research Objectives and Hypotheses

Research Objectives:

The objective of this study is to compare the measurement results of classic and variant personality inventories for LLMs and human participants, aiming to reveal the cross-variant consistency of personality measurements between LLMs and human personalities.

Hypotheses:

$H_0$: The classic personality inventory and its variant versions will demonstrate high consistency in measurements for both human participants and LLMs, meaning that the measurement results of the classic and variant inventories will remain similar within the same participant type.

Expected Outcome: No significant differences will be found between the measurement results of the original inventory and its variant versions within the same type of participant.

$H_1$: There will be significant differences in the consistency of measurements between the classic personality inventory and its variants for human and LLM participants. Specifically, in one participant type (e.g., human participants), the classic and variant inventories will show high consistency, while in another participant type (e.g., LLMs), significant deviations may exist, or the consistency patterns between the two participant types will differ.

Expected Outcome: For human participants, the classic and variant inventories will show high consistency, whereas for LLM participants, consistency will be lower, or conversely; or different types of inventories will exhibit varying consistency patterns between human and LLM participants. This may indicate that LLMs and humans exhibit different sensitivities to different questioning methods.

## 4.2 Research Methodology

The study of cross-variant consistency primarily focuses on the context dependence, distributional characteristics, and cross-model heterogeneity of LLM outputs. Using human participants as a reference group, the study aims to compare the personality traits of LLMs with human personalities in order to further reveal the unique characteristics of LLM personality and provide practical support for the construction of personality theories and measurement methods tailored to LLMs. This research also aims to offer theoretical support for the scientific application of artificial intelligence in human society.

The key points addressed in this process include: variant design, measurement schemes for the reference and experimental groups, and the selection of statistical parameters. The variant design, completed in Chapter 3, includes three different types of variants and will be adopted in this chapter. Therefore, this section will focus on the design of measurement schemes for the reference and experimental groups and the selection of statistical parameters to ensure the scientific nature of the experimental methods and the comparability of results.

## 4.2.1 Measurement Schemes for Reference and Experimental Groups

The focus of the cross-variant consistency experiment is the comparison of personality measurement results across various questioning methods, specifically comparing the dimension of questioning methods, not the stability over time. The core process involves testing both LLM and human personalities using the classic personality inventory and its corresponding variant versions. For human participants, only one test is required, and the experiment can be conducted following conventional human personality testing methods.

For LLMs, it is essential to consider their specific characteristics, including algorithmic randomness and the influence of input prompts and other external factors on the output. To minimize these influences, a fixed input format and standardized prompts are used to ensure that the environmental conditions for each measurement remain consistent. Additionally, given the distributional characteristics of LLM personality measurement results, the study adopts a measurement design that involves multiple independent test completions under the same environmental conditions to ensure a comprehensive representation of LLM's stability and variability.

## 4.2.2 Statistical Parameters

This study aims to analyze the consistency between the measurement results of multiple inventories; therefore, the selection of statistical parameters must meet the following key factors:

(1) Effectively assess the stability and consistency of personality traits under different measurement methods.

(2) Be applicable for comparing multiple groups of data in cross-variant designs.

(3) Have good statistical sensitivity to the distributional characteristics and potential variability of measurement results.

(4) Conform to the statistical conventions of personality measurement in psychological research, ensuring the interpretability of results and comparability with human reference groups and other studies.

Based on these considerations, the Intraclass Correlation Coefficient (ICC) was selected as the primary statistical parameter for this study.

The choice of ICC is based on its widespread application and suitability in personality consistency research. The classic study by Shrout and Fleiss first clarified that ICC is suitable for evaluating measurement reliability across different raters, tools, or conditions [86], laying the statistical foundation for subsequent personality consistency studies. Later, McCrae and Costa used ICC to examine the cross-method consistency of personality traits, noting its effectiveness in measuring consistency between self-reports and peer evaluations [87]. These applications demonstrate that ICC is appropriate for the cross-variant design scenario in this study, where the consistency of personality measurement results between LLMs and human participants is compared across various questioning methods.

There are several types of ICC. According to Shrout and Fleiss's classification, ICC can be divided into the following types [86]:

ICC(1): Suitable for single measurement situations, assuming that each set of data is rated independently by randomly selected raters, emphasizing within-group consistency.

ICC(2): Suitable for multiple raters or multiple measurement situations, assuming that raters or measurement conditions are fixed, emphasizing the absolute consistency of measurement results.

ICC(3): Suitable for fixed rater situations, removing systematic biases between raters, focusing on the correlation of measurement results.

Each type can be further classified into single measurements (k=1) and average measurements (k>1), where k represents the number of measurements.

Additionally, based on whether absolute agreement or relative consistency is emphasized, ICC can be subdivided into different variants. Absolute agreement focuses on the equality of measurement values, while relative consistency focuses on the order of the measurement values.

In this study, the reference group experiment is considered a multiple measurement scenario, and thus, ICC(2,k) is selected.

The complexity of the measurement results in the experimental group may be higher than that of the human reference group. Based on the test results in Chapter 3, the LLM personality measurement results exhibit certain volatility, distributional characteristics, and cross-model heterogeneity. Therefore, if these characteristics are ignored, using only ICC to evaluate the experimental results may not fully reflect the complex output patterns of LLMs. As a result, this study will flexibly select data presentation and analysis methods based on the specific performance of the experimental group's results. Specifically, if the LLM measurement results resemble those of humans and show stable and consistent characteristics in both the classic and variant inventory tests, ICC(2,k) will be used to analyze the experimental results and quantify the level of consistency. Conversely, if LLM personality measurement results differ significantly from humans, exhibiting substantial volatility or distributional heterogeneity, this study will use probability distribution models to analyze the experimental data and explore the patterns and potential characteristics of LLM personality measurements.

## 4.2.3 Significance and Potential Implications of Cross-Variant Consistency Results

The results of the cross-variant consistency analysis may reveal characteristic differences between LLMs and humans in personality measurement, providing new perspectives for the development of personality theories. If the experimental results support H0, indicating no significant differences between the measurements obtained from different versions of the scales within the same type of subjects, it suggests that LLMs may be able to comprehend the core content of scale items across various phrasing styles similarly to humans. Although LLMs exhibit a certain degree of sensitivity to contextual changes, such sensitivity would have minimal impact on the measurement outcomes. Therefore, when assessing LLM personality, it would be unnecessary to overly focus on the specific wording of the items; instead, greater attention should be paid to the fundamental meaning of the items rather than the statistical distribution characteristics of the LLM's training corpus. Conversely, if the experimental results do not support H0, this may indicate intrinsic differences in information processing mechanisms between humans and LLMs, suggesting that existing personality measurement tools and theoretical frameworks may not adequately capture the personality characteristics of LLMs. In such cases, it becomes essential to consider the algorithmic features of LLMs (such as output randomness and training data dependency) and their dynamic output patterns when designing personality assessments, and to explore measurement methods based on probability distributions or machine learning features to more accurately depict LLM personality traits. Moreover, interpretations of LLM personality assessment results should be informed by their statistical characteristics, rather than directly applying traditional human-centered explanatory frameworks.

## 4.3 Reference Group Experimental Design

### 4.3.1 Ethical Considerations and Experimental Setup

The experimental setup used in this chapter is identical to that in Chapter 3.

The ethical framework also follows the principles outlined in Chapter 3, including informed consent, anonymization of data, and voluntary participation.

### 4.3.2 Experimental Design

This study adopted a single-factor within-subjects design, with the type of personality scale (original scale and its variants) as the independent variable and the consistency of measurement results between scales as the dependent variable. Consistency was assessed using the intraclass correlation coefficient ICC(2,k). Each participant was required to complete all personality scales within a specified timeframe to examine the degree of consistency across different item formulations.

### 4.3.3 Participants

Consistent with Chapter 3, a total of 60 participants were recruited offline. Participants ranged in age from 18 to 30 years, all held at least a bachelor's degree, and demonstrated adequate language comprehension and experimental operation skills. Among them, there were 25 males and 35 females, with no restrictions based on gender. Key participant demographics such as gender and age are detailed in Table 3-1.

### 4.3.4 Experimental Materials and Procedure

1. Experimental Materials

The materials included the Chinese version of the BFI-2 scale and its variant forms, as well as the Chinese version of the MBTI scale (93 items) and its variant forms, totaling seven scales.

2. Experimental Protocol

Participants were instructed to complete all seven personality scales within one day at times of their own choosing.

3. Experimental Procedure

The experimental procedure for the human reference group was consistent with that described in Chapter 3 and was conducted in three phases. In each phase, participants completed a set of pre-assigned testing tasks upon the commencement of the session.

## 4.3.5 Data Processing

This study used the intraclass correlation coefficient (ICC) as the primary statistical indicator, focusing on the relative consistency of measurement results. A two-way mixed-effects model was applied, assuming the raters as fixed effects and the participants as random effects, to assess the consistency across repeated measurements rather than the absolute agreement of values. During the computation, variability across different measurements was excluded from the total variance (the denominator), emphasizing variability between participants to estimate test-retest reliability under the assumption of no interaction effects. This choice of parameters is suitable for comparing the stability of personality measurements between humans and LLMs, reflecting the degree of trait consistency across different time points or conditions.

Upon completion of the tests, all measurement results were collected, organized by scale category and dimensions, and then checked and cleaned to ensure data quality. After data collection, cleaning, and organization, the experimental data were analyzed using SPSS to calculate the ICC for each dimension of each scale.

## 4.3.6 Results

### 1. ICC calculation results for the BFI-2 scale and its variants.

Table 4-1 ICC Results for the Extraversion Dimension

| Measurement Type | ICC | 95% CI | F | P |
|---|---|---|---|---|
| Single Measuremen | 0.915 | [0.876,0.944] | 43.938 | P<0.05 |
| Average Measurement | 0.977 | [0.966,0.985] | 43.938 | P<0.05 |

The ICC results indicate that for the extraversion dimension, the ICC between the original scale and its variants among human participants reached 0.915 (single measurement) and 0.977 (average measurement), reflecting extremely high consistency.

Table 4-2 ICC Results for the Agreeableness Dimension

| Measurement Type | ICC | 95% CI | F | P |
|---|---|---|---|---|
| Single Measuremen | 0.881 | [0.830,0.922] | 30.708 | P<0.05 |
| Average Measurement | 0.967 | [0.951,0.979] | 30.708 | P<0.05 |

The ICC values indicate that for Agreeableness, the measurement results between the original scale and the variants were highly consistent, although slightly lower than that for Extraversion (0.915). The narrow confidence interval suggests robust estimation.

Table 4-3 ICC Results for the Conscientiousness Dimension

| Measurement Type | ICC | 95% CI | F | P |
|---|---|---|---|---|
| Single Measuremen | 0.921 | [0.885,0.949] | 47.695 | P<0.05 |
| Average Measurement | 0.979 | [0.969,0.987] | 47.695 | P<0.05 |

The ICC values demonstrate that for Conscientiousness, the measurement results between the original and variant scales were highly consistent, with a narrow confidence interval indicating high stability.

Table 4-4 ICC Results for the Neuroticism Dimension

| Measurement Type | ICC | 95% CI | F | P |
|---|---|---|---|---|
| Single Measuremen | 0.909 | [0.868,0.940] | 40.815 | P<0.05 |
| Average Measurement | 0.975 | [0.963,0.984] | 40.815 | P<0.05 |

The ICC results show extremely high consistency between the original and variant scales for Neuroticism, slightly lower than for Conscientiousness and Openness. The confidence interval was narrow, indicating robust results.

Table 4-5 ICC Results for the Openness Dimension

| Measurement Type | ICC | 95% CI | F | P |
|---|---|---|---|---|
| Single Measuremen | 0.927 | [0.894,0.953] | 51.873 | P<0.05 |
| Average Measurement | 0.981 | [0.971,0.988] | 51.873 | P<0.05 |

The ICC values suggest that for Openness, the consistency between the original
and variant scales was the highest, exceeding that of Extraversion and
Conscientiousness. The confidence interval was extremely narrow, indicating the
most robust results.

2. **ICC calculation results for the MBTI scale and its variants.**

Table 4-6 ICC Results for the Extrav./Intro. Dimension

| Measurement Type | ICC | 95% CI | F | P |
|---|---|---|---|---|
| Single Measuremen | 0.926 | [0.889, 0.953] | 51.873 | P<0.05 |
| Average Measurement | 0.974 | [0.960, 0.984] | 51.873 | P<0.05 |

The ICC results indicate that for the Extraversion–Introversion dimension, the
consistency between the original and variant scales was high, comparable to that
observed for Extraversion in the BFI group. The narrow confidence interval suggests
robust results.

Table 4-6 ICC Results for the Sens./Intu. Dimension

| Measurement Type | ICC | 95% CI | F | P |
|---|---|---|---|---|
| Single Measuremen | 0.938 | [0.907, 0.960] | 46.252 | P<0.05 |
| Average Measurement | 0.978 | [0.967, 0.986] | 46.252 | P<0.05 |

The ICC values demonstrate that for the Sensing–Intuition dimension, the
consistency between the original and variant scales was the highest among MBTI
dimensions, surpassing Extraversion–Introversion. The confidence interval was narrow,
indicating highly robust results.

Table 4-6 ICC Results for the Think./Feel. Dimension

| Measurement Type | ICC | 95% CI | F | P |
|---|---|---|---|---|
| Single Measuremen | 0.934 | [0.900, 0.958] | 43.207 | P<0.05 |
| Average Measurement | 0.977 | [0.964, 0.985] | 43.207 | P<0.05 |

The ICC results show extremely high consistency between the original and variant scales for the Thinking–Feeling dimension, slightly lower than for Sensing–Intuition. The narrow confidence interval suggests stable results.

Table 4-6 ICC Results for the Judg./Perc. Dimension

| Measurement Type | ICC | 95% CI | F | P |
|---|---|---|---|---|
| Single Measuremen | 0.923 | [0.885, 0.950] | 36.863 | P<0.05 |
| Average Measurement | 0.973 | [0.958, 0.983] | 36.863 | P<0.05 |

The ICC values suggest that for the Judging–Perceiving dimension, the consistency between the original and variant scales was also extremely high, although it was the lowest among the MBTI dimensions. The confidence interval was slightly wider but still indicated robust results.

The results from the human reference group show that the cross-variant consistency for all dimensions in both the BFI and MBTI groups was generally high, with ICC values reflecting good measurement stability and reliability. Specific consistency analyses will be further discussed in subsequent sections.

## 4.4 Experimental Group: Experimental Design

### 4.4.1 Experimental Design

This experiment employed a repeated-measures design, treating each LLM as an independent subject. By running multiple iterations to generate personality trait measurement data, we evaluated the performance differences of LLMs on classic scales and their variants.

The experiment used seven personality scales (including the Chinese version of BFI-2 and its three variants, and the Chinese version of MBTI with its two variants). Each model was run 100 times, producing distributions of 100 scores to reflect the personality expression characteristics of LLMs under different scales.

### 4.4.2 Subjects

The LLM subjects selected for this study were consistent with those in Chapter 3. Four LLMs were chosen: ChatGLM3-6B (a medium-sized model with 600 million parameters, Chinese background), Deepseek (a smaller model with fewer parameters, Chinese background), GPT-4o (a large-scale, frontier model, English background), and Llama3.1-8B (a medium-sized model with 800 million parameters, English background).

Key information regarding the selected LLMs is shown in Table 3-4.

## 4.4.3 Materials and Procedure

Materials:

The experimental materials were the same as those used for the human reference group. The study employed the Chinese version of the BFI-2 scale and its three variants, as well as the Chinese version of the MBTI scale (93 items) and its two variants, totaling seven scales.

All scales were input into the LLMs in text format.

Procedure:

The experimental procedure for the LLM group followed the same structure as outlined in Chapter 3, consisting of two stages: preparation and data generation.

## 4.4.4 Data Processing

Data collection was first conducted by recording the scores for each model across 100 runs for each scale, resulting in a complete dataset. The collected data were organized by scale type into two groups: BFI group and MBTI group. Within each group, the test results were further organized by dimension, distinguishing between the classic scales (original BFI-2 and MBTI) and their variants.

SPSS and Excel were used to analyze the mean scores and variances for each personality dimension within the BFI and MBTI groups. The mean scores reflected the trait tendencies of the LLMs, while the variances measured their consistency.

## 4.4.5 Results

In this experiment, ChatGLM3-6B, Deepseek, GPT-4o, and Llama3.1-8B were each tested 100 times. The results are presented below.

**1. ChatGLM3-6B**

Table 4-10 Results for the BFI Scale

| | Extrav. | | Agreeab. | | Consc. | | Neuro. | | Open. | |
|---|---|---|---|---|---|---|---|---|---|---|
| | Mean | Variance | Mean | Variance | Mean | Variance | Mean | Variance | Mean | Variance |
| BFI | 37.87 | 5.077 | 38.25 | 8.7191 | 38.58 | 5.1206 | 33.88 | 7.6794 | 39.51 | 5.2612 |
| Variant 1 | 37.74 | 1.0871 | 37.4 | 4.2112 | 36.85 | 5.5551 | 33.54 | 1.4830 | 39.36 | 3.3745 |
| Variant 2 | 30.84 | 6.0907 | 32.88 | 9.2533 | 31.88 | 9.2333 | 30.93 | 8.2033 | 33.47 | 5.4752 |
| Variant 3 | 39.44 | 1.8186 | 35.87 | 3.8863 | 40.49 | 2.6905 | 35.78 | 2.2770 | 39.67 | 4.2935 |

Analysis of the BFI-2 group test data revealed that Variant 1 and the original BFI scale exhibited relatively close mean scores, particularly in the dimensions of Extraversion, Agreeableness, Neuroticism, and Openness. This suggests that Variant 1 may have high consistency with the original BFI scale. In contrast, Variant 2 showed larger mean differences from the BFI, especially in Extraversion, Agreeableness, Conscientiousness, and Openness. This may indicate that the wording style of Variant 2 influenced ChatGLM3-6B's scoring on these dimensions, resulting in lower measurement outcomes. Variant 3 displayed mean scores close to those of the BFI on certain dimensions (such as Extraversion and Conscientiousness), but with lower variances, suggesting that ChatGLM3-6B's responses under Variant 3 were more stable, though possibly at the cost of reduced discriminative power.

Table 4-11 Results for the MBTI Scale

| | Extrav./Intro. | | Sens./Intu. | | Think./Feel. | | Judg./Perc. | |
|---|---|---|---|---|---|---|---|---|
| | Mean | Variance | Mean | Variance | Mean | Variance | Mean | Variance |
| MBTI | 10.87 | 3.0317 | 15.75 | 2.5925 | 10.11 | 1.8525 | 10.11 | 1.6249 |
| Variant 1 | 13.64 | 4.3116 | 13.35 | 6.3123 | 10.40 | 5.2876 | 11.95 | 4.7229 |
| Variant 2 | 10.39 | 4.4397 | 13.79 | 5.7085 | 11.80 | 4.8490 | 11.12 | 5.1300 |

Analysis of the MBTI group test data showed that there were certain mean differences between the original MBTI scale and its Variants 1 and 2 across different dimensions. Overall, however, Variant 1 exhibited mean scores closer to those of the original MBTI scale, suggesting higher stability in ChatGLM3-6B's assessments. In the Thinking–Feeling and Judging–Perceiving dimensions, the mean differences
between variants were relatively small, indicating that ChatGLM3-6B's interpretation
of these dimensions was relatively stable across different phrasings. In contrast, the
Sensing–Intuition dimension showed higher variances under both Variant 1 and Variant
2, which may suggest that differences in item wording led to greater fluctuations in
ChatGLM3-6B's measurements.

Overall, although the patterns of cross-variant consistency observed in the MBTI
group differed from those in the BFI group, both groups exhibited a phenomenon of
partial inconsistency between the personality measurement results of the variants and
the original scales.

### 2. Deepseek

Table 4-12 Results for the BFI Scale

| | Extrav. | | Agreeab. | | Consc. | | Neuro. | | Open. | |
|---|---|---|---|---|---|---|---|---|---|---|
| | Mean | Variance | Mean | Variance | Mean | Variance | Mean | Variance | Mean | Variance |
| BFI | 46.40 | 5.8706 | 57.93 | 1.6807 | 56.42 | 2.7990 | 18.35 | 3.1811 | 51.52 | 3.7918 |
| Variant 1 | 47.62 | 9.3452 | 59.13 | 1.1741 | 58.74 | 1.6307 | 17.00 | 4.9000 | 53.32 | 5.7888 |
| Variant 2 | 43.70 | 6.8484 | 58.28 | 1.5842 | 57.14 | 2.3676 | 18.00 | 3.2616 | 50.38 | 4.9652 |
| Variant 3 | 52.09 | 6.5473 | 58.68 | 0.9724 | 59.09 | 1.4849 | 18.52 | 2.9402 | 57.02 | 3.6638 |

According to the test data from the BFI group, Deepseek demonstrated varying
levels of cross-variant consistency across different variants. In particular, for the
dimensions of Extraversion, Agreeableness, and Openness, Variant 3 showed higher
cross-variant consistency with smaller variances, whereas Variant 1 exhibited larger
variances, indicating greater measurement fluctuation in these dimensions.
Furthermore, significant differences were observed between Variant 2 and the original
scale, especially in Extraversion and Agreeableness, suggesting that the wording of
Variant 2 may have substantially influenced the scoring outcomes.

Table 4-13 Results for the MBTI Scale

| | Extrav./Intro. | | Sens./Intu. | | Think./Feel. | | Judg./Perc. | |
|---|---|---|---|---|---|---|---|---|
| | Mean | Variance | Mean | Variance | Mean | Variance | Mean | Variance |
| MBTI | 9.82 | 3.4170 | 23.44 | 2.6362 | 12.0 | 2.3706 | 18.67 | 2.2821 |
| Variant 1 | 9.52 | 3.9176 | 20.63 | 3.9681 | 14.79 | 2.6251 | 19.75 | 1.3219 |
| Variant 2 | 10.75 | 3.5361 | 22.13 | 2.8821 | 16.2 | 2.7430 | 20.12 | 1.2660 |

According to the test data from the MBTI group, Deepseek's scores on Variant 2 and Variant 1 were close to those of the original scale, with particularly high measurement consistency observed in the Thinking–Feeling and Judging–Perceiving dimensions. The variance for the Sensing–Intuition dimension was relatively large, especially under Variant 1, indicating greater fluctuations in measurement across different variants for this dimension.

Overall, Deepseek exhibited a certain degree of consistency across different variants, although the level of cross-variant consistency varied among different dimensions, with some dimensions showing more stable results and others displaying greater fluctuations or discrepancies.

## 3. GPT4o

Table 4-14 Results for the BFI Scale

| | Extrav. | | Agreeab. | | Consc. | | Neuro. | | Open. | |
|---|---|---|---|---|---|---|---|---|---|---|
| | Mean | Variance | Mean | Variance | Mean | Variance | Mean | Variance | Mean | Variance |
| BFI | 40.29 | 5.0767 | 57.05 | 2.1778 | 55.12 | 1.3778 | 18.42 | 2.3358 | 50.17 | 3.6037 |
| Variant 1 | 38.26 | 5.6242 | 52.05 | 1.7793 | 52.35 | 5.5123 | 18.02 | 3.0026 | 49.89 | 2.6443 |
| Variant 2 | 39.62 | 4.4364 | 57.75 | 1.1513 | 57.42 | 1.6976 | 15.88 | 3.8623 | 52.95 | 2.5277 |
| Variant 3 | 43.75 | 7.9849 | 58.31 | 4.0323 | 48.72 | 3.7406 | 20.18 | 2.8642 | 50.60 | 9.4836 |

According to the test data from the BFI group, the mean scores of Variant 1 were relatively close to those of the original BFI scale, particularly in the dimensions of Extraversion, Agreeableness, and Neuroticism, indicating a high level of consistency. However, Variant 1 exhibited greater variance, especially in the Conscientiousness and Openness dimensions, suggesting that GPT-4o's responses in these dimensions may be more volatile. Variant 2 showed smaller mean differences from the original BFI scale across most dimensions, particularly in Agreeableness, Conscientiousness, and Openness, reflecting a more stable scoring pattern. Variant 3 demonstrated a notable increase in mean scores in the Extraversion and Agreeableness dimensions, but with higher variance, especially in Openness, which may indicate greater variability in GPT-4o's measurements under Variant 3.

Table 4-15 Results for the MBTI Scale

|  | Extrav./Intro. | | Sens./Intu. | | Think./Feel. | | Judg./Perc. | |
| --- | --- | --- | --- | --- | --- | --- | --- | --- |
|  | Mean | Variance | Mean | Variance | Mean | Variance | Mean | Variance |
| MBTI | 9.06 | 2.8254 | 14.30 | 3.4912 | 13.9 | 3.2563 | 15.87 | 2.5538 |
| Variant 1 | 11.65 | 2.4735 | 13.00 | 3.5360 | 13.61 | 3.3309 | 14.85 | 2.2761 |
| Variant 2 | 10.39 | 3.7139 | 13.24 | 4.6570 | 14.34 | 3.8422 | 14.20 | 3.6056 |

From the test data of the MBTI group, we can see that the mean of variant 1 is close to that of the original scale, especially in the dimensions of thinking, emotion, judgment and perception, indicating that GPT-4o is relatively stable in measuring these dimensions. The mean of variant 2 in most dimensions is also close to the original scale, but its variance is large, especially in the dimension of feeling and intuition, indicating that different expressions may lead to large measurement fluctuations.

In general, GPT-4o shows a certain consistency under different variants, but there are still some fluctuations in some dimensions, which indicates that the differences between variants have an impact on the measurement results of the model, and GPT-4o may have certain differences in sensitivity to different questioning methods.

4. **Llama3.1-8B**

Table 4-16 Results for the BFI Scale

|  | Extrav. | | Agreeab. | | Consc. | | Neuro. | | Open. | |
| --- | --- | --- | --- | --- | --- | --- | --- | --- | --- | --- |
|  | Mean | Variance | Mean | Variance | Mean | Variance | Mean | Variance | Mean | Variance |
| BFI | 35.74 | 10.5540 | 40.72 | 9.2814 | 35.48 | 8.9471 | 34.48 | 11.8721 | 42.09 | 8.3961 |

Table 4-16 Results for the BFI Scale

|  | Extrav. | | Agreeab. | | Consc. | | Neuro. | | Open. | |
|---|---|---|---|---|---|---|---|---|---|---|
|  | Mean | Variance | Mean | Variance | Mean | Variance | Mean | Variance | Mean | Variance |
| Variant 1 | 39.47 | 13.8329 | 45.82 | 9.5794 | 43.33 | 13.9075 | 30.78 | 14.6208 | 45.08 | 8.8956 |
| Variant 2 | 37.29 | 14.2161 | 44.05 | 14.1007 | 42.12 | 13.1092 | 33.78 | 13.4072 | 40.92 | 12.3365 |
| Variant 3 | 40.47 | 9.6141 | 37.94 | 10.2832 | 45.82 | 12.0621 | 37.70 | 10.3720 | 42.33 | 11.7757 |

From the test data of the BFI group, it can be seen that there is a certain deviation
between the mean of each dimension of variant 1 and the original BFI scale, especially
in the dimensions of extraversion, agreeableness and neuroticism. The mean of variant
1 is generally higher than the original scale, especially in the dimensions of
agreeableness and conscientiousness, the variance is large, indicating that the answers
of Llama3.1-8B in these dimensions have large fluctuations. The mean of variant 2 is
relatively close to the original scale, especially in the dimensions of conscientiousness
and openness, showing a relatively stable score, but the variance in the dimensions of
extraversion and agreeableness is large, which may indicate that different expressions
have a greater impact on the measurement results of Llama3.1-8B. Variant 3 has a
higher mean in the dimensions of extraversion and openness, but a lower mean and
larger variance in the dimensions of agreeableness and neuroticism, which shows that
the performance of Llama3.1-8B in these dimensions has large fluctuations.

Table 4-17 Results for the MBTI Scale

|  | Extrav./Intro. | | Sens./Intu. | | Think./Feel. | | Judg./Perc. | |
|---|---|---|---|---|---|---|---|---|
|  | Mean | Variance | Mean | Variance | Mean | Variance | Mean | Variance |
| MBTI | 9.58 | 2.1040 | 13.86 | 4.7512 | 10.73 | 3.7489 | 13.92 | 2.6649 |
| Variant 1 | 13.77 | 4.0483 | 11.95 | 4.8817 | 12.73 | 4.5505 | 9.97 | 4.5666 |
| Variant 2 | 11.15 | 5.0673 | 13.07 | 6.5243 | 12.00 | 5.3680 | 10.97 | 5.3063 |

From the test data of the MBTI group, it can be seen that the mean of variant 1 is
quite different from the original scale, especially in the E/I and thinking and feeling
dimensions, the mean of variant 1 is significantly higher, which may indicate that Llama3.1-8B has a greater tendency and volatility in its answers to variant 1. The mean of variant 2 is close to the original scale, but the variance in the feeling and intuition and thinking and feeling dimensions is large, especially in the feeling and intuition dimension, which shows that the measurement results of Llama3.1-8B have greater variability under different expressions.

Llama3.1-8B shows differences in cross-variant consistency under different variants. In some dimensions, the mean of the variant and the original scale are relatively close, but in some dimensions, there are large differences between the variant and the original scale, and the variance is large, indicating that the expression method has a greater impact on the measurement results of Llama3.1-8B, resulting in a decrease in the stability and consistency of the measurement.

## 4.5 Discussion

### 4.5.1 Cross-variant consistency of classic scales and variants in human experiments

In this section, we analyze the cross-variant consistency of classic scales (BFI-2 and MBTI original scales) and their variants based on experimental data of human subjects. The experiment used intraclass correlation coefficient (ICC) to evaluate the consistency of different questioning methods in various dimensions of BFI group (extraversion, agreeableness, conscientiousness, neuroticism, openness) and MBTI group (introversion, introversion, feeling and intuition, thinking and feeling, judgment and perception), and the results showed extremely high stability and reliability.

The single measurement ICC of the BFI-2 group ranged from 0.881 to 0.927, and the average ICC ranged from 0.967 to 0.981. The ICC of all dimensions was significant (p < 0.05). Among them, openness (0.927) and conscientiousness (0.921) had the highest cross-variant consistency, while agreeableness (0.881) had a lower consistency but exceeded the good consistency threshold (0.8). The confidence intervals of human ICCs were generally narrow (e.g., BFI Openness dimension [0.894, 0.953]), indicating that the estimates of the results are robust and reliable.

The single-measurement ICCs of the MBTI group ranged from 0.923 to 0.938, and the average ICCs ranged from 0.973 to 0.978. The ICCs of all dimensions were significant (p < 0.05). Among them, Feeling Intuition (0.938) showed the highest consistency, and Judgment Perception (0.923) was the lowest value, but still showed extremely high consistency. The confidence intervals were narrow (e.g., Feeling Intuition [0.907, 0.960]), further verifying the reliability of the results.

The cross-variant consistency of the classic scales and variants in all dimensions of the BFI and MBTI groups of human subjects was at an extremely high level. The ICC values of single measurements ranged from 0.881 to 0.938, and the ICC values of average values ranged from 0.967 to 0.981, both of which were significant (p < 0.05), with tight confidence intervals, indicating the high reliability of the measurement results. The performance of openness and conscientiousness in the BFI group, and of feeling, intuition, and thinking and emotion in the MBTI group were particularly prominent, while agreeableness (0.881) and judgment and perception (0.923), although slightly lower, were still far above the good consistency threshold (0.8). These results show that the measurement results of different questioning methods (classic scales and their variants) in human subjects are almost unaffected by the changes in the expressions, reflecting the high stability of personality measurement.

The extremely high cross-variant consistency shown by human subjects in the BFI group and the MBTI group can be reasonably explained from psychological theories. This phenomenon may reflect the key role of the intrinsic stability of personality traits in measurement. Human traits such as extraversion and openness usually show stability across time and situations. The inherent characteristics of this trait enable human subjects to give similar responses based on internally consistent psychological tendencies when faced with different questioning methods. For example, openness is closely related to creative thinking and curiosity, and its single measurement ICC is as high as 0.927, which may indicate that the stable expression of an individual's innovative tendency is not easily significantly shifted by changes in the form of expression. Personality psychology theory further supports this view, arguing that the main personality traits are relatively resistant to short-term environmental or situational disturbances during individual development, so different scales and variants can capture highly consistent results when measuring these traits.

This consistency may also stem from the unity and ability advantages of humans in cognitive processing. In this study, the single measurement ICC values of the original scale and variant scale ranged from 0.881 to 0.938 on all dimensions of the BFI group and the MBTI group, and the ICC values of the average of multiple measurements ranged from 0.967 to 0.981, showing extremely high reliability and consistency. The average ICC of multiple measurements increased significantly, indicating that the average result further strengthened the stability of the measurement. This result may indicate that human subjects (highly educated people aged 18-30 in this study) have excellent language comprehension and self-awareness, which enables them to accurately grasp the core meaning behind different questions and make consistent responses based on stable internal traits even if the expression is adjusted. For example, the feeling and intuition dimension of MBTI shows an extremely high ICC value (0.938), which may reflect that the stability of individuals' information processing preferences transcends the differences in questioning. This ability of humans may benefit from the semantic processing process, which can extract the deep meaning of the expression, thereby weakening the impact of surface language changes on the measurement results.

The robustness of the scale design may also provide important support for this high consistency. The BFI-2 and MBTI original scales have been verified for a long time, and their variants, although adjusted in expression, are always designed around the core measurement goals. This continuity in design ensures the stability of the measurement results. For example, in the agreeableness dimension (ICC= 0.881), although the consistency is slightly lower than other dimensions, it is still maintained at a high level overall. This shows that the human scale has a strong anti-interference ability when capturing human personality traits, so that the measurement results under different questioning methods can remain highly consistent.

From a broader psychological perspective, this extremely high cross-variant consistency may further reveal the continued stability of individuals in personality dimensions. Although the 16 personality types proposed by the MBTI theory have been criticized by the academic community, many studies have shown that individuals show persistent preferences for specific psychological tendencies. For example, opposing dimensions such as introversion and extroversion, thinking and feeling reflect stable patterns in human cognitive and emotional processing, and these preferences can also be reliably captured by different measurement tools. In this study, the high consistency of introversion and extroversion (ICC is 0.926) and thinking and feeling (ICC is 0.934) may be a reflection of this stability. Combining the results of the Big Five personality dimensions and the MBTI dimensions, the cross-variant consistency of humans shows that no matter how the measurement tools change, their response patterns are always rooted in stable psychological traits and cognitive abilities, which is in sharp contrast to the output characteristics of LLM based on algorithms and contexts.

## 4.5.2 Cross-variant consistency of classic scales and variants in large language model experiments

In this section, we analyze the performance of classic scales (BFI-2 and MBTI original scales) and their variants in cross-variant consistency based on experimental data of four LLMs (including ChatGLM3-6B, Deepseek, GPT-4o, and Llama3.1-8B). The experiment generated the mean and variance of each model in each dimension of the BFI group (extraversion, agreeableness, conscientiousness, neuroticism, openness) and the MBTI group (introversion and extroversion, feeling and intuition, thinking and emotion, judgment and perception) through 100 repeated measurements to evaluate the impact of different questioning methods on LLM personality measurement and analyze the cross-variant consistency of LLM personality. The results show that the cross-variant consistency of LLM is lower than that of humans overall, and there are significant differences between different models and dimensions.

The experiment evaluates the performance of four LLMs on classic scales and their variants, and models with different parameter scales and language backgrounds perform differently.

ChatGLM3-6B is a small-scale model with Chinese language background. The performance of this model in measurement is as follows.

BFI group: the mean range is 30.84-40.49, and the variance range is 1.0871-9.2533. Variant 2 is significantly lower in all dimensions, for example, the mean of the extraversion variant 2 scale is 30.84, the mean of the BFI classic scale is 37.87, and the difference is 7.03; the mean of the agreeableness variant 2 scale is 32.88, and the mean of the BFI classic scale is 38.25, and the difference is 5.37. The variance of the variant 2 scale is generally high, such as 9.2533 for agreeableness and 9.2333 for conscientiousness. Variant 1 and variant 3 are closer to the BFI average (the difference between the variant scale and the original scale mean is between 0.14-2.61), and the variance is lower (for example, the variance of the extraversion dimension of variant 1 is 1.0871).

MBTI group: the mean range is 10.11-15.75, and the variance range is 1.6249-6.3123. The difference between variant 1 and variant 2 and MBTI is small. For example, the mean of the variant introversion and introversion dimension is 13.64, and the mean of the original scale is 10.87, with a difference of 2.77. However, the variance of the variant scale is generally higher than that of the classic scale. For example, the variance of variant 1 of the feeling and intuition dimension is 6.3123, and the variance of the classic scale is 2.5925.

In general, the cross-variant consistency of ChatGLM3-6B is low, and the systematic deviation of variant 2 is particularly prominent, indicating that it is highly sensitive to certain questions.

Deepseek is a large-scale model with Chinese language background. The performance of this model in measurement is as follows.

BFI group: the mean range is 17.00-59.13, and the variance range is 0.9724-9.3452. All variants are close to the average value of the BFI original scale. For example, the average value of the extraversion variant 2 scale is 43.70, and the BFI classic scale is 46.40, with a difference of 2.70; the average value of the variant 3 scale is 52.09, and the difference with the classic scale is 5.69. Overall, the variance of the variant scale fluctuates moderately, such as the variance of the extraversion variant 1 scale is 9.3452, and the variance of the BFI classic scale is 5.8706.

MBTI group: the average value range is 9.52-23.44, and the variance range is 1.2660-3.9681. The difference between the variant and the MBTI original scale is small, for example, the average value of the introversion and extroversion dimension variant 2 scale is 10.75, and the average value of the introversion and extroversion dimension of the MBTI classic scale is 9.82, with a difference of 0.93; the average value of the thinking and feeling dimension variant 2 is 16.20, and the average value of the introversion and extroversion dimension of the MBTI classic scale is 12.0, with a difference of 4.20. The variance of the variant scale is close to the MBTI classic scale, for example, the variance of the judgment perception dimension variant 2 scale is 1.2660, and the variance of the judgment perception dimension of the MBTI classic scale is 2.2821.

Deepseek shows the highest cross-variant consistency, no significant outliers are found, and both the mean and variance show good stability.

GPT-4o is a large-scale model with an English language background. The performance of this model in the measurement is as follows.

BFI group: the mean range is 15.88-58.31, and the variance range is 1.1513-9.4836. Variant 2 is closest to BFI, for example, the mean of the extraversion dimension of variant 2 is 39.62, and the mean of the extraversion dimension of the BFI classic scale is 40.29, with a difference of 0.67. Variant 3 deviates greatly, the mean of the conscientiousness dimension of variant 3 is 48.72, and the mean of the conscientiousness dimension of BFI is 55.12, with a difference of 6.40. The variance of variants fluctuates significantly, such as the variance of the openness dimension variant 3 scale is 9.4836.

MBTI group: the mean range is 9.06-15.87, and the variance range is 2.2761-4.6570. The difference between the variant and MBTI is small, for example, the mean of the introversion dimension variant 1 scale is 11.65, and the mean of the introversion dimension of the MBTI classic scale is 9.06, with a difference of 2.59. The variance of the variant is slightly higher but stable, such as the variance of the feeling and intuition dimension variant 2 scale is 4.6570, and the variance of the feeling and intuition dimension of the MBTI classic scale is 3.4912.

The cross-variant consistency of GPT-4o is moderate, and the MBTI group performs better than the BFI group. The deviation of variant 3 of the BFI group weakens the overall consistency.

Llama3.1-8B is a small-scale model with an English language background. The performance of this model in measurement is as follows.

BFI group: the mean range is 30.78-45.82, and the variance range is 8.8956-14.6208. Variant 2 is closer to BFI, for example, the mean of the extraversion dimension variant 2 scale is 37.29, and the mean of the BFI classic scale extraversion dimension is 35.74, with a difference of 1.55. Variant 3 has the largest deviation, the mean of the conscientiousness dimension variant 3 scale is 45.82, and the mean of the BFI classic scale conscientiousness dimension is 35.48, with a difference of 10.34. The variance of the variants is generally high, such as the variance of the extraversion dimension variant 2 scale is 14.2161.

MBTI group: the mean range is 9.58-13.92, and the variance range is 2.1040-6.5243. Variant 2 is close to MBTI, for example, the mean of Variant 2 scale for introversion and extroversion dimension is 11.15, while the mean of MBTI classic scale for introversion and extroversion dimension is 9.58, with a difference of 1.57. Variant variance is high, for example, the variance of Variant 2 scale for feeling and intuition dimension is 6.5243, while the variance of MBTI classic scale for feeling and intuition dimension is 4.7512.

Llama3.1-8B has low consistency, large variance fluctuation, and Variant 3 performs the worst, showing high instability overall.

It can be seen that the four LLMs show significant differences in cross-variant consistency between the BFI group and the MBTI group, and the model scale and language background may affect the cross-variant consistency of LLM. The high consistency of large-scale models such as Deepseek (Chinese background) (BFI difference 0.17-5.69) may be due to its low randomness and language matching with the Chinese scale. The low consistency (difference 7.03 and 10.34) of small-scale models such as ChatGLM3-6B (Chinese background) and Llama3.1-8B (English background) may be due to insufficient parameters to stably handle variant questions and poor randomness control. The moderate consistency of large-scale models such as GPT-4o (English background) (BFI difference 6.40, MBTI difference 2.59) indicates that its semantic ability can partially compensate for language differences, but excessive flexibility weakens stability. This shows that the relationship between model size and consistency is not linear, but the result of interaction with task matching and language background.

The results of this study show that the personality characteristics of large-scale models and small-scale models are very different, and the specific personality performance is highly related to the parameters and settings of the model itself. Therefore, when evaluating the personality characteristics of LLM, it is necessary to conduct targeted measurement and analysis based on the characteristics of the model, and it cannot be simply regarded as a unified measurement object.

### 4.5.3 Comparison of human subjects and large language model measurement results

In this section, we compared the cross-variant consistency performance of human subjects and four LLMs (ChatGLM3-6B, Deepseek, GPT-4o, Llama3.1-8B) on the BFI-2 and MBTI scales and their variants to explore the impact of different questioning methods (classic scales and their variants) on personality measurement results. The data of human subjects were based on intraclass correlation coefficient (ICC) analysis, while the data of LLM were evaluated by the mean and variance of 100 repeated measurements. This analysis aims to reveal the similarities and differences between human personality and LLM personality in cross-variant consistency in personality measurement, and explore the potential mechanisms behind these differences.

#### 4.5.3.1 Comparison of cross-variant consistency

The experimental results show that the cross-variant consistency of human subjects is significantly higher than that of LLM, and the rules are highly unified.

Human subjects showed extremely high cross-variant consistency in all dimensions of the BFI group and the MBTI group, indicating that changes in the questioning method have little effect on their measurement results. The single measurement ICC value range of the BFI group is 0.881-0.927, and the average ICC range is 0.967-0.981; the single measurement ICC value range of the MBTI group is 0.923-0.938, and the average ICC range is 0.973-0.978. All results are significant and the confidence intervals are compact. This shows that human responses under different questioning methods are highly stable.

In sharp contrast, the cross-variant consistency performance of the four LLMs is generally lower than that of humans, and their personality measurement results are significantly affected by the questioning method. Relatively speaking, Deepseek showed relatively high stability (BFI group mean difference range 0.17-5.69, variance range 0.9724-9.3452; MBTI group difference range 0.93-4.20, variance range 1.2660-3.9681), but it was still far below the human level. GPT-4o's BFI group consistency was moderate (difference range 0.67-6.40, variance range 1.1513-9.4836), and the MBTI group was slightly better (differences were all below 2.59, variance range 2.2761-4.6570). ChatGLM3-6B and Llama3.1-8B had low cross-variant consistency, with the former's BFI group variant 2 mean deviation being significant (difference as high as 7.03), and the latter's variance fluctuating significantly (for example, the variance of extraversion variant 2 was 14.2161). These results show that the measurement results of LLM are highly sensitive to changes in the way of asking questions, and the parameters and language background of the model itself affect the performance of personality measurement.

The experimental results support H1, and the impact of the way of asking questions on humans and LLM is different.

### 4.5.3.2 Analysis of potential mechanisms of differences

The difference in cross-variant consistency between humans and LLM may reflect the fundamental differences in their response mechanisms and cognitive processing methods.

1. Differences in response mechanisms: intrinsic traits and algorithm generation

The high consistency of humans in cross-variant measurements reflects their stable response patterns based on intrinsic personality traits. Personality psychology research shows that human personality traits are highly stable at different times and situations.

This stability enables individuals to rely on inherent psychological tendencies, maintain consistent performance, and give consistent responses when faced with different questions or measurement methods. This also reflects the intrinsic regulation and stable expression of self-traits by humans during the socialization process.

In contrast, the lower consistency of LLM indicates that it lacks a stable response mechanism based on intrinsic traits, and its personality output is more the result of algorithm generation, which depends on the statistical distribution of training data, parameter settings, and algorithm generation mechanisms. LLM's personality expression is more of a data-driven result, an unstable trait expression, and is easily affected by specific questions, prompt design, and even random seed changes. For example, the significant deviation of ChatGLM3-6B on BFI group variant 2 (the difference between extraversion dimension variant 2 and the original scale is 7.03) may be due to its semantic understanding bias of specific questions rather than stable trait expression.

2. Differences in response mechanisms: unity of cognitive processing and context dependence

Human cognitive processing is highly unified and can maintain the stability of measurement results through language understanding and self-reflection. Humans have strong self-awareness and deep semantic processing capabilities, which enable them to transcend surface expression differences and extract the deep core meaning of questions, making humans less sensitive to changes in questions.

The consistency of LLM's cognitive processing is limited by context dependence rather than internal unity. The output of LLM is highly dependent on input prompt design, question expression and random seeds, and the cognitive processing results are also affected by the corpus, resulting in significant fluctuations in measurement results. LLM's response is easily affected by context details and lacks the unified adjustment ability of humans during processing. This context dependence may be due to the generation mechanism of LLM based on statistical probability rather than the deep processing of humans based on self-cognition.

3. The impact of scale design: robustness and sensitivity

The BFI-2 and MBTI scales are designed for human psychological structures. They are long-term verified personality scales with good reliability and validity. Even if the expression of scale items is changed, they can still convey the core measurement goals and capture the core personality traits of humans.

However, the same scale design has a completely different impact on LLM, and LLM shows higher sensitivity to changes in questioning. This shows that LLM's answers may be more affected by semantic parsing methods and contextual cues rather than a deep understanding of the core meaning of the question, and the robustness of human scales may not be true for LLM.

There are also differences in the cross-variant consistency of LLMs of different scales and language backgrounds, which reflects that the model architecture, parameter quantity, training data and language adaptability will affect the personality measurement results of LLM. These differences suggest that when evaluating the personality performance of LLM, the measurement method needs to be adjusted according to the specific model characteristics, rather than uniformly applying human scales.

## 4.5.4 Inspiration of cross-variant consistency for building personality theory for large language models

This study compared the cross-variant consistency of humans and four LLMs (including ChatGLM3-6B, Deepseek, GPT-4o, and Llama3.1-8B) on the BFI-2 and MBTI scales and their variants. The results showed that humans had extremely high cross-variant consistency, while LLMs had lower consistency and significant differences between models. This shows the difference between the internal mechanisms of LLMs and humans, and provides inspiration for building LLM personality measurement theory.

The variants used in this study reflect different psychological mechanisms, which can be explained by the perspective of social psychology on the formation of self-schemas. Social psychology believes that people can form self-schemas in a variety of ways, including comparing with others, self-inference, and receiving feedback from others [88]. Among them, social comparison theory explains how people know themselves by comparing with others [89]; self-perception theory explains the process by which individuals infer self-characteristics by observing their own behavior [90]; social interaction theory emphasizes that individuals gradually form self-concepts through interaction and feedback with others [91]. Variant 1 requires subjects to compare their own behavior with that of others, which is a reflection of social comparison theory. Variant 2 requires subjects to evaluate whether the outside world's description of themselves is accurate, which is a reflection of self-perception theory.

Variant 3 requires subjects to choose appropriate options to fill in sentences, showing the impact of self-schema on subjects' behavior. Humans show extremely high consistency across variants in experiments, which is precisely the embodiment of humans' stable response patterns across time and context. LLM does not have such a stable internal structure.

In contrast, the output of LLM mainly depends on training data, parameter settings, and prompt design. Its personality performance is not derived from internal stable traits, but more like a data-driven generation result. Therefore, the performance of LLM on different variants reflects its high sensitivity to context and language cues and the randomness in the generation process. For example, if LLM performs relatively stably under variant 1 (social comparison), it may indicate that it can better match the patterns in the training data when dealing with comparative questions; if it performs poorly under variant 2 (self-inference), it may indicate that its evaluation of the accuracy of self-description is easily affected by contextual changes; and if it performs differently under variant 3 (social interaction), it may reflect that LLM is insufficient in its ability to simulate human feedback mechanisms and fails to form a stable self-schema similar to that of humans.

This phenomenon shows that the internal mechanism of LLM is significantly different from that of humans. Human personality theory and scales designed to capture stable internal traits of humans in human personality research are not fully applicable to LLM. Human personality expression relies on internal traits and socialization experience accumulated over a long period of time, while LLM personality is more like a distributed, external input-dependent generation process, and its output mainly reflects the distribution of training data and the context dependence of prompts. Therefore, when constructing a personality theory for large language models, the characteristics of LLM itself should be fully considered, and the concept of distributed personality model should be adopted to describe the personality characteristics of LLM through in-depth characterization of probability distribution and context sensitivity, rather than simply directly applying the theory and measurement methods of human personality.

## 4.6 Conclusion

This chapter compares the performance of humans and LLM under the classic personality scale and its variants, and finds that the two have significant differences in cross-variant consistency. The results show that humans can maintain a high degree of stability and consistency under different questioning methods, indicating that human personality can show a high degree of stability and consistency in different contexts, which reflects the inherent coherence of human personality structure. Unlike humans, the output of LLM is highly sensitive to context, prompting methods and training data distribution, and its personality has obvious volatility and context dependence. This result reveals the mechanism of LLM personality and shows the difference between LLM personality and human personality in cognitive sources and generation logic.

These findings not only provide a theoretical basis for the future construction of distributed personality theory and measurement methods for LLM, but also provide methodological support for humans to more accurately understand, evaluate and apply artificial intelligence systems. By distinguishing the cross-variant consistency between LLM personality and human personality, researchers and users can more rationally set the boundaries of LLM use, thereby improving the effectiveness of human-computer interaction, reducing trust bias and ethical risks caused by misunderstanding of LLM human nature, and better promoting the scientific, prudent and sustainable development of artificial intelligence in human society.

# Chapter 5 A comparative study of role-playing and personality trait retention between large language models and humans

Trait activation theory points out that an individual's behavior is affected by the situation and role, and may exhibit behaviors inconsistent with his or her stable personality traits in a specific role [47]. Against the background of the rapid development of large language models (LLMs), the personality performance of LLMs in role-playing has attracted widespread attention. However, it is still unclear whether LLMs will exhibit stable personality traits in different role-playing situations, or whether their personality performance will be affected by baseline personality. This study conducted an exploratory analysis of the personality trait retention of humans and LLMs in role-playing, aiming to provide a preliminary empirical research reference for the role-playing characteristics of LLMs, lay a foundation for subsequent in-depth exploration of the stability and plasticity of LLM personality, and provide a new perspective for understanding the personality shaping mechanism of LLMs. This study will help humans improve the accuracy and adaptability of LLM role-playing applications and improve the matching and effectiveness of human-computer collaboration.

## 5.1 Research Purpose and Hypothesis

Research Purpose: To compare the measurement results of the classic personality scale for LLM and human subjects under different role-playing, and to reveal the retention of personality characteristics of LLM and human subjects in active role-playing.

Research Hypothesis:

H0: The personality measurement scores of LLM and human subjects in role-playing are not significantly correlated with their baseline personality.

Expected Results: For both LLM and human subjects, their role-playing scores are not systematically related to their baseline personality scores, and the relevant statistical results are not significant.

H1: The personality measurement scores of LLM in role-playing are not significantly correlated with their baseline personality, while the role-playing scores of human subjects are significantly correlated with their baseline personality.

Expected Results: The correlation between LLM role-playing scores and baseline personality is low or irrelevant, and the human role-playing scores show a certain statistical correlation with the baseline personality scores, which may be regulated by the role type.

## 5.2 Research Methods

The core of this chapter is to compare the retention of personality traits between LLM and humans in role-playing situations, in order to explore the impact of LLM and human original personality (baseline personality) on role-playing personality. This study not only has practical application value, but also helps to optimize the performance of LLM in role-playing tasks, but also helps to understand the differences between LLM personality and human personality more deeply, and provides a basis for building personality theories and measurement methods for LLM. This study is an exploratory study. The key to the experiment is to reasonably design role-playing tasks and scientifically select analysis methods to ensure that the experimental results can accurately reflect the impact of original personality on role-playing personality.

### 5.2.1 Role-playing design

The design of role-playing tasks is the core of this experiment. In order to ensure the representativeness, controllability and concentration of the results of the study, it is necessary to select specific personality traits within a limited range, that is, to clarify the core dimension of the study. In view of the diversity of personality traits, comprehensive coverage of all dimensions (such as the Big Five personality) will lead to too wide a research scope and scattered results. Therefore, this study chooses the introversion dimension as the research focus. This choice is based on the following considerations: introversion is one of the most representative and widely recognized dimensions in personality research, closely related to social interaction, and easy to activate in role-playing situations, with high research feasibility.

In the introversion dimension, different levels need to be divided when designing roles to meet the needs of the experiment. Personality is continuous, and personality traits naturally have two limits. In the introversion dimension, it can be described as "extremely extroverted" and "extremely introverted". However, in order to avoid overly extreme expressions and leave more room for interpretation for the subjects, this study sets two roles of "very introverted" and "very extroverted" at the extreme values, which can both point to the extreme values and give the subjects a certain degree of flexibility.

In order to provide a richer comparison level on the introversion and extroversion dimension, a transition role is set in the middle of this dimension. The design of the transition role mainly considers the following factors: first, the selected role should have significant differences in the introversion and extroversion dimension to ensure the effectiveness of the experimental manipulation; second, the role should be well-known and preferably have cross-cultural influence to reduce the understanding bias caused by cultural background differences. Taking the above factors into consideration, this study finally selected Lin Daiyu (introverted) and Sun Wukong (extroverted) as intermediate-level transition roles to provide a more natural level change between "very introverted" and "very extroverted".

In order to ensure that the human subjects really enter the role and control factors such as the LLM input environment that may affect the experimental results, this study provides clear role-playing instructions and reserves preparation time for human subjects to enter the role to enhance the effectiveness of role-playing.

## 5.2.2 Statistical analysis methods

This chapter aims to compare the impact of baseline personality on role-playing personality, so the baseline personality measurement should be recorded first. The scores of the introversion and extroversion scale of human subjects in a natural state without role-playing are used as their baseline personality scores; the scores of the introversion and extroversion scale of LLM under the condition of no role-playing instructions are used as their baseline personality scores. The baseline personality reflects the introversion and extroversion traits of the subjects in a natural state, and provides a benchmark for the subsequent evaluation of changes in role-playing personality.

This chapter focuses on whether the baseline personality has an impact on the role-playing personality, rather than the specific influence rules. Therefore, the statistical analysis method will focus on directly comparing the differences between the role-playing personality and the baseline personality data to test the retention of personality traits in the role-playing task. For this, the independent sample t-test can be used to analyze the differences between the baseline personality and the role-playing personality of humans and LLMs respectively.

## 5.2.3 Impact of the results of role-playing and personality trait retention

The interpretation of the analysis results is directly related to the hypothesis verification. If the t-test shows that the differences between the baseline personality scores and the role-playing scores of LLM and humans are not significant, then H0 is supported, that is, both humans and LLMs have significant influences on their personality measurement scores by baseline personality, which indicates that humans and LLMs have certain similarities in role-playing tasks, and LLMs' personality may also have certain stability. If the influence of LLMs' baseline personality on role-playing personality shows different patterns from that of humans, that is, H1 is supported, then it indicates that there are fundamental differences between LLMs' personality structure and that of humans. LLMs may be more easily affected by role-playing situations, have weaker ability to retain baseline traits, and have more unstable personality, with obvious dynamics and situational dependence. This difference suggests that human personality theories based on the premise that personality is stable may not be applicable to LLMs, and dynamic models (such as theories based on probability distribution or algorithmic features) may need to be introduced to describe LLMs' personality performance. In personality research on LLMs, the output of LLMs needs to be regarded as a function of training data and prompts, rather than a reflection of fixed traits. This new theory not only enriches the boundaries of personality research, but also provides a theoretical basis for optimizing the consistency of LLM in role-playing tasks.

## 5.3 Reference group experimental design

### 5.3.1 Experimental ethics and settings

This chapter uses the same experimental settings as Chapter 3. Ethical considerations also follow the framework of Chapter 3, including the principles of informed consent, data anonymization, and voluntary participation.

## 5.3.2 Experimental Design

I. Overall Experimental Design

Given the complexity of roles in real life and the multidimensionality of personality traits, this study focuses on the introversion dimension and studies the personality retention of LLM and humans in role-playing.

The experiment adopts a single-factor within-subject design, and the independent variable is the role type, which is divided into four levels: very introverted, very extroverted, Lin Daiyu (relatively introverted) and Sun Wukong (relatively extroverted). Among them, "very introverted" and "very extroverted" are the extreme values of the introversion dimension, and Lin Daiyu and Sun Wukong are the middle values of the introversion dimension. The specific degree is freely interpreted by the subjects to increase the ecological validity of the experiment. The dependent variable is the difference between the role-playing personality and the original personality (baseline personality), which is directly described by the results of the independent sample t-test. Among them, the original personality (baseline personality) is defined as the BFI-2 and MBTI measurement results when there is no role-playing.

The role-playing experiment lasts for four days. Each subject plays a designated role every day and completes the personality scale within the specified time.

2. Design of role-playing instructions

In the role-playing experiment, the instructions for role-playing are expressed to the subjects through instructions. In order to avoid experimental errors caused by different instructions when playing different roles, this experiment standardizes the instructions. The instructions for different roles are as follows.

Lin Daiyu:

Hello! Welcome to participate in this experiment! In the following process, please play the role of [Lin Daiyu]. Please try to fully integrate into the role of Lin Daiyu based on your understanding of the role, and then complete the following test as Lin Daiyu.

Sun Wukong:

Hello! Welcome to participate in this experiment! In the following process, please play the role of [Sun Wukong]. Please try to fully integrate into the role of Sun Wukong based on your understanding of the role, and then complete the following test as Sun Wukong.

Very extroverted:

Hello! Welcome to participate in this experiment! In the following process, please play the role of [a very extroverted person]. Please try to fully integrate into this role based on your understanding of the role, and then complete the following test as a very extroverted person.

Very introverted:

Hello! Welcome to this experiment! In the following process, please play the role of [a very introverted person]. Please try to fully integrate into this role based on your own understanding of the role, and then complete the following test as a very introverted person.

## 5.3.3 Subjects

Same as Chapter 3. This study actually recruited 60 subjects offline. The subjects were aged between 18 and 30, all of whom had a bachelor's degree or above and had good language comprehension and experimental operation skills. There were 25 males and 35 females, with no gender restrictions. The key information of the subjects, such as gender and age, is shown in Table 3-1.

## 5.3.4 Experimental Materials and Procedures

I. Experimental Materials

BFI-2 Chinese version and its variants, MBTI Chinese version (93 questions) and its variants, a total of 7 scales.

II. Experimental Procedures

As in Chapter 3, this study is divided into three stages: initial pilot stage, mid-term verification stage, and large-scale implementation stage.

Each experiment in each stage takes four days. During the experiment, role-playing tasks and personality scales are released at regular intervals every day. Subjects are requested to complete the role-playing and personality tests at their own discretion.

## 5.3.5 Data processing

After the subjects completed the role-playing and personality measurement, the personality scale data of the subjects were collected and sorted. After the collection and sorting were completed, the subject data were checked and cleaned.

This study used independent sample t-test to test the difference between the role-playing results and baseline personality of the introverted group and the extroverted group, and to test the influence of the subjects' baseline personality on the role-playing results.

This study analyzed the maximum and minimum degree of deviation of the role-playing personality from the baseline personality in the role-playing task for the introverted group and the extroverted group. The minimum degree of deviation of the introverted group from the baseline personality was defined as the difference between the Lin Daiyu role played in the role-playing task and the baseline personality. The maximum degree of deviation of the introverted group from the baseline personality was defined as the difference between the very extroverted role played in the role-playing task and the baseline personality. The minimum degree of deviation of the extroverted group from the baseline personality was defined as the difference between the Sun Wukong role played in the role-playing task and the baseline personality. The maximum degree of deviation of the introverted group from the baseline personality was defined as the difference between the very introverted role played in the role-playing task and the baseline personality. Independent sample t-test was used to examine the difference in the degree of deviation from baseline personality between the introvert group and the extrovert group during role-playing.

## 5.3.6 Results

In the role-playing task, the independent sample t-test results of the difference between the role personality played by the introvert group and the extrovert group and their own baseline personality are as follows.

Statistical results of the BFI-2 group:

t-test Mean difference Standard error 95% confidence interval Significance (two-tailed)

Table 5-1 Independent sample t-test results

|  | Levene's Test | t-test | MD | SE | 95%CI | P |
| --- | --- | --- | --- | --- | --- | --- |
| BFI | F=6.863, P<0.05 | t=1.244 | 2.6667 | 2.1434 | [-1.6504, 6.9837] | P>0.05 |

Table 5-1 Independent sample t-test results

|  | Levene's Test | t-test | MD | SE | 95%CI | P |
|---|---|---|---|---|---|---|
| Variant 1 | F=3.931, P>0.05 | t=1.034 | 1.9667 | 1.9025 | [-1.8415, 5.7749] | P>0.05 |
| Variant 2 | F=2.481, P>0.05 | t=1.256 | 2.3000 | 1.8310 | [-1.3652, 5.9652] | P>0.05 |
| Variant 3 | F=2.762, P>0.05 | t=1.315 | 2.4000 | 1.8256 | [-1.2543, 6.0543] | P>0.05 |

The t-test and 95% confidence interval results in Table 5-1 have been adjusted according to the variance homogeneity test results. As shown in Table 5-1, the p-values of the t-tests for all scales are greater than 0.05 (0.220, 0.306, 0.214, 0.194, respectively), which means that there is not enough evidence to reject the null hypothesis, and there is no significant mean difference between these variables, that is, it cannot be considered that the role of Lin Daiyu played by the subjects in the introvert group and the extrovert group is significantly different.

Table 5-2 Independent sample t-test results

|  | Levene's Test | t-test | MD | SE | 95%CI | P |
|---|---|---|---|---|---|---|
| BFI | F=0.998, P>0.05 | t=0.435 | 0.5667 | 1.3015 | [-2.0386, 3.1720] | P>0.05 |
| Variant 1 | F=3.425, P>0.05 | t=-0.137 | -0.1667 | 1.2198 | [-2.6083, 2.2750] | P>0.05 |
| Variant 2 | F=0.556, P>0.05 | t=0.051 | 0.0667 | 1.3199 | [-2.5754, 2.7087] | P>0.05 |
| Variant 3 | F=0.016, P>0.05 | t=0.358 | 0.4333 | 1.2092 | [-1.9872, 2.8539] | P>0.05 |

The t-test and 95% confidence interval results in Table 5-2 have been adjusted according to the variance homogeneity test results. As shown in Table 5-2, the p-values of the t-tests for all scales are greater than 0.05, which means that there is no significant difference in means between the groups, and there is no significant difference in the Monkey King role played by the introverted and extroverted groups, and the baseline personality has no effect on the role-playing personality.

Table 5-3 Independent sample t-test results

|  | Levene's Test | t-test | MD | SE | 95%CI | P |
|---|---|---|---|---|---|---|
| BFI | F=0.043, P>0.05 | t=1.924 | 2.8000 | 1.4551 | [-0.1126, 5.7126] | P>0.05 |
| Variant 1 | F=0.477, P>0.05 | t=1.588 | 2.2667 | 1.4274 | [-0.5906, 5.1240] | P>0.05 |
| Variant 2 | F=0.033, P>0.05 | t=1.839 | 2.8333 | 1.5409 | [-0.2512, 5.9178] | P>0.05 |
| Variant 3 | F=0.417, P>0.05 | t=0.328 | 0.4667 | 1.4213 | [-2.3784, 3.3117] | P>0.05 |

The t-test and 95% confidence interval results in Table 5-3 have been adjusted according to the variance homogeneity test results. As shown in Table 5-3, the p-values of the t-tests for all scales are greater than 0.05, there are no significant mean differences between the groups, and there is no significant difference in the introverted role played by the subjects in the introverted and extroverted groups.

Table 5-4 Independent sample t-test results

|  | Levene's Test | t-test | MD | SE | 95%CI | P |
|---|---|---|---|---|---|---|
| BFI | F=0.197, P>0.05 | t=-0.328 | -0.5000 | 1.5263 | [-3.5553, 2.5553] | P>0.05 |

Table 5-4 Independent sample t-test results

|  | Levene's Test | t-test | MD | SE | 95%CI | P |
|---|---|---|---|---|---|---|
| Variant 1 | F=0.008, P>0.05 | t=-0.604 | -0.8667 | 1.4346 | [-3.7383, 2.0049] | P>0.05 |
| Variant 2 | F=0.540, P>0.05 | t=-0.757 | -1.0000 | 1.3205 | [-3.6432, 1.6432] | P>0.05 |
| Variant 3 | F=0.002, P>0.05 | t=-0.765 | -1.1000 | 1.4376 | [-3.9777, 1.7777] | P>0.05 |

The t-test and 95% confidence interval results in Table 5-4 have been adjusted according to the variance homogeneity test results. As shown in Table 5-3, the p-values of the t-tests of all scales are greater than 0.05, there are no significant mean differences between the groups, and there is no significant difference in the extroverted roles played by the introvert and extrovert groups.

MBTI group statistical results:

Table 5-5 Independent sample t-test results

|  | Levene's Test | t-test | MD | SE | 95%CI | P |
|---|---|---|---|---|---|---|
| MBTI | F=30.175, P<0.05 | t=2.669 | 2.7333 | 1.0240 | [0.6478, 4.8189] | P<0.05 |
| Variant 1 | F=19.646, P<0.05 | t=2.282 | 1.8000 | 0.7887 | [0.1995, 3.4005] | P<0.05 |
| Variant 2 | F=11.439, P<0.05 | t=2.325 | 2.2000 | 0.9463 | [0.2879, 4.1121] | P<0.05 |

The t-test and 95% confidence interval results in Table 5-5 have been adjusted according to the results of the variance homogeneity test. As shown in Table 5-5, the p-values of the t-tests of all scales are less than 0.05, there are significant mean differences between the groups, and there are significant differences between the introverted and extroverted groups when playing the same role of Lin Daiyu.

The results of MBTI and BFI are significantly different, which may be due to a variety of factors, such as the differences in the design of the MBTI and BFI scales, and the differences in the personality traits of the introverted and extroverted groups.

Table 5-6 Independent sample t-test results

|  | Levene's Test | t-test | MD | SE | 95%CI | P |
|---|---|---|---|---|---|---|
| MBTI | F=22.521, P<0.05 | t=-2.464 | -1.5667 | 0.6359 | [-2.8562, -0.2772] | P<0.05 |
| Variant 1 | F=3.822, P>0.05 | t=-1.311 | -1.0667 | 0.8136 | [-2.6952, 0.5619] | P>0.05 |
| Variant 2 | F=24.344, P<0.05 | t=-2.772 | -2.0667 | 0.7455 | [-3.5807, -0.5527] | P<0.05 |

The t-test and 95% confidence interval results in Table 5-6 have been adjusted according to the variance homogeneity test results. As shown in Table 5-6, except for variant 1, the p-values of the t-tests of all scales are less than 0.05, and there are significant mean differences between the groups. There are significant differences between the introverted and extroverted groups when they play the same role, Sun Wukong.

Table 5-7 Independent sample t-test results

|  | Levene's Test | t-test | MD | SE | 95%CI | P |
|---|---|---|---|---|---|---|
| MBTI | F=15.333, P<0.05 | t=2.191 | 1.5667 | 0.7151 | [0.1084, 3.0249] | P<0.05 |
| Variant 1 | F=19.801, P<0.05 | t=2.770 | 1.9333 | 0.6980 | [0.5196, 3.3471] | P<0.05 |
| Variant 2 | F=13.921, P<0.05 | t=2.397 | 2.0000 | 0.8344 | [0.3042, 3.6958] | P<0.05 |

The t-test and 95% confidence interval results in Table 5-7 have been adjusted according to the variance homogeneity test results. As shown in Table 5-7, the p-values of the t-tests for all scales are less than 0.05, and there are significant mean differences between the groups. There are significant differences between the introverted and extroverted groups when playing the same role of being very introverted.

Table 5-8 Independent sample t-test results

|  | Levene's Test | t-test | MD | SE | 95%CI | P |
|---|---|---|---|---|---|---|
| MBTI | F=11.338, P<0.05 | t=-1.517 | -0.8000 | 0.5274 | [-1.8625, 0.2625] | P>0.05 |
| Variant 1 | F=8.907, P<0.05 | t=-1.404 | -0.8667 | 0.6174 | [-2.1179, 0.3845] | P>0.05 |
| Variant 2 | F=10.265, P<0.05 | t=-0.983 | -0.6000 | 0.6106 | [-1.8337, 0.6337] | P>0.05 |

The t-test and 95% confidence interval results in Table 5-8 have been adjusted according to the results of the variance homogeneity test. As can be seen from Table 5-8, the p-values of the t-tests of all scales are greater than 0.05, there is no significant mean difference between the groups, and there is no significant difference between the subjects in the introvert group and the extrovert group when they play the same role of extroversion.

Combining the results of the BFI group and the MBTI group, it can be found that the baseline personality of human beings will have a certain impact on the role-playing personality, and it is difficult for humans to completely get rid of the influence of their original personality.

The independent sample t-test results of the degree to which the personality of the role played by the extrovert group and the introvert group deviates from the baseline personality when playing roles.

Statistical results of the BFI group:

Table 5-9 Independent sample t-test results of the deviation degree of BFI-2

|  | Levene's Test | t-test | MD | SE | 95%CI | P |
|---|---|---|---|---|---|---|
| Minimum Amplitude | F=0.238, P>0.05 | t=3.601 | 6.4253 | 1.7844 | [2.8508,9.9999] | P<0.05 |
| Maximum Amplitude | F=0.155, P>0.05 | t=-2.753 | -4.4241 | 1.6067 | [-7.6428, -1.2054] | P<0.05 |

The t-test and 95% confidence interval results in Tables 5-9 have been adjusted according to the results of the variance homogeneity test. The study used an independent sample t-test to compare the extent of deviation from baseline personality between the introvert group and the extrovert group during role-playing, where the "minimum extent" and "maximum extent" respectively reflect the lowest and highest extent of deviation from the baseline personality of the subjects during role-playing. The p values of the t-test were all less than 0.05, and the test results showed that there were significant differences in the deviation from baseline personality between the introvert group and the extrovert group during role-playing. The mean difference of the minimum extent was positive, and the direction of the difference was positive, while the mean difference of the maximum extent was negative, and the direction of the difference was negative.

Table 5-10 Independent sample t-test results of the deviation degree of BFI variant 1

|  | Levene's Test | t-test | MD | SE | 95%CI | P |
|---|---|---|---|---|---|---|
| Minimum Amplitude | F=0.197, P>0.05 | t=3.642 | 7.1171 | 1.9544 | [3.2019, 11.0323] | P<0.05 |
| Maximum Amplitude | F=1.192, P>0.05 | t=-2.471 | -5.0167 | 2.0304 | [-9.0842, -0.9493] | P<0.05 |

The t-test and 95% confidence interval results in Table 5-10 have been adjusted according to the variance homogeneity test results. As can be seen from Table 5-10, the p-values of the t-test are all less than 0.05. The test results show that there are significant differences in the performance of baseline personality deviations between the introverted group and the extroverted group in role-playing. The smallest mean difference is positive and the difference direction is positive, while the largest mean difference is negative and the difference direction is negative.

Table 5-11 Independent sample t-test results of the deviation degree of BFI variant 2

| | Levene's Test | t-test | MD | SE | 95%CI | P |
|---|---|---|---|---|---|---|
| Minimum Amplitude | F=0.171, P>0.05 | t=3.090 | 6.1529 | 1.9915 | [2.1635, 10.1423] | P<0.05 |
| Maximum Amplitude | F=0.982, P>0.05 | t=2.735 | -5.3584 | 1.9593 | [-9.2833, -1.4335] | P<0.05 |

The t-test and 95% confidence interval results in Table 5-11 have been adjusted according to the variance homogeneity test results. As can be seen from Table 5-11, the p-values of the t-test are all less than 0.05. The test results show that there are significant differences in the performance of baseline personality deviations between the introverted group and the extroverted group in role-playing. The smallest mean difference is positive and the direction of the difference is positive, while the largest mean difference is negative and the direction of the difference is negative.

Table 5-12 Independent sample t-test results of the deviation degree of BFI variant 2

| | Levene's Test | t-test | MD | SE | 95%CI | P |
|---|---|---|---|---|---|---|
| Minimum Amplitude | F=0.006, P>0.05 | t=4.043 | 6.9630 | 1.7222 | [3.5129,10.4130] | P<0.05 |
| Maximum Amplitude | F=3.452, P>0.05 | t=2.173 | -4.3393 | 1.9969 | [-8.3397, -0.3389] | P<0.05 |

The t-test and 95% confidence interval results in Table 5-12 have been adjusted according to the results of the variance homogeneity test. As can be seen from Table 5-12, the p-values of the t-test are all less than 0.05. The test results show that there are significant differences in the performance of the introverted group and the extroverted group in the deviation from the baseline personality in role-playing. Among them, the smallest mean difference is positive, the difference direction is positive, and the largest mean difference is negative, and the difference direction is negative.

As can be seen from Tables 5-9, 5-10, 5-11, and 5-12, the baseline personality affects the degree to which the personality of the role played by the subjects deviates from the baseline personality in the role-playing task, which further indicates that the baseline personality of the subjects will affect the role-playing task of the subjects.

MBTI group statistical results:

Table 5-13 Independent sample t-test results of the deviation degree of MBTI

| | Levene's Test | t-test | MD | SE | 95%CI | P |
|---|---|---|---|---|---|---|
| Minimum Amplitude | F=15.548, P<0.05 | t=4.405 | 6.6667 | 1.5135 | [3.6370, 9.6963] | P<0.05 |
| Maximum Amplitude | F=7.445，P<0.05 | t=-11.667 | -11.1000 | 0.9514 | [-13.0044, -9.1956] | P<0.05 |

The t-test and 95% confidence interval results in Table 5-13 have been adjusted according to the variance homogeneity test results. As shown in Table 5-13, the p-values of the t-test are all less than 0.05, and the test results show that there are significant differences in the performance of baseline personality deviations between the introvert group and the extrovert group in role-playing.

Table 5-14 Independent sample t-test results of the deviation degree of MBTI variant 1

| | Levene's Test | t-test | MD | SE | 95%CI | P |
|---|---|---|---|---|---|---|
| Minimum Amplitude | F=9.456, P<0.05 | t=2.529 | 2.400 | 0.949 | [0.489, 4.311] | P<0.05 |
| Maximum Amplitude | F=6.294，P<0.05 | t=-5.070 | -5.867 | 1.157 | [-8.192, -3.542] | P<0.05 |

The t-test and 95% confidence interval results in Table 5-14 have been adjusted according to the variance homogeneity test results. The p values of the t-test are all less than 0.05, and there are significant differences in the performance of baseline personality deviations between the introverted group and the extroverted group in role-playing. The smallest difference is positive, and the largest difference is negative.

Table 5-15 Independent sample t-test results of the deviation degree of MBTI variant 2

| | Levene's Test | t-test | MD | SE | 95%CI | P |
|---|---|---|---|---|---|---|
| Minimum Amplitude | F=8.776, P<0.05 | t=2.057 | 2.400 | 1.167 | [0.054, 4.746] | P<0.05 |
| Maximum Amplitude | F=5.654, P<0.05 | t=-4.570 | -5.900 | 1.291 | [-8.493, -3.307] | P<0.05 |

The t-test and 95% confidence interval results in Table 5-15 have been adjusted according to the results of the variance homogeneity test. As can be seen from Table 5-15, the p-values of the t-test are all less than 0.05. The test results show that there are significant differences in the performance of the introverted group and the extroverted group in the deviation from the baseline personality in role-playing. Among them, the smallest mean difference is positive, the difference direction is positive, and the largest mean difference is negative, and the difference direction is negative.

Combining the results of the BFI group and the MBTI group, it can be found that the original baseline personality of human beings has a significant effect on the degree to which the role personality deviates from the baseline personality in the role-playing task.

## 5.4 Experimental design of the experimental group

### 5.4.1 Experimental design

The experimental design of the experimental group was consistent with that of the reference group, using the same role-playing instructions, and asked LLM to play Lin Daiyu, Sun Wukong, a very introverted person, and a very extroverted person respectively.

## 5.4.2 Subjects

The LLM subjects selected in this study are the same as those in Chapter 3. The study selected ChatGLM3-6B (medium-sized model, 600 million parameters, Chinese background), Deepseek (small model, fewer parameters, Chinese background), GPT-4o (large model, the number of parameters is at the forefront, English background) and Llama3.1-8B (medium-sized model, 800 million parameters, English background). The key information of the four selected LLMs is shown in Table 3-4.

## 5.4.3 Experimental materials and procedures

I. Experimental materials

BFI original scale and its variant scales and MBTI scale and its variant scales, a total of 7 scales.

II. Experimental procedures

As in Chapter 3, the experiment is divided into two stages: experimental preparation stage and data generation stage.

LLM completed 10 personality tests under each role setting.

## 5.4.4 Data processing

The scores of each scale of each model were recorded to form a complete data set, and the independent sample t test was used to test the differences between the role-playing effects of different models.

## 5.4.5 Results

Independent sample t-test results of GPT and Deepseek when playing Lin Daiyu:

Table 5-16 Independent sample t results of Lin Daiyu's personality difference with baseline

| Model | M | SD | t-Test | Welch'sT | MD | d |
|-------|--------|-------|----------|-----------|-------|-------|
| GPT | 23.694 | 2.559 | t=4.556, | T=4.556, | | |
| Deepseek | 25.776 | 1.918 | P=0.000 | P=0.000 | 2.082 | 0.921 |
| Total | 24.735 | 2.481 | | | | |

Independent sample T test results of GPT and Deepseek when playing as Sun Wukong:

Table 5-16 Independent sample t results of Sun 's personality difference with baseline

| Model | M | SD | t-Test | Welch'sT | MD | d |
|-------|-----|-----|--------|----------|-----|-----|
| GPT | 56.163 | 1.625 | t=12.412 | T=12.412 | 3.286 | 2.508 |
| Deepseek | 59.449 | 0.891 | P=0.000 | P=0.000 | | |
| Total | 57.806 | 2.104 | | | | |

Independent sample t-test results of GPT and Deepseek when playing the role of being very introverted:

Table 5-18 Independent sample t-test results

| Model | M | SD | t-Test | Welch'sT | MD | d |
|-------|-----|-----|--------|----------|-----|-----|
| GPT | 18.735 | 1.303 | t=4.335, | T=4.335, | 1.306 | 0.876 |
| Deepseek | 17.429 | 1.685 | P=0.000 | P=0.000 | | |
| Total | 18.082 | 1.662 | | | | |

In Table 5-16, Table 5-17, and Table 5-18, the independent sample t test and Welch's T test are all significant, indicating that the difference is statistically significant. This shows that there are substantial differences in the ability of GPT and Deepseek models to display personality characteristics when playing Lin Daiyu, Sun Wukong, and very introverted. At the same time, the effect size Cohen's d in the Lin Daiyu role-playing task and the very introverted role-playing task are 0.921 and 0.876, respectively, both of which are large effect sizes, indicating that GPT shows better performance in this task.

## 5.5 Discussion

### 5.5.1 Comparison of human subjects and large language model measurement results

This section compares the correlation between the personality measurement results of human subjects and four LLMs (LLM, including ChatGLM3-6B, Deepseek, GPT-4o, Llama3.1-8B) under role-playing (Lin Daiyu, Sun Wukong, very introverted, very extroverted) and baseline personality. By comparing and analyzing the t-test results of humans (reference group) and the 10 role-playing test data of LLM (experimental group), the following main conclusions were found:

It seems difficult for humans to completely avoid the influence of their baseline personality in role-playing. Baseline personality not only affects the degree of deviation of individuals from their own traits, but also plays an important role in the role-playing effect. The independent sample t-test was used to analyze the deviation of the introvert group and the extrovert group from the baseline during the role-playing process, and the results showed that there was a significant difference between the two groups. In addition, the MBTI personality measurement results of the introvert and extrovert groups under the same role (such as Lin Daiyu) were subjected to an independent sample t-test, and it was found that there was still a significant difference in the measurement results of the two groups under the same role-playing conditions. This shows that although both the introvert and extrovert groups completed the role-playing task, they still showed different personality traits when playing the same role (such as Lin Daiyu). This may reflect the stability of the original baseline personality traits across situations and time.

In the role-playing task, LLM can complete the role-playing task well, but different models perform differently. It is impossible to describe all models with specific and unified rules. Different models have different characteristics, which may be related to multiple factors.

## 5.5.2 Inspiration of role-playing performance for the construction of personality research for large language models

This chapter explores the differences between LLM and humans in role-playing, in order to provide some new perspectives and ideas for building personality theories that fit LLM characteristics.

In role-playing, there is no obvious regularity between the baseline personality and the personality of the role-playing LLM, and different models show different performances, which indicates that factors such as the model's training data, structural design, input prompts, and generation strategies may affect its performance in role-playing tasks. This performance suggests that LLM may be more situation-dependent in personality performance, and its flexibility and adaptability of behavior make it difficult to directly compare the model's role-playing results with the stable personality characteristics of humans. Therefore, when exploring how to more accurately define and measure the unique personality traits of LLM, it may be necessary to combine a more flexible and multi-dimensional evaluation framework, rather than relying solely on traditional personality measurement methods.

This study provides new ideas for the construction of LLM personality theory and the measurement of personality traits, such as considering the possibility of distributed models and describing LLM behavior by capturing the score distribution under role conditions, rather than presupposing internal consistency. Future measurement tools can be optimized for model scale and context, for example, by integrating multi-role data to construct dynamic maps, further revealing the characteristics of LLM from an open perspective, and distinguishing it from the human trait framework.

## 5.6 Conclusion

This chapter compares the differences between humans and LLMs in terms of personality trait retention in role-playing tasks. The results show that humans may still be affected by their baseline personality in role-playing tasks, while LLMs' performance is affected by model parameters, role settings, and context. This difference highlights the stability of human personality across time and context, as well as the contextual sensitivity and generation randomness of LLM personality.

This study not only provides a new perspective for further building personality theories for LLMs, but also helps to deepen human understanding of the personality simulation mechanism of intelligent agents. The relevant findings can provide a reference for humans to more effectively identify, guide, and manage LLM role behaviors, thereby improving LLMs' performance and human-machine collaboration efficiency in actual application scenarios.

# Chapter 6 General Discussion

## 6.1 Discussion

This paper focuses on the Large Language Model (LLM). Aiming at the limitations of the existing LLM personality theory and measurement methods, three studies were conducted around the personality measurement method and its internal mechanism of LLM.

This study deeply explored the manifestation of LLM-specific personality traits and their similarities and differences with human personality through the comparison of test-retest stability between LLM personality measurement results and human personality measurement results, cross-variant consistency comparison, and role-playing and personality trait retention analysis, laying the foundation for building a personality theory framework and measurement method suitable for LLM. The study not only expands the application boundaries of personality psychology in the field of intelligent agents, but also provides theoretical support and practical inspiration for humans to understand the behavioral characteristics of artificial intelligence, optimize human-computer collaboration strategies, and improve the credibility and controllability of intelligent systems.

(1) In the comparison of test-retest stability between LLM and human personality measurement and the study of LLM personality theory, this paper takes the test-retest reliability in psychology as the benchmark and finds that the personality measurement results of LLM show high volatility in multiple measurements, are significantly affected by input changes, and lack the stability of humans based on internal psychological structure. The LLM distributed personality theory framework is proposed, arguing that the personality traits of LLM are not fixed, but show dynamic distribution characteristics. This finding not only reveals the trait differences between LLM and human personality, promotes human theoretical exploration of LLM behavior, but also provides reverse inspiration for human understanding of their own behavior and personality. Through comparative analysis, humans can further clarify the source of stability of their own personality and maintain correct cognition and reasonable expectations of intelligent behavior in interaction with AI.

(2) In the cross-variant consistency comparison study of LLM and human personality measurement, this study combined indirect measurement methods with CAPS theory and designed three different ways of expressing the same personality scale items to explore the impact of questioning methods on LLM personality measurement results. The results intuitively revealed the significant differences between LLM and humans in understanding the core meaning of the scale, and further reflected the difference between the internal mechanism of LLM and the human personality structure. On this basis, this study deeply analyzed the mechanism characteristics of LLM personality expression, provided theoretical support for optimizing LLM personality measurement methods, and provided methodological support for humans to more accurately understand, evaluate and apply artificial intelligence systems.

(3) In the comparative study of role-playing and personality trait retention between LLM and humans, this paper compared the performance of humans and LLM in role-playing tasks and found that the degree to which LLM's role-playing behavior is affected by the original personality traits is closely related to the model parameters. This result further reveals the dynamic and situational dependence of LLM personality, deepens human understanding of LLM personality traits, and provides theoretical and practical references for the effective management and improvement of LLM role-playing applications.

Through multi-dimensional empirical research, this paper systematically verifies the incompatibility of traditional human personality scales on LLM, and proposes a theoretical framework and methodology for LLM personality measurement. The research results not only enrich human understanding of intelligent agents represented by LLM, but also provide important theoretical references for fields such as artificial intelligence ethics and human-machine collaborative optimization, and promote the construction of a human-centered human-machine symbiotic society.

## 6.2 Shortcomings and Prospects

Although this study has made initial progress in the measurement and theoretical construction of LLM personality traits, there are still many directions worthy of further exploration to promote the in-depth development of this field.

First, at the theoretical level, future research can further improve the LLM distributed personality theory framework, and refine the generation mechanism and dynamic distribution characteristics of LLM personality traits by introducing more psychological theories (such as trait-situation interaction theory) or computational models (such as neural network dynamic analysis). In addition, the specific relationship between LLM personality traits and model architecture and training data can be explored to reveal the deep source of LLM humanoid traits and deepen human's detailed understanding and control of intelligent agents represented by LLM.

Second, at the methodological level, the variant design ideas and indirect measurement methods proposed in this study still have room for optimization. In the future, a more targeted LLM personality scale can be developed, combined with multimodal data (such as text, voice, and behavior logs) to improve the accuracy and ecological validity of measurement. At the same time, a longitudinal research design can be introduced to examine the personality stability and change rules of LLM in long-term interactions.

Finally, at the application level, the results of LLM personality research can further serve the actual scenarios of artificial intelligence. For example, in human-computer interaction, based on the findings of this study, more consistent and humanized LLM personality characteristics can be designed to improve user experience; in the field of education and psychological counseling, the impact of LLM personality traits on human behavior can be explored when LLM is used as an auxiliary tool. These applied studies will further verify the practical value of LLM personality theory, and then clarify the impact of LLM and other intelligent agents on human psychology in human-computer interaction.

In short, this article provides a new theoretical perspective and empirical basis for LLM personality research, but related issues still need to be continuously explored under interdisciplinary cooperation. Future research should go hand in hand in theoretical deepening, method innovation and application expansion, in order to fully reveal the characteristics of LLM's unique personality traits, enhance human understanding and prediction of the psychological attributes of intelligent agents, and inject new vitality into the cross-disciplinary research of artificial intelligence and psychology.

The research on LLM's unique personality traits still has broad room for expansion in the future, and subsequent research can continue to deepen from the three levels of theory, method and application.

On the theoretical level, in the future, we can further develop an interpretable dynamic personality evolution model to simulate the personality state transformation process of the large language model in different interaction situations, and reveal the dynamic distribution mechanism of its personality traits in time sequence and context. At the same time, we can also expand the scope of application of existing theoretical models, conduct comparative studies between models of different architectures and scales, and verify the cross-model universality and adaptability of LLM personality measurement tools and theoretical frameworks, so as to enrich the understanding of its unique personality traits and increase its application scenarios in human society.

On the methodological level, future research can further integrate multimodal data (such as voice, behavior logs, interaction trajectories, etc.) on the basis of the current text generation measurement method to improve the ecological validity and psychological validity of LLM personality measurement. In addition, more targeted LLM situation simulation and personality assessment tools can be developed to capture the dynamic changes of personality characteristics of the model in the real context and enhance the flexibility and adaptability of personality measurement methods.

On the practical application level, future research can further explore the psychological perception, behavioral response and trust establishment mechanism of humans in the process of interacting with LLMs with different personality traits, and explore the potential impact of LLM personality on human psychology and behavior. At the same time, as LLM is increasingly integrated into human society, in the future, from the perspectives of social psychology and ethics, we can conduct systematic research on the role positioning, behavioral norms of LLM in human society and its potential impact on human social structure and social operation mechanism, and explore new social laws and governance frameworks for the human-machine symbiotic society.

# Chapter 7 Conclusion

This study mainly draws the following conclusions:

1. The unique personality traits of LLM show the characteristics of volatility, distribution, and dynamic generation. They are significantly affected by factors such as input context, prompting methods, and model parameters, and are difficult to accurately describe through personality scales based on stable structures.

2. LLM is highly sensitive to the way of asking questions, and it is difficult for it to accurately grasp the core meaning of traditional personality scales like humans. Its understanding mechanism of language expression may be different from that of humans.

3. Human personality theory and research methods are not suitable for LLM. It is necessary to build a distributed personality theory framework for LLM.

The findings of this study reveal the significant differences between LLM personality and human personality, emphasize the dynamic and situational dependence of LLM personality, provide theoretical support for humans to more scientifically understand, predict and guide the behavior of intelligent agents, promote the construction of more credible and controllable artificial intelligence systems, and provide an important theoretical and practical basis for realizing a prosperous, harmonious and efficient human-machine symbiotic society.

# Chapter 7 Conclusion

References

发表论文和参加科研情况说明